\documentclass[11pt]{article}
\usepackage{empheq}
\usepackage{mathtools}
\usepackage{placeins}
\usepackage{epsfig}
\usepackage{amsfonts}
\usepackage{mathrsfs, bbm}
\usepackage{amssymb}
\usepackage{amstext}
\usepackage{amsmath}
\usepackage{xspace}
\usepackage{comment}
\usepackage{multirow}
\usepackage{setspace}
\usepackage[compact]{titlesec}
\usepackage{comment}
\usepackage{version}
\usepackage{enumitem}
\usepackage{multirow,subfigure}
\usepackage[pagebackref=true]{hyperref}
\hypersetup{
pageanchor=true,
bookmarks,
bookmarksnumbered,
colorlinks=true,
menubordercolor= [rgb]{0, 0, 0.8},
linkbordercolor= [rgb]{0, 0.8, 0.8},
citecolor = [rgb]{0.8, 0, 0},
linkcolor= [rgb]{0, 0, 0.8}}
\usepackage{url}
\usepackage{natbib}

\usepackage{smile}

\usepackage[mathlines]{lineno}[4.41]
  \newcommand*\patchAmsMathEnvironmentForLineno[1]{%
    \expandafter\let\csname old#1\expandafter\endcsname\csname #1\endcsname
    \expandafter\let\csname oldend#1\expandafter\endcsname\csname end#1\endcsname
    \renewenvironment{#1}%
    {\linenomath\csname old#1\endcsname}%
    {\csname oldend#1\endcsname\endlinenomath}}%
  \newcommand*\patchBothAmsMathEnvironmentsForLineno[1]{%
    \patchAmsMathEnvironmentForLineno{#1}%
    \patchAmsMathEnvironmentForLineno{#1*}}%
  \patchBothAmsMathEnvironmentsForLineno{equation}%
  \patchBothAmsMathEnvironmentsForLineno{align}%
  \patchBothAmsMathEnvironmentsForLineno{flalign}%
  \patchBothAmsMathEnvironmentsForLineno{alignat}%
  \patchBothAmsMathEnvironmentsForLineno{gather}%
  \patchBothAmsMathEnvironmentsForLineno{multline}

\makeatletter
\renewcommand{\section}{\@startsection{section}{1}{0mm}%
                                   {2ex plus -1ex minus -.2ex}%
                                   {1.3ex plus .2ex}%
                                   {\normalfont\Large\bfseries}}%
 \renewcommand{\subsection}{\@startsection{subsection}{2}{0mm}%
                                     {1ex plus -1ex minus -.2ex}%
                                     {1ex plus .2ex }%
                                     {\normalfont\large\bfseries}}%
 \renewcommand{\subsubsection}{\@startsection{subsubsection}{3}{0mm}%
                                     {1ex plus -1ex minus -.2ex}%
                                     {1ex plus .2ex }%
                                     {\normalfont\normalsize\bfseries}}
 \renewcommand\paragraph{\@startsection{paragraph}{4}{0mm}%
                                    {1ex \@plus1ex \@minus.2ex}%
                                    {-1em}%
                                    {\normalfont\normalsize\bfseries}}
\titlespacing{\section}{2pt}{*0}{*0}
\titlespacing{\subsection}{1pt}{*0}{*0}
\titlespacing{\subsubsection}{0pt}{*0}{*0}

\makeatother

\renewcommand{\theequation}{\arabic{equation}}

\allowdisplaybreaks

\graphicspath{{./figures/}}

\usepackage{cleveref}

\includeversion{paperonly}
\excludeversion{techreportonly}

\newcommand{\stab}{\textsf{stab}}
\newcommand{\bias}{\textsf{bias}}
\newcommand{\var}{\textsf{var}}
\newcommand{\bad}{\textsf{bad}}
\newcommand{\good}{\textsf{good}}
\newcommand{\Reg}{\mathcal{R}}
\newcommand{\dynalg}{\textsc{Dyn-LQR}}

\def\##1\#{\begin{align}#1\end{align}}
\def\*#1\*{\begin{align*}#1\end{align*}}
\def\cI{{\mathcal{I}}}
\def\cJ{{\mathcal{J}}}
\newcommand{\red}{\color{red}}

\usepackage[textsize=tiny]{todonotes}

\begin{document}
\title{Dynamic Regret Minimization for Control of Non-stationary Linear Dynamical Systems} 
\author{Yuwei Luo\thanks{Stanford University, Graduate School of Business} , Varun Gupta\thanks{The University of Chicago, Booth School of Business} , Mladen Kolar\footnotemark[2]}
\date{\today}
\maketitle
\vspace{-0.1 in}


\begin{abstract}
\begin{onehalfspace} 
We consider the problem of controlling a Linear Quadratic Regulator (LQR) system over a finite horizon $T$ with fixed and known cost matrices $Q,R$, but unknown and non-stationary dynamics $\{A_t, B_t\}$. The sequence of dynamics matrices can be arbitrary, but with a total variation, $V_T$, assumed to be $o(T)$ and unknown to the controller. Under the assumption that a sequence of stabilizing, but potentially sub-optimal controllers is available for all $t$, we present an algorithm that achieves the optimal dynamic regret of $\tilde{\cO}\left(V_T^{2/5}T^{3/5}\right)$. With piecewise constant dynamics, our algorithm achieves the optimal regret of $\tilde{\cO}(\sqrt{ST})$ where $S$ is the number of switches. The crux of our algorithm is an adaptive non-stationarity detection strategy, which builds on an approach recently developed for contextual Multi-armed Bandit problems. We also argue that non-adaptive forgetting (e.g., restarting or using sliding window learning with a static window size) may not be regret optimal for the LQR problem, even when the window size is optimally tuned with the knowledge of $V_T$. The main technical challenge in the analysis of our algorithm is to prove that the ordinary least squares (OLS) estimator has a small bias when the parameter to be estimated is non-stationary. Our analysis also highlights that the key motif driving the regret is that the LQR problem is in spirit a bandit problem with linear feedback and locally quadratic cost. This motif is more universal than the LQR problem itself, and therefore we believe our results should find wider application.
\end{onehalfspace}
\end{abstract}

\clearpage 
\tableofcontents

\setlength{\parskip}{.1in}

\clearpage
\section{Introduction}
\label{sec:introduction}

We look at the control of a Linear Quadratic Regulator (LQR) system with unknown and time-varying linear dynamics:
\[  
x_{t+1} = A_t x_t + B_t u_t + w_t, 
\]
with state $x_t \in \RR^n$ and control $u_t \in \RR^d$, stochastic \textit{i.i.d.}~sub-Gaussian noise process $\{w_t\}$, and a time-invariant known quadratic cost function $c(x,u) = x^\top Q x + u^\top R u$ over a horizon of $T$ periods.
LQR systems are perhaps the simplest Markov Decision Processes (MDPs) and one of the most fundamental problems studied in control theory. To quote \cite[Chapter 8]{tedrake2009underactuated}, ``one of the most powerful applications of time-varying LQR involves linearizing around a nominal trajectory of a nonlinear system and using LQR to provide a trajectory controller.'' More precisely, given a desired trajectory $x_t^0, u_t^0$ that one desires to track for a system with non-linear dynamics:
\[ \expct{x_{t+1} \mid x_t, u_t} = x_t + f(x_t,u_t), \]
we define the centered trajectories $\bar{x}_t = x_t - x^0_t, \bar{u}_t = u_t - u^0_t$, so that:
\begin{align*}
    \expct{\bar{x}_{t+1} \mid \bar{x}_t, \bar{u}_t} & = \bar{x}_t + f(x_t,u_t) - f(x_t^0, u_t^0) \\ 
    & \approx \bar{x}_t + \frac{\partial f(x_t^0,u_t^0)}{\partial x_t^0} (x_t-x_t^0)+ \frac{\partial f(x_t^0,u_t^0)}{\partial u_t^0} (u_t - u_t^0) =: A_t \bar{x}_t + B_t \bar{u}_t. 
\end{align*}
See also \citep{athans1971role} for a tutorial  treatment of use of LQR in engineering design. LQR systems, and linear dynamical systems more broadly,  have been used to model diverse  applications, such as controlling robots \citep{levine2016end}, cooling data centers \citep{cohen2018online}, control of brand dynamics in marketing \cite{naik2014marketing}, and macroeconomic policy \citep{chow1976control} to name a few. As a result, LQR systems have also been the subject of a lot of research on reinforcement learning: from model-free vs.~model-based approaches in episodic learning setting, to learning and control under unknown stationary dynamics, to robust control in the presence of an adversarial (non-stochastic) noise process. See related work in Section~\ref{sec:related}. The ability to adapt to changing dynamics lends another, arguably stronger, robustness to the control policy. However, to the best of our knowledge, the problem of learning non-stationary dynamics while controlling an LQR system has not been studied yet. We take the first steps towards this problem.

We quantify the non-stationarity of the sequence $\{ \Theta_t = [ A_t \ B_t]\}$ by the total variation $V_T = \sum_{t=1}^{T-1}\Delta_t$ with $\Delta_t := \norm{\Theta_{t+1} - \Theta_t}_F$ denoting the Frobenius norm of change of dynamics matrix $A$ and input matrix $B$ from time $t$ to $t+1$. In the case of piecewise constant dynamics, we measure the non-stationarity by the number of pieces $S_T \geq 1$.

We  measure the performance of a control (and learning) policy $\pi$ via dynamic regret metric:
\begin{align}
    \label{eqn:regret_defn}
    \Reg^{\pi}(T) &= \sum_{t=1}^T c(x_t, u_t) - J_t^*,
\end{align}
where $u_t$ denotes the action taken by policy $\pi$, and $J_t^*$ denotes the optimal average steady-state cost of the \textit{stationary} LQR system with dynamics fixed as $\Theta_t$. We also show that $\sum_t J^*_t$ is at most $\mathcal{O}(V_T)$ larger than the expected cost of the dynamic optimal policy. A fundamental result in the theory of LQR systems states that the optimal policy for an LQR system is a linear feedback control policy $u_t = K_t x_t$ for some sequence of matrices $K_t$ (see, e.g., \citep{bertsekas2012dynamic}). If the LQR system is stationary, then the infinite horizon optimal policy satisfies $K_t = K^*$. Our central result states that, given access to a nominal sequence of controllers that are potentially sub-optimal but are guaranteed to stabilize the non-stationary LQR dynamics, the proposed algorithm $\dynalg$ guarantees:
\[ \expct{\Reg^{\dynalg}(T)} = \tilde{\cO}\left( V_T^{2/5} T^{3/5} \right), \]
\textit{without} the knowledge of $V_T$ upfront.  We also demonstrate an instance showing that this regret rate is tight for any online learner/controller. The same algorithm guarantees $\expct{\Reg^{\dynalg}(T)} = \tilde{\cO}\left( \sqrt{ST} \right)$ when the dynamics are piece-wise constant with at most $S$ switches. The dependence of the regret on the dimensions $n,d$ for our algorithm and analysis is $n^2d^2$, but we believe this can be improved with a better choice of the tuning parameters in our algorithm.\footnote{For stationary LQR, \cite{simchowitz2020naive} prove that the optimal dependence is $d\sqrt{n}$, we leave the task of achieving the same dependence in non-stationary LQR as a question for subsequent research.}

The design philosophy behind our algorithm $\dynalg$ is of using \textit{certainty equivalent controllers}, that is, using the controller based on a point estimate of the model parameter (as opposed to confidence ellipsoids, for example). At a typical time $t$, $\dynalg$ employs a linear feedback control $\hat{K}_t$ based on an estimate $\hat{\Theta}_t$ of the current dynamics, with some extra exploration noise: $u_t = \hat{K}_t x_t + \sigma_t \eta_t$. Here $\eta_t \sim \mathcal{N}(0,I_d)$, and $\sigma_t$ denotes the ``exploration energy.'' A fairly simple regret decomposition lemma shows that if the policies $\hat{K}_t$ do not change very often, then the regret is dominated by \textit{(i)} the total exploration energy $\sum_t \sigma_t^2$, and  \textit{(ii)} $\sum_{t} J_t(\hat{K}_t) - J_t^*$, where $J_t(\hat{K}_t)$ denotes the average steady-state cost of the \textit{stationary} LQR system with time-invariant dynamics $\Theta_t$ and control $\hat{K}_t$. A result of \cite{simchowitz2020naive} shows 
that $J_t(\hat{K}_t) - J_t^* \lessapprox C \cdot \norm{\hat{\Theta}_t - \Theta_t}_F^2$, \textit{if} the estimation error is small enough. Thus, if we strip away the complexity introduced due to the dynamics itself, the essence of the non-stationary LQR problem is that of tracking $\Theta_t$, which boils down to \textit{a bandit problem with linear feedback and a locally quadratic loss function.} In Section~\ref{sec:conclusion} we give an example of a queueing system which also exhibits this motif, and for which we believe a similar algorithm as $\dynalg$ can give optimal dynamic regret.

Under non-stationary dynamics, it is important to forget the distant history when constructing an estimate of the current dynamics. Our approach for doing so is to adaptively restart the learning problem when ``sufficient'' change in the dynamics has accumulated, using a scheme motivated by the algorithm of \citet{chen2019new} developed for contextual multi-armed bandits. The algorithm of \citet{chen2019new} runs multiple tests in parallel, each tailored to detect changes of a different scale, by replaying (with carefully tailored probabilities) an older strategy and then comparing the new estimated reward distribution with the older reward distribution. As a result, \citet{chen2019new} were the first to obtain the optimal dynamic regret for contextual bandit problems as a function of the total variation of the reward distribution {\it without the knowledge of the variation budget}. For the LQR problem, we modify this procedure in at least two directions. First, we keep using the current controller but inject a higher exploration noise. This change is critical for our regret analysis at two places: our current analysis includes a term involving the number of policy switches and minimizing the number of policy switches impacts the regret guarantee; and, we mention below, our analysis of the estimation error of dynamics crucially relies on the linear feedback control matrix being fixed throughout the interval of estimation. Second, the probabilities with which the exploration is carried out are different for the LQR problem owing to the quadratic cost. More recently, the authors in \cite{wei2021non} outline that for many classes of episodic reinforcement learning problems, a similar strategy can be used to convert any Upper Confidence Bound (UCB) type stationary reinforcement learning algorithm to a dynamic regret minimizing algorithm. There are quite a few differences between \cite{wei2021non} and our work: the LQR problem is not covered by the classes of MDPs they consider, we look at a non-episodic version of the LQR problem, and our algorithm is certainty equivalent controller-based and not a UCB-type.  

\paragraph{Technical challenges and novelty:} We next point out three areas where the analysis in the current paper contributes to the existing literature on online learning and control. 
\begin{enumerate}
    \item {\it Ordinary Least Squares (OLS) under non-stationarity:} The biggest challenge we overcome is to prove a bound on the error of the estimated parameters $\hat{\Theta}_t$. In particular, based on the observations in some interval $\cI$, the OLS estimate $\hat{\Theta}_\cI$ of the dynamics is given by:
\[ 
\hat{\Theta}_{\cI} = \argmin_\Theta \sum_{t \in \cI} \norm{ x_{t+1} - \Theta (x_t^\top\ u_t^\top)^\top }^2 = \argmin_\Theta \sum_{t \in \cI} \norm{ ( \Theta_t - \Theta) \cdot (x_t^\top\ u_t^\top)^\top + w_t}^2.  
\]
A linear feedback controller $u_t = {K}_\cI x_t$, with ${K}_\cI$ fixed during the interval $\cI$, allows estimating the component of $\Theta_t$ parallel to the $n$-dimensional column space of $[x_t^\top \ u_t^\top]=[I_n\ K_\cI^\top]^\top x_t$, but not in the orthogonal subspace. This problem shows up even in stationary LQR, and is the reason we use the exploration noise $\sigma_t \eta_t$ in $u_t$. However, for stationary LQR, this is only a mild problem -- the estimate is unbiased by default and the condition number of the (ill-conditioned) Hessian is sufficient to bound the variance of the OLS estimator. Under non-stationary $\Theta_t$, even proving that the OLS estimate $\hat{\Theta}_{\cI}$ is ``unbiased,'' i.e., close to $\Theta_t$ for $t \in \cI$ even when all the $\Theta_t$ in $\cI$ are close to each other, is not trivial. Naively using the condition number of the Hessian would require a larger $\sigma_t$, and, thus, result in a suboptimal regret. A major chunk of the technical analysis is to show that a small exploration cost is sufficient to guarantee that $\hat{\Theta}_\cI$ has small bias. This requires quite a delicate analysis of the geometry of the Hessian, as well as an interplay with the algorithm itself where we need to keep the policy $K_\cI$ fixed so that the column space of $[I_n \ K_\cI^\top]^\top$ is fixed. This is where we crucially take advantage of the fact that instead of replaying an old policy as in \cite{chen2019new} to detect non-stationarity, we continue playing the same linear feedback controller and only increase the exploration noise. 
\item {\it Continuous and unbounded state space:} The second challenge comes from the fact that the LQR system has unbounded state space. A particular complication this creates is that the certainty equivalent controller need not stabilize the dynamics under non-stationarity, and therefore the norm of the state can blow up. Algorithmically, we solve this problem by falling back on the nominal sequence of controllers when the norm of the state crosses a threshold, and until it falls below another threshold. Analytically, this requires some careful analysis to bound the total cost incurred during such phases.
\item {\it An impossibility result for non-adaptive restart algorithms:} We prove a novel regret lower bound that outlines a shortcoming of a popular strategy for non-stationary bandits/reinforcement learning. As we mentioned earlier, to forget distant history for non-stationary bandits and episodic reinforcement learning, almost all existing algorithms restart learning at a fixed schedule, or use sliding window based estimators with a fixed window size. For all the flavors of non-stationary bandit or reinforcement learning problems studied in the literature, this strategy yields the optimal regret {\it if the window size is tuned optimally with the knowledge of the variation budget, or using a bandit-on-bandit technique}. In Theorem~\ref{thm:lowerbound_window_based} we prove that for the non-stationary LQR problem, for a wide class of fixed window size based algorithms, this approach can not give the optimal regret rate even with the knowledge of $V_T$. This crucially uses the fact that the LQR problem behaves like a bandit problem with non-linear (in particular quadratic) loss function. We believe that the same lower bound should extend to non-linear bandit problems more generally.
\end{enumerate}




\paragraph{Paper Outline:} We survey some of the relevant literature in Section~\ref{sec:related}. In Section~\ref{sec:stationary}, we first present some classical results on control of stationary LQR and recent results on learning and control. Then in Section~\ref{sec:model} we present the model assumptions for the non-stationary LQR problem that is the subject of our study. In Section~\ref{sec:algorithm}, we present our proposed algorithm \dynalg. We devote Section~\ref{sec:ols_esti} to highlighting the technical challenge in studying the error of the OLS estimator for non-stationary LQR. In Section~\ref{sec:analysis} we present the regret upper  bound for \dynalg, and in Section~\ref{sec:lower} we present two lower bound results.

\paragraph{Notation:}  All vectors are column vectors. For a matrix $A$, we use $\norm{A} = \sup_{\norm{x}=1} \norm{A x}$ to denote the operator norm and $\norm{A}_F = \sqrt{\sum_{i,j} a^2_{ij}}$ to denote the Frobenius norm.  For two square matrices $A, B$, we use $A \preccurlyeq B$ to denote that the matrix $B-A$ is positive semidefinite. The $\cO()$ notation will used to suppress problem dependent constants, including the dimensions $d,n$; the $\widetilde{\cO}()$ notation further suppresses $\polylog T$ factors.  

\section{Related Work}
\label{sec:related}

Our work touches on many themes in online learning and control. For each, we mention only a few papers relevant to the present work and make no attempt to present an exhaustive survey. 

\paragraph{Learning and control of stationary LQR:} The study of learning and control of LQRs was initiated in \citet{abbasi2011regret}, who presented an $\cO(\sqrt{T})$ regret algorithm based on the Optimism in the Face of Uncertainty (OFU) principle, but with an exponential dependence on the dimensionality of the problem. \citet{ibrahimi2012efficient} improved dependence on the dimensionality to polynomial. \citet{cohen2019learning} was the first paper that provided a computationally efficient algorithm with $\cO(\sqrt{T})$ regret for the stationary LQR problem by solving for the optimal steady-state covariance of $[x_t^\top u_t^{\top}]$ via a semi-definite program and extracting a controller from this covariance.  \citet{faradonbeh2020input} and \citet{mania2019certainty} proved that the certainty equivalent controller is efficient and yields $\cO(\sqrt{T})$ regret. \citet{simchowitz2020naive} proved a matching upper and lower bound on the regret of the stationary LQR problem of $\tilde{\Theta}(\sqrt{nd^2T})$, settling the open question of whether logarithmic regret may be possible for LQR (due to the strongly convex loss function). Notably,  the upper bound in \citet{simchowitz2020naive} was achieved by a variant of the certainty equivalent controller. \citet{cassel2020logarithmic} proved an $\Omega(\sqrt{T})$ lower bound and showed that naive exploration based algorithms can indeed attain logarithmic regret when the problem is sufficiently non-degenerate. \cite{jedra2021minimal} developed a certainty equivalent controller based strategy for stationary LQR, but allow the controller to change arbitrarily quickly, rather than according to a fixed doubling schedule as in prior work. 

\paragraph{Dynamic regret minimization for experts and bandits:} Due to the weakness of static regret as a metric for environments with non-stationary or adversarial losses/rewards, numerous stronger notions of regret have been proposed and studied. One of the first such results was in the seminal paper of \citet{zinkevich2003online}, where a regret parameterized by the total variation of the comparator sequence of actions was proved. \citet{herbster1998tracking} proposed the \textsc{FixedShare} algorithm for prediction with expert advice problem, where the best expert may switch during the time horizon. \citet{hazan2009efficient} looked at online convex optimization with changing loss functions, and proposed a metric for adaptive regret, defined to be the maximum over all windows of the regret of the algorithm on that window compared to the best fixed action for that window. \citet{daniely2015strongly} introduced a metric of strongly adaptive regret and proved that no algorithm can be strongly adaptive in the bandit feedback setting. For the bandit setting, the most common approach towards dynamic regret is to assume that the non-stationary sequence has bounded total variation, and providing min-max regret guarantees as a function of the variation, e.g., \citet{besbes2014stochastic}. The common design technique is to use periodic restarts or discounting with the knowledge of the variation of rewards, e.g., \citep{garivier2011upper,yoan2019weighted}, or a bandit-on-bandit technique to learn the optimal window size as in \citet{cheung2019learning}, but with a suboptimal regret guarantee. A recent breakthrough was achieved by the algorithm of \citet{chen2019new}, which performs a very delicate exploration and uses an adaptive restart argument to attain the optimal regret rate for contextual multi-armed bandits without any prior knowledge of the variation.

\paragraph{Reinforcement learning for non-stationary MDPs:} While there is some literature on regret minimization for MDPs with fixed transition kernel, but a changing sequence of cost functions \citep{yu2009markov, ortner2020variational}, the work on unknown non-stationary dynamics is much more recent \citep{gajane2018sliding, cheung2019non}. The main idea is to use sliding window based estimators of the transition kernel and design a policy based on an optimistic model of the transition dynamics within the confidence set. As we mentioned earlier, sliding window based algorithms are provably regret-suboptimal for the LQR problem due to the quadratic cost function. In parallel with this work, \cite{wei2021non} proposed an adaptive restart approach  for non-stationary reinforcement learning that uses any UCB-type algorithm for stationary reinforcement learning as a black box. The authors show that for many tabular or linear MDP settings, their approach gives the state-of-the-art regret without knowledge of variation of the input instance. While the LQR problem is neither tabular nor linear, our approach is similar in its spirit to \cite{wei2021non} -- however, we use point estimates and explicit exploration instead of using a UCB-like approach. 

\paragraph{Robust control of LQR under adversarial noise:} While we consider the robust control of LQR systems from the perspective of changing transition dynamics, there have been some recent results on robust control of LQR when the noise $w_t$ is adversarial. \citet{hazan2020nonstochastic} considered a ``stationary'' LQR system with known $A,B$, but with adversarial noise, and proposed an algorithm with $\cO(T^{2/3})$ regret against the best linear controller in hindsight.  \citet{simchowitz2020improper} looked at the same problem when the $A,B$ matrices may or may not be known, and proposed a Disturbance Feedback Control based online  control policy with sublinear regret against all stabilizing policies. Finally, \citet{goel2020regret, gradu2020adaptive} looked at non-stationary LQR problems with adversarial noise. \citet{goel2020regret} assumed that the sequence $A_t, B_t$ is known upfront and proposed a controller with optimal dependence of regret on the total noise. \citet{gradu2020adaptive} assumed that the dynamics matrices $A_t, B_t$ are observed after the action $u_t$ is taken and proposed a policy with strongly adaptive regret guarantee. Finally, we would like to point to \cite{boffi2021regret} as a recent example of a work on learning and control of non-stationary non-linear dynamical systems, although in this work the non-stationary dynamics are linearly parameterized by a known non-stationary sequence of basis matrices and an unknown stationary parameter.


\section{Preliminaries -- Stationary LQR}
\label{sec:stationary}

In this section, we give a brief summary of the classical theory of stationary LQR systems and some recent work on learning and control for stationary LQR systems that lays the groundwork for our work on non-stationary LQR. The stationary dynamics, parameterized by $\Theta = [A \ B]$, are given by:
\[ x_{t+1} = A x_t + B u_t + w_t, \quad t\in[T], \]
and the cost function by: 
\[ c(x_t, u_t) = x_t^\top Q x_t + u_t^\top R u_t, \]
where $x_t \in \RR^n$ denotes the state, $u_t \in \RR^d$ the control (or input), $w_t$ are i.i.d.~stochastic noise (disturbance) with covariance matrix $W$,  and $Q,R$ are positive-definite matrices.

A classical result in the theory of LQR problems is that the value function of the LQR problem is a quadratic function of the state. This is true even for non-stationary dynamics and can be most easily seen by solving for the optimal control for a finite horizon problem via backward Dynamic Programming. As a consequence, the optimal controller turns out to be a linear feedback controller $u_t= K_t x_t$, for some sequence of control matrices $\{K_t\}$. In the special case of infinite horizon average cost minimization, the control is stationary with $K_t = K^*$. For an arbitrary linear feedback controller $K$ that is stabilizing, i.e., the spectral radius of $A+BK$ is upper bounded away from 1, we denote by $J(\Theta, K)$  the infinite horizon average cost and by the symmetric positive definite matrix $P(\Theta, K)$  we denote the quadratic \textit{relative value function} (also called the \textit{bias function}) for the infinite horizon average cost problem, satisfying the following Bellman equation:
\begin{align*}
    x^\top P(\Theta, K)x &=  c(x,Kx) - J(\Theta, K) + \expct{ x_1^\top P(\Theta, K) x_1 | x_{0} = x } \\
    &=  x^\top (Q+K^\top R K) x  - J(\Theta, K) + x^\top(A+BK)^\top P(\Theta, K) (A+BK)x + \expct{ w^\top P(\Theta,K) w }.
\end{align*}
Matching the quadratic and the constant terms, we get that $P(\Theta,K)$ solves the following equation 
\[ P = Q + K^\top R K + (A+BK)^\top P (A+BK) \]
and $J(\Theta, K) = \Tr(P(\Theta, K)W)$. Let the optimal bias function be denoted by $P^*(\Theta)$ and the optimal linear feedback controller by $K^*(\Theta)$.  Given $P^*(\Theta) = P^*$, the optimal linear feedback controller $K^*=K^*(\Theta)$ can be obtained by solving for the cost minimizing action in the Bellman equation:
\#
\label{eqn:Kstar}
K^* = -(R+B^\top P^* B)^{-1}B^\top P^*A.
\#
Plugging the above in the equation for $P(\Theta, K)$ gives a fixed point equation (called the Discrete Algebraic Ricatti Equation) for $P^*(\Theta)$:
\# 
\label{eqn:Pstar}
P^* = Q + A^\top P^* A - A^\top P^* B (R + B^\top P^* B) B^\top P^* A.
\#
While the explicit forms of $K^*(\Theta), P^*(\Theta)$ are not essential for following the results in the paper, we would like to point out that neither of them depend on the covariance of the noise process, even though the optimal cost $J^*(\Theta)$ does. 

Finally, consider the policy $u_t = K x_t + \sigma \eta_t$, where  $\eta_t$ are \textit{i.i.d.}~with covariance $I_d$ and $\sigma > 0$. Denote the average cost for this policy by $J(\Theta, K, \sigma)$ and the relative value function by $P(\Theta, K, \sigma)$. Then, 
\begin{align}
\nonumber    P(\Theta, K, \sigma) &= P(\Theta, K), \\
    J(\Theta, K, \sigma) &= J(\Theta, K) + \sigma^2 \Tr{\left( R + B^\top P(\Theta, K) B\right)}.
    \label{eqn:noise_J}
\end{align} 
That is, the effect of additive noise in the controller completely decouples from the cost of the noiseless control $K x_t$.

\paragraph{Cost of model estimation error:} The following lemma from \cite{simchowitz2020naive} will be central for the intuition and analysis behind learning and control of LQR.
\begin{lemma}[{\citet[Theorem 5]{simchowitz2020naive}}]
\label{lem:quadratic}
Let $\Theta = [A \  B]$ be a stabilizable system and $\hat{\Theta} = [ \hat{A} \ \hat{B} ]$ be an estimate of $\Theta$. Then there exist constants $C_1, C_2$, depending on $R, Q, W$, such that if $\max\left\{\norm{A - \hat{A}}, \norm{B - \hat{B}} \right\} \leq  C_1 \norm{P^*(\Theta)}^{-5}$, then
\[ J^*(\Theta) - J(\Theta,K^*(\hat{\Theta})) \leq C_2 \norm{P^*(\Theta)}^8 \norm{\Theta - \hat{\Theta}}_F^2.
\]
\end{lemma}
The lemma implies that the certainty equivalent controller $K^*(\hat{\Theta})$ based on the estimate $\hat{\Theta}$ with sufficiently small error $\epsilon$ leads to a suboptimality of at most a problem-dependent constant times $\epsilon^2$. Note that the closer the spectral norm of the closed loop $A+BK^*(\Theta)$ is to 1, the larger is $\norm{P^*(\Theta)}$, and the harder it is to satisfy the condition in Lemma~\ref{lem:quadratic}.

\paragraph{A naive exploration algorithm:} 
To get some intuition on the fundamental exploration-exploitation trade-off for the LQR problem, we describe a bare bones version of the algorithm from \cite{simchowitz2020naive} for the stationary setting. The authors assume (as is common in the literature) access to a stabilizing, but suboptimal controller $K_0$. The algorithm begins by playing $u_t = K_0x_t + \sigma_0 \eta_t$ with $\eta_t \stackrel{i.i.d.}{\sim} \mathcal{N}(0,I_d)$ and $\sigma^2_0 = 1$ for a sufficiently long warm-up period $L$. Based on this warm-up period, an initial estimate $\hat{\Theta}_1$ is constructed using the ordinary least squares (OLS) estimator. The quantity $\sigma^2_0$ denotes the \textit{exploration noise/energy}.  Even though the LQR dynamics adds \textit{i.i.d.} noise $w_t$ to the state, the exploration noise $\sigma^2_0 \eta_t$ is necessary because the vector $[x_t^\top \   u_t^\top]^\top = [I_n\ K_0^\top]^\top x_t$ lives in an $n$-dimensional subspace instead of the full $(n+d)$-dimensional subspace.
The algorithm then proceeds in blocks of doubling length, indexed by $i=1,2,\ldots$. Block $i$ is of length $\tau_i = L \cdot 2^i$. In block 1, the control is chosen as $u_t = K_1 x_t + \sigma_1 \eta_t$ where $K_1 = K^*(\hat{\Theta}_1)$ and $\sigma^2_1 = 1/\sqrt{\tau_1}$. The observations from block 1 are used to construct an estimate $\hat{\Theta}_2$ and the control in block 2 is $u_t = K_2x_t +\sigma_2 \eta_2$ with $K_2 = K^*(\hat{\Theta}_2)$ and $\sigma^2_2 = 1/\sqrt{\tau_2}$. More generally, observations from block $(i-1)$ are used to create an estimate $\hat{\Theta}_{i}$ and controller $K_{i}=K^*(\hat{\Theta}_i)$. The control in block $i$ is $u_t = K_i x_t + \sigma_i \eta_t$, with exploration noise $\sigma^2_i = 1/\sqrt{\tau_i}$. The intuition behind the choice of exploration noise is the following. The total exploration energy invested in block $i$ is $\tau_i \sigma^2_i$, which, by \eqref{eqn:noise_J}, increases the cost by an order $\tau_i \sigma^2_i$. Furthermore, the variance of the OLS estimator $\hat{\Theta}_{i+1}$ is inversely proportional to the exploration noise, and is therefore $\cO(1/\tau_i \sigma^2_i)$. Lemma~\ref{lem:quadratic} then says that the per step exploitation cost from using controller $K_{i+1}$ based on $\hat{\Theta}_{i+1}$ is of the order $1/\tau_{i}\sigma^2_{i}$. Therefore, the total regret is of order $1/\sigma^2_i$ during block $(i+1)$. Balancing the exploration cost $\tau_i \sigma^2_i$ during block $i$ and the total exploitation cost $1/\sigma^2_i$ during block $i+1$ gives the choice $\sigma^2_i \approx \tau_i^{-1/2}$.

\section{Model and Preliminaries -- Non-stationary LQR}
\label{sec:model}

The non-stationary LQR problem has dynamics: 
\[ 
x_{t+1} = A_t x_t + B_t u_t + w_t, \quad t\in[T],
\]
and time-invariant cost function: 
\[ 
c(x_t, u_t) = x_t^\top Q x_t + u_t^\top R u_t, 
\]
where $x_t \in \RR^n$ denotes the state, $u_t \in \RR^d$ the control (or input), $w_t \stackrel{i.i.d.}{\sim} \mathcal{N}(0, W)$ denotes the stochastic noise (disturbance)  with covariance matrix $W = \psi^2 I_n$ (the assumption on $w_t$ is for exposition purposes; our results readily extend to sub-Gaussian $w_t$ with $\psi^2 I_n \preccurlyeq W \preccurlyeq \Psi^2 I_n$ for $0 < \psi < \Psi < \infty$). We use $\{\mathcal{F}_t\}_{t \in [T]}$ to denote the filtration generated by $\{w_1, \ldots, w_T\}$. We will use $\Theta_t = [A_t\ B_t]$ to succinctly denote the dynamics of the LQR at time period $t$. Cost matrices $Q, R$ are assumed to be symmetric positive definite with $ r_{\min} I_d \preccurlyeq R \preccurlyeq r_{\max} I_d, q_{\min} I_n \preccurlyeq Q \preccurlyeq q_{\max} I_n$.

The learner/controller knows the cost matrices $Q,R$, but not the dynamics $\{\Theta_t\}_{t\in [T]}$. For any interval $\cI = [s,e]$, we define the total variation of the model parameter within the interval as
\[  
\Delta_\cI = \Delta_{[s,e]} := 
\sum_{t=s}^{e-1} \Delta_{s} = \sum_{t=s}^{e-1} \norm{\Theta_s - \Theta_{s+1}}_F, 
\]
so that the total variation $V_T = \Delta_{[1,T]}$. In the case of piecewise constant dynamics, we use $S_\cI \geq 1$ to denote the number of such constant dynamics pieces in interval $\cI$.

A common assumption in the literature on online learning and control of stationary LQR systems is the availability of a baseline controller $K_0$ that may be suboptimal, but stabilizes the system. Such a controller can be played in an initial warm-up phase until a good initial estimate of the dynamics can be learned. This assumption allows one to focus on the algorithmic challenge of minimizing regret and not worry about the stability of the system. From the point of view of applications, often there are default actions which guarantee this condition (e.g., shutting a data center will prevent over-heating of servers), or crude forecasts of the dynamics may be enough to derive such controls. Theoretically, a stabilizing controller can be found by following the strategy proposed in \cite{faradonbeh2018finite}. Similarly, we also assume that our algorithm is given a sequence of controllers $\{K_t^\stab\}$ that stabilizes the dynamics given by $\{\Theta_t\}$. More formally, Assumption~\ref{assum:stabilization} states that the exogenous sequence of controllers satisfies a property called \textit{sequentially strong stability}.  

\begin{definition}[Sequentially Strong Stability \cite{cohen2018online}]
\label{defn:SSS} 
For the non-stationary LQR problem with parameters $\{\Theta_t\}= \{[A_t \ B_t]\}$, a sequence of controllers $\{K_1, \ldots, K_T \}$ is called $(\kappa, \gamma)$ sequentially strongly-stabilizing (for $\kappa \geq 1 $ and $0 < \gamma \leq  1$) if there exist matrices $H_1, H_{2}, \ldots, H_T \succ 0$ and $L_1, L_{2}, \ldots, L_T$ such that $A_t + B_t K_t = H_t L_t H_t^{-1}$ for all $t\in[T]$, and the following properties hold:
\begin{enumerate}[label=(\roman*),topsep=0ex]
\item $\norm{L_t} \leq 1-\gamma$ and $\norm{K_t} \leq \kappa$ for $t\in[T]$;
\item $\norm{H_t} \leq B_0$ and $\norm{H_t^{-1}} \leq 1/b_0$ with $\kappa = B_0/b_0$ for $t\in[T]$;
\item $\norm{H_{t+1}^{-1} H_t} \leq 1 + \gamma/2$ for $t\in[T-1]$.
\end{enumerate}
\end{definition}

\begin{assumption}
\label{assum:stabilization}
The online algorithm has access to a sequence of $(\kappa, \gamma)$ sequentially strongly-stabilizing controllers $\{K^\stab_1, K^\stab_2, \ldots, K^\stab_T \}$, for  constants $\kappa \geq 1$ and $0 < \gamma \leq  1$.
\end{assumption}

A $(\kappa',\gamma')$ sequentially strongly stabilizing sequence of controllers is also $(\kappa, \gamma)$ strongly stabilizing for $\kappa \geq \kappa'$ and $\gamma \leq \gamma'$. Therefore, we take $\kappa \geq 1$ as a convenient convention. An intuitive explanation for this assumption is the following. Denote 
\[ 
\Phi_t  := A_t + B_t K_t^\stab 
\qquad \mbox{and} \qquad 
\Phi_{b:a} := \Phi_b \Phi_{b-1} \cdots \Phi_{a}, \quad \mbox{for } 1\leq a \leq b \leq T.
\]
Then
\begin{align}
\nonumber
\norm{\Phi_{b:a}} &= \norm{H_b L_b (H_{b}^{-1} H_{b-1}) L_{b-1}   \cdots  (H_{a+1}^{-1} H_{a})L_a H_{a}^{-1}} \\
\nonumber
& \leq \norm{H_b}\cdot \norm{L_b}\cdot \norm{H_{b}^{-1} H_{b-1}} \cdot\norm{L_{b-1}}  \cdots  \norm{H_{a+1}^{-1} H_{a}}\cdot\norm{L_a} \cdot \norm{H_{a}^{-1}} \\
& \leq \kappa \left(1 + \frac{\gamma}{2}\right)^{b-a}\left(1 -  \gamma\right)^{b-a+1} \ \leq \ \kappa \left(1 - \frac{\gamma}{2}\right)^{b-a}.
\label{eqn:normPhi}
\end{align}
 As a consequence of \eqref{eqn:normPhi} and noting that
\[ x_{b} = \Phi_{b-1:a}x_a + \Phi_{b-1:a+1} w_{a} + \Phi_{b-1:a+2} w_{a+1} + \cdots + \Phi_{b-1:b-1} w_{b-2}  + w_{b-1}, 
\]
we can bound the norm of the state under
the stabilizing controllers as:
\begin{align}
    \nonumber
    \norm{x_b} & \leq \kappa e^{-\gamma (b-a)/2} \norm{x_a} + \frac{2\kappa}{\gamma} \max_{a\leq t \leq b-1} \norm{w_t}, \\
    \intertext{and,}
    \nonumber
    \expct{\norm{x_b}^2} & = \norm{\Phi_{b-1:a}x_a}^2 + \expct{\norm{\Phi_{b-1:a+1} w_{a}}^2} + \cdots + \expct{\norm{\Phi_{b-1:b-1} w_{b-2}}^2}  + \expct{\norm{w_{b-1}}^2}  \\ 
    & \leq \ \kappa^2 e^{-\gamma (b-a)} \norm{x_a}^2 + \frac{2\kappa^2}{\gamma} \expct{\norm{w}^2}.
    \label{eqn:ss_norm_x}
\end{align}
While assuming $\norm{\Phi_t} < 1-\gamma/2$ also ensures \eqref{eqn:normPhi}, it is a much more restrictive condition. A weaker condition is that the spectral radius is bounded: $\rho(\Phi_t) \leq 1-\gamma/2$, but the spectral radius is not submultiplicative and does not imply \eqref{eqn:normPhi}.

A second assumption we will make is on the stability of the controller derived from an \emph{accurate} estimate of the true dynamics. 
\begin{assumption}
\label{assum:approx_stabilization}
For any $t \in [T]$, let $\Theta_t$ be the true dynamics, $\hat{\Theta}_t$ be an estimate of the true dynamics, and $\hat{K} = K^*(\hat{\Theta}_t)$ be the optimal closed-loop controller for the estimated dynamics. Then, there exist constants $C_3, C_4$ such that $\norm{\hat{\Theta}_t - \Theta_{t}}_F^2 \leq  C_3$ implies $ J^*(\Theta_t) - J(\Theta_t,\hat{K}) \leq C_4 \norm{\Theta_t - \hat{\Theta}_t}_F^2$. For convenience, we assume $C_3 \leq 1$, since the assumption continues to hold if we choose a smaller value of $C_3$ than sufficient. 
\end{assumption}
Assumption~\ref{assum:approx_stabilization} is without loss of generality due to Lemma~\ref{lem:quadratic}. As mentioned earlier, the constants $C_3, C_4$ depend on the maximum operator norm of $P^*_t$, which we assume to be bounded independent of $T$ and $V_T$. The constants $C_3, C_4$ are only used in the analysis, not as a part of the algorithm.

Just like Assumption~\ref{assum:stabilization}, under non-stationary dynamics, we need a stronger sequential stability property for controllers $K^*(\hat{\Theta}_t)$ than in Assumption~\ref{assum:approx_stabilization}. 
Towards that end, we introduce a strengthening of the $(\kappa,\gamma)$ sequential strong stability criterion. The main difference is that condition \textit{(iii)} involves the variation $\|\Theta_{t+1} - \Theta_{t}\|$ and hence allows us to prove exponential stability for non-stationary dynamics with small total variation.

\begin{definition}[$(\kappa, \gamma, \nu)$-Sequentially Strong Stability] 
For the non-stationary LQR problem and an interval $[a,b]$, a sequence of controllers $\{K_a, \ldots, K_b \}$ is called $(\kappa, \gamma, \nu)$-sequentially strongly-stabilizing (for $\kappa \geq 1 $ and $0 < \gamma \leq  1$) if there exist matrices $H_a, H_{a+1}, \ldots, H_b \succ 0$ and $L_a, L_{a+1}, \ldots, L_b$ such that $A_t + B_t K_t = H_t L_t H_t^{-1}$ for all $t\in[a,b]$, and the following properties hold:
\begin{enumerate}[label=(\roman*),topsep=0ex]
\item $\norm{L_t} \leq 1-\gamma$ and $\norm{K_t} \leq \kappa$ for $t\in[a,b]$;
\item $\norm{H_t} \leq B_0$ and $\norm{H_t^{-1}} \leq 1/b_0$ with $\kappa = B_0/b_0$ for $t\in[a,b]$;
\item $\norm{H_{t+1}^{-1} H_t} \leq 1 + \nu \cdot \norm{\Theta_{t+1}-\Theta_t}$ for $t\in[a,b-1]$.
\end{enumerate}
\end{definition}

The next lemma states that if the provided estimate $\hat{\Theta}$ satisfies $\|\hat{\Theta}-\Theta_t\|_F^2\leq C_3$ for all $t$ in an interval $\cI$, then the controller $\hat{K}  = K(\hat{\Theta})$ is $(\kappa,\gamma, \nu)$-sequentially strongly stable for the dynamics in $\cI$.
\begin{lemma}
\label{lem:nonstat_seq_stg_stab} For an interval $\cI$, let $\hat{\Theta}$ be an estimate of the dynamics such that $\|\hat{\Theta}-\Theta_t\|_F^2\leq C_3 $ for all $t\in \cI$. Let $\hat{K}  = K^*(\hat{\Theta})$ be the optimal linear feedback controller with respect to the estimate $\hat{\Theta}$.  Define 
\*
\nu = \frac{2(1-\gamma)^2}{1-(1-\gamma)^2}\left((1-\gamma)+ (\kappa+1)\right).
\*
Then $\hat{K}$ is a $(\kappa,\gamma, \nu)$-sequentially strongly stable control sequence for interval $\cI$ with the following setting of parameters: $H_{t} = P_{t}^{1 / 2}$ and $L_t = P_{t}^{-1 / 2}\left(A_{t}+B_{t} \hat{K}\right) P_{t}^{1 / 2}$, where $P_t := P(\Theta_t, \hat{K} )$, $\kappa = \sqrt{\frac{\tilde{J}^*_{\cI}}{\psi^{2} r_{\min}}}$, $\gamma =  \frac{q_{\min}\psi^{2}}{2J^*_{\cI}}$, $J^*_\cI= \max_{t\in\cI} J^*(\Theta_t)$, and $\tilde{J}^*_\cI = J^*_\cI + C_3C_4$.
\end{lemma} 

As a corollary, similar to the calculations in \eqref{eqn:ss_norm_x}, the following lemma bounds the norm of $x_t$. 
\begin{lemma}
\label{lem:nonstat_seq_stg_stab_bdd_norm}
Let the controller $\hat{K}$ and interval $\cI = [s_\cI, e_\cI]$ satisfy the conditions in Lemma~\ref{lem:nonstat_seq_stg_stab}. Then for an action sequence $u_t = \hat{K}x_t + \sigma_t\eta_t$, $t\in\cI$, there exists a constant $C_{ss}$ such that
\*
\left\|x_{t}\right\| \leq   \kappa e^{-\gamma(t-1)+C_{ss} V_{[1,t-1]}} \|x_1\| +
\frac{\kappa e^{-\gamma(t-s) + C_{ss} V_{[s,t-1]}} }{\gamma} \max _{1<s<t}\left\|w_{s} + \sigma_s B_s \eta_s\right\|, \quad t \in \cI.
\*
\end{lemma}

Later we will see that the controllers used in our proposed Algorithm~\ref{alg:DYN-LQR} satisfy the conditions of Lemmas~\ref{lem:nonstat_seq_stg_stab} and \ref{lem:nonstat_seq_stg_stab_bdd_norm}, and hence stabilize the dynamics and the state has bounded norm with high probability.

Finally, we introduce some constants that we will use as a parameterization of the input instance. We assume that they are known to the learner/controller.

\paragraph{Additional Constants:}  Let the norm upper bounds for the parameters of the instance be given by:  $A_u=\max_{t\in [T]}\norm{A_t}$, $B_u=\max_{t\in [T]}\norm{B_t}$, $\Theta_u=\max_{t\in [T]}\norm{\Theta_t}$, and $P_u =\max_{t\in [T]} \norm{P^*_t}$. Define
    \[ \beta := \max\left\{ \psi,  \max_{i,t} \beta_{i,t} \right\},  \]
    where $\beta_{i,t}$ are singular values of $B_t$. 
    Define $K_u$ as:
    \[  K_u = \max_{t} \left\{ \norm{K_t^\stab} , \max_{\hat{\Theta} : \norm{\hat{\Theta} - \Theta_t}_F^2 \leq C_3} \norm{K^*(\hat{\Theta})}  \right\}. 
    \]
    Finally, define $\rho_0 = 1 - \gamma_{\min}/2$ and $\kappa = \kappa_{\max}$, where $\gamma_{\min}$ is the smaller of $\gamma$ values from Assumptions~\ref{assum:stabilization} and  Lemma~\ref{lem:nonstat_seq_stg_stab}, and similarly $\kappa_{\max}$ is the larger of the $\kappa$ values.

\section{Algorithm \dynalg}
\label{sec:algorithm}

\begin{figure}[ht!]
    \centering
    \includegraphics[width=6.8in]{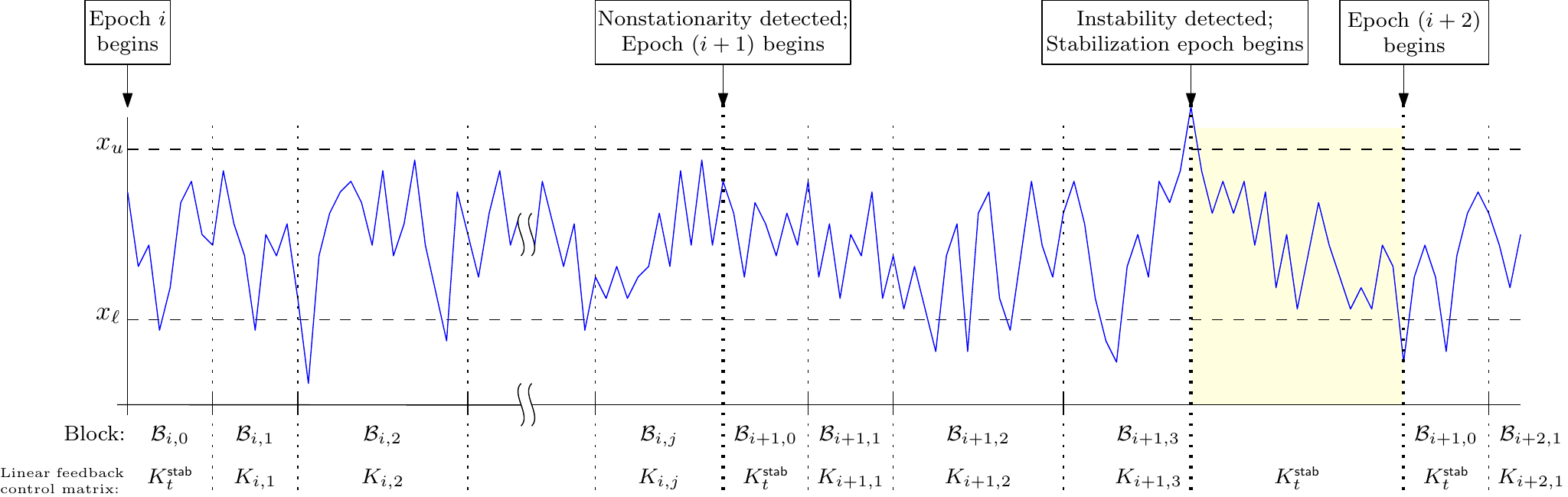}
    \caption{Illustration of Regular epochs, blocks, and stabilization epochs for Algorithm $\dynalg$. Epoch $i$ ends in block $\cB_{i,j}$ when an \textsc{EndOfExplorationTest} fails. Epoch $i+1$ ends because $\norm{x_t}$ exceeds the threshold $x_u$, indicating that the current controller $K_{i+1,3}$ is potentially unstable. This triggers a stabilization epoch which ends the first time $\norm{x_t}$ falls below $x_\ell$, and starts epoch $i+2$. }
    \label{fig:algorithm_illustration}
\end{figure}

Our algorithm \dynalg\ is presented as Algorithm~\ref{alg:DYN-LQR}. At a high level, the algorithm divides the time horizon into epochs $\{\cE_1, \cE_2, \ldots\}$ where the squared total variation $\Delta^2_{\cE_i}$ within epoch $\cE_i$ is of the order $\sqrt{1/|\cE_i|}$. This should be reminiscent of the trade-off described in the last paragraph of Section~\ref{sec:stationary} where the variance of the OLS estimator for a block was proportional to the inverse square root of the length of the block. The end of an epoch signals that a sufficient change in $\Theta_t$ has accumulated and the algorithm starts a new epoch, whereby it forgets the past history and restarts the procedure to estimate the dynamics $\Theta_t$. Since the length of an epoch is unknown to the online controller a priori, within each epoch we follow a doubling strategy (again similar to the naive algorithm in Section~\ref{sec:stationary}) by further splitting it into non-overlapping blocks (indexed by $j = 0,1,\ldots$) of geometrically increasing duration. We denote the $j$-th block of epoch $i$ as $\cB_{i,j}$. During block 0, or the \textit{warm-up block}, the algorithm plays an action $u_t = K^\stab_t x_t + \nu_0 \eta_t$ where $\eta_t \sim \mathcal{N}(0, I_d)$ are \textit{i.i.d.}~Gaussian random vectors, and $\nu_0=1$ is the added exploration noise. We denote by $\{\cG_t\}_{t\in[T]}$ the filtration generated by $\{\eta_1, \ldots, \eta_T\}$. The duration of the warm-up blocks is $L = \cO((n+d)\log^3 T)$. The $\cO(1)$ exploration noise reduces the estimation error of the OLS estimate computed at the end of the block. Observations from block $j$ are used to create an estimate $\hat{\Theta}_{i,j}$ of the dynamics, which in turn gives the linear feedback controller for block $j+1$ as $K_{i,j+1} := K^*(\hat{\Theta}_{i,j})$, and action $u_t = K_{i,j+1} x_t + \nu_{j+1} \eta_t$. For a block $\cB_{i,j}$ with $j \geq 1$, we choose $\nu^2_{j} \approx \frac{1}{\sqrt{|\cB_{i,j}|}}$ as the exploration noise similar to the stationary LQR case. If the estimate based on a block $\cB_{i,j}$ ``differs statistically''   from the estimate from the previous block $\cB_{i,j-1}$ (Algorithm~\ref{alg:endofblock}), epoch $\cE_i$ is ended and $\cE_{i+1}$ started. Figure~\ref{fig:algorithm_illustration} gives an illustration of epochs and blocks.

\begin{algorithm}[!hbtp]
\begin{minipage}{\linewidth}
\small
   \caption{\dynalg}
   \label{alg:DYN-LQR}
   \LinesNumbered
 \SetKw{KwDef}{Definition:} 
 \SetKw{KwInit}{Initialize:} 
\SetAlgoLined
   \KwIn{Horizon $T$, stabilizing controllers $\{K^\stab_t\}$, input instance parameters $\rho_0, \psi,\kappa,\beta$}
   \KwDef{} $\nu_0=1$; $\nu_j^2 = \sqrt{\frac{C_0}{2^j L}}$ for $j\geq 1$ where $C_0 = 4\log T$, $L = \frac{16 (n+d) \log^3 T}{1-\rho_0}$\;
    \hspace{0.2in}   $\mathcal{B}_{i,j} =[\tau_i + 2^{j-1} L, \tau_i + 2^{j} L - 1]$, where $\tau_i$ is the start of exploration epoch $\cE_i$\;
   \hspace{0.2in} Bounds on $\norm{x_t}$ for stabilization epochs: $x_u = 2\kappa e^{C_{ss}} \left( \frac{\sqrt{8(n+d)}\beta}{\sqrt{1-\rho_0}} \sqrt{\log T} + \frac{(n+d)B}{1-\rho_0} \right), x_\ell = \frac{2\psi \kappa \sqrt{n}}{1-\rho_0}$\;
    \hspace{0.2in}  $\eta_t \stackrel{i.i.d.}{\sim} \mathcal{N}(0,I_n)$\;
    \KwInit{} $t=1, i=1$\;
    $\tau_i \leftarrow t$ \tcc*[r]{Start of exploration epoch $\cE_i$}
    \nllabel{algline:epochstart} 
    \For(\tcc*[f]{Block $0$ (warm-up)}){$t=\tau_i, \ldots, \tau_i+L-1$}{
    Play $u_t = K^\stab_t x_t + \nu_0 \eta_t$\;
    }
    \For{$j=1, 2, \dots$}{
    Let $\hat{\Theta}_{i,j-1}$ be the OLS estimator based on $\cB_{i,j-1}$, and define $K_{i,j} = K^*(\hat{\Theta}_{i,j-1})$\;
 $\mathcal{M}  \leftarrow \emptyset$ \tcc*[r]{    Initialize the set of exploration phases }
    \While{$t\leq \tau_i + 2^{j} L - 1$}{
    $E\sim \textsf{Ber}\left(\frac{1}{L} 2^{-j/2} \sum_{m = 0}^{j-1} 2^{-m/2}\right)$ \tcc*[r]{Sample exploration indicator}
    \If{$E = 1$}{
    Sample exploration scale index $m\in\{0,1,2,\dots,j-1\}$ with probability $\text{Pr}(m = b) \propto 2^{-b/2}$\;
    $\mathcal{M}  \leftarrow \mathcal{M} \cup\{(m,t)\}$\;
    }
    Let $M_t = \{ (m,s)\in \cM \ |\ s \leq t \leq s+2^m L -1 \}$ \tcc*[r]{Active exploration phases}
    \eIf{$M_t  \neq  \emptyset$}{
    Set $m_t = \min\{ m \ |\ \exists (m, s) \in M_t\}$\;
    Play $u_t = K_{i,j}x_t + \nu_{m_t} \eta_t $\;
    }{
    Play $u_t = K_{i,j}x_t + \nu_{j} \eta_t$\;
    }
     Observe $x_{t+1}$\;
     \For{$(m,s) \in \cM \text{ with } t = s+2^m L-1$}{
     \If{$\textsc{EndOfExplorationTest}(i,j,m,s) = \textit{Fail}$}{
     $t \leftarrow t +1$, $i\leftarrow i+1$\ ; Go to line \ref{algline:epochstart} \tcc*[r]{Start a new epoch}
     \nllabel{algline:epochendnonstat1} 
     }
     }
      \If{$t = \tau_{i}+2^{j} L-1$ {\bfseries and} $\textsc{EndOfBlockTest}(i,j) = \textit{Fail}$ }{
      $t \leftarrow t +1$, $i\leftarrow i+1$\ ; Go to line \ref{algline:epochstart} \tcc*[r]{Start a new epoch}
           \nllabel{algline:epochendnonstat2} 
      }
      $t \leftarrow t +1$\;
       \If(\tcc*[f]{Instability detected}){$\norm{x_{t}} \geq x_u$}{
        \While{$\norm{x_{t}} \geq x_\ell$}{
         Play $u_t = K^\stab_t x_t$, observe $x_{t+1}$\;
         $t \leftarrow t+1$\;
        }
        $i \leftarrow i+1$\ ; Go to line \ref{algline:epochstart} \tcc*[r]{Start a new epoch}
             \nllabel{algline:epochendinstab} 
       }
    }
    }
\end{minipage}
\end{algorithm}

\begin{algorithm}[t]
\small
 \caption{\textsc{EndOfExplorationTest}($i,j,m,s$)} 
 \label{alg:endofphase}
\SetAlgoLined
 Construct OLS estimator $\hat{\Theta}_{i,j,(m,s)}$\;
 $\hat{\Theta}_{i,j,(m,s)} = \argmin_{\Theta} \sum_{t=s}^{s+2^m L -1} \norm{ x_{t+1} - \Theta [x_t^\top u_t^\top]^\top }_F^2$\;
    \If(\tcc*[f]{See  \eqref{eq:C_defs}}){$\norm{\hat{\Theta}_{i,j-1} - \hat{\Theta}_{i,j,(m,s)}}_F^2 \geq  (1 + \bar{C}_\bias + 2 \bar{C}_{\var})^2 (2^m L)^{-1/2}$}
    {
    Return \textit{Fail}\;
    }
    {
    Return \textit{Pass}\;
    }
\end{algorithm}

\begin{algorithm}[t]
\small
 \caption{\textsc{EndOfBlockTest}($i,j$)}
 \label{alg:endofblock}
\SetAlgoLined
 Construct OLS estimator $\hat{\Theta}_{i,j}$ \;
 $\hat{\Theta}_{i,j} = \argmin_{\Theta} \sum_{t \in \mathcal{B}_{i,j}} \norm{ x_{t+1} - \Theta [x_t^\top u_t^\top]^\top }_F^2$\;
    \eIf(\tcc*[f]{See \eqref{eq:C_defs}}){$\norm{\hat{\Theta}_{i,j-1} - \hat{\Theta}_{i,j}}_F^2 \geq  (1 + \bar{C}_\bias + 2 \bar{C}_{\var})^2(2^{j-1} L)^{-1/2}$}
    {
    Return \textit{Fail}\;
    }
    {
    Return \textit{Pass}\;
    }
\end{algorithm}

The vanilla policy mentioned above suffers from the problem that we could potentially commit to a controller for a long block -- and hence fail to detect a large change, which could in turn potentially lead to $\cO(T)$ regret. This is where the crucial novelty of the scheme of \cite{chen2019new} (designed for contextual multi-armed bandits) comes into play: to detect non-stationarity, which may happen at different scales (few large or many small changes), at each time within the block $\cB_{i,j}$, the authors' algorithm enters a \emph{replay phase} where the policy from an earlier block in the same epoch (together with the larger exploration noise) is played. If at the end of some replay phase, the estimate of reward differs significantly from the history, the current epoch is ended. The algorithm could potentially be in multiple replay phases simultaneously, in which case the policy to replay is picked uniformly at random from active replays. Replay phases with different indexes are intended to detect changes of different magnitudes. 

To adapt to the LQR setting, we simplify the above strategy. In particular, at any time $t$ in a block $\cB_{i,j}$, we enter an \textit{exploration phase} with probability proportional to $1/\sqrt{|\cB_{i,j}|}$ and given this event happens, the `scale' of the exploration phase is chosen to be $m$ with probability proportional to $1/\sqrt{2^m}$. A scale $m$ exploration phase lasts for $2^m L$ time steps, during which we play the action $u_t = K_{i,j} x_t + \sigma_t \eta_t$. That is, we keep playing the same linear feedback controller, but with exploration noise increased to $\sigma^2_t \approx \frac{1}{\sqrt{2^m}}$. Therefore, a scale $m$ exploration phase allows us to detect variation in $\Theta_t$ of size $\sqrt[\leftroot{-3}\uproot{3}4]{1/2^m}$. There can be multiple exploration phases active at any time $t$. We denote them by $M_t = \{(m_1, t_1),(m_2,t_2), \ldots\}$ where $m_k$ denotes the scale and $t_k$ denotes the starting time of the $k$-th active exploration phase. In this case, we play the most aggressive (i.e., the smallest $m$) exploration phase, with the feedback used by all active exploration phases to improve their estimates. At the end of the exploration phase $(m,s)$, we first compute the OLS estimator $\Theta_{i,j,(m,s)}$, and declare non-stationarity and end the epoch if $\norm{\hat{\Theta}_{i,j-1} - \hat{\Theta}_{i,j,(m,s)}}^2 \gtrapprox \frac{1}{\sqrt{2^{m}}}$ (Algorithm~\ref{alg:endofphase}).  


One crucial difference between LQR and the contextual bandit setting off \cite{chen2019new} is that LQR has a quadratic cost, while contextual bandit is a special case of a linear bandit problem, which affects the choice of $\sigma_t$. Yet another crucial difference from the contextual bandit setting is that since the LQR system has a state, the system could potentially become unstable through an inaccurate estimate before the non-stationarity is detected. We thus create a third criterion for ending an epoch: whenever $\norm{x_t} \geq x_u = \cO\left( \frac{\sqrt{(n+d)\log T}}{1-\rho_0}\right)$, we end the current epoch and enter a \textit{stabilization epoch}. In a stabilization epoch we keep playing the stabilizing controllers without any exploration noise until $\norm{x_t}$ drops below $x_\ell = \cO\left( \frac{n}{1-\rho_0}\right)$. At this point, we begin a regular \textit{exploration epoch}.


\section{Estimation error for OLS with non-stationary $\Theta_t$}
\label{sec:ols_esti}

A central ingredient of our algorithm is the ordinary least squares estimator used to learn the approximate dynamics. While the study of the variance of the OLS estimator is a well-understood topic, when the parameter sequence is non-stationary, the OLS estimator can be biased. Studying this bias is quite non-trivial, especially for the LQR problem.

We state our results on the estimation error of the OLS estimator for non-stationary LQR at the end of this section and devote Appendix~\ref{sec:ols_est_proof} to the formal proofs of the results. However, we will highlight in brief the reason that these results are challenging and non-trivial. For intuition, the reader should keep the trade-off we pointed to at the end of Section~\ref{sec:stationary} in mind: during an interval $\cI$ of length $|\cI|$, to balance the exploration-exploitation trade-off we would like to create an estimator that has error of order $|\cI|^{-1/4}$. With a non-stationary parameter sequence, this error comes from both the variance of the estimator as well as the bias. Therefore, if the variation in $\Theta_t$ during this interval, $\Delta_{\cI}$, is of smaller order than $|\cI|^{-1/4}$, then we would like the bias of our estimator to be $\cO(\Delta_{\cI})$. 

\paragraph{Failure of a naive proof-strategy.} We first show that an obvious first line of attack to bound the estimation error of OLS does not work. Define $z_t = [x_t^\top, u_t^\top]^\top$ and $\Upsilon_\cI \coloneqq \sum_{t\in\cI} z_t z_t^\top$ for an interval $\cI=[s,e]$. Then we can write the error in the OLS estimator compared to a `representative' $\bar{\Theta}$ (e.g., $\bar{\Theta} = \Theta_e$) as: 
\*
\hat{\Theta}_{\cI}-\bar{\Theta} = \underbrace{\left(\sum_{t\in \cI}\left(\Theta_{t}-   \bar{\Theta}\right) z_{t} z_{t}^{\top}\right) \Upsilon_{\cI}^{-1}}_{\footnotesize \mbox{``bias''}}+ \underbrace{\left(\sum_{t\in \cI } w_{t} z_{t}^{\top}\right)\Upsilon_{\cI}^{-1}}_{\footnotesize \mbox{``variance''}}.
\*
The above shows that if $\Theta_t$ is constant in $\cI$, then the estimator is unbiased. Lacking that, we may try to bound the first term as follows (this proof strategy was followed in \cite{cheung2019learning}). Let $\bar{\Theta} = \Theta_e$, then
\*
& \norm{ \left(\sum_{t\in \cI}\left(\Theta_{t}-  \Theta_e\right) z_{t} z_{t}^{\top}\right)\Upsilon_{\cI}^{-1}}_F
= \norm{ \left(\sum_{t\in \cI}\sum_{p = t}^{e-1}\left(\Theta_{p}-   \Theta_{p+1}\right)  z_{t} z_{t}^{\top}\right)\Upsilon_{\cI}^{-1}}_F \\
 & \qquad \qquad = \ \norm{ \sum_{p=s}^{e-1} \left(\Theta_{p}-   \Theta_{p+1}\right) \left(\sum_{t = s}^{p-1}z_{s} z_{s}^{\top}\right)\Upsilon_{\cI}^{-1}}_F 
 \  \leq \  \sum_{p=s}^{e-1}  \norm{ \left(\Theta_{p}-   \Theta_{p+1}\right) }_F \lambda_{\max}\left(\left(\sum_{t = s}^{p-1}z_{s} z_{s}^{\top}\right)\Upsilon_{\cI}^{-1}\right).
\*
If $\lambda_{\max}\left(\left(\sum_{t = s}^{p-1}z_{s} z_{s}^{\top}\right)\Upsilon_{\cI}^{-1}\right) \leq 1$, then the analysis above would bound the bias by $\Delta_{\cI}$. While this may seem intuitive (e.g., it is true if $z_s$ are scalars), this was shown to be false for an arbitrary $\{z_s\}$ sequence even for the case of $z_s \in \RR^2$ by \cite{zhao2021non}.

\begin{figure}[t]
    \centering
    \includegraphics[width=1.9in]{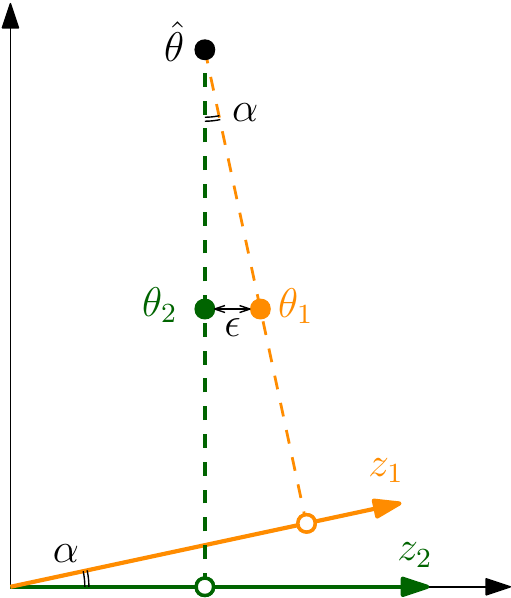}
    \caption{Illustration for the large bias of OLS estimator with non-stationary parameters.} 
    \label{fig:OLS_illustration}
\end{figure}

\paragraph{An illustrative example.} To further highlight why a technically challenging analysis is necessary for the study of OLS with a non-stationary parameter sequence, we consider a simple example of OLS estimation \textit{without noise}. Consider a 2-dimensional example with two data points: 
\begin{align*}
    \theta_1 \  =\  [1\ \  1], \quad \theta_2 \ =\ [1-\epsilon \ \ 1]; \qquad   z_1 \ =\ [\cos \alpha \ \ \sin \alpha], \quad  z_2 \ =\  [1 \ \ 0].
\end{align*} 
Figure~\ref{fig:OLS_illustration} shows the geometric intuition behind the OLS estimator. In this specific example, the estimate is given by the intersection of two (for $t=1,2$) lines: perpendicular to $z_t$ and passing through $\theta_t$. The bias of the OLS estimate $\hat{\theta}$ in this noiseless case is given as $|\hat{\theta} - \theta_2| = \epsilon/\tan \alpha$. With $\alpha \approx \epsilon \ll 1$, the bias approaches $1$ even though $\theta_1,\theta_2$ are $\epsilon$-close to each other. The matrix $\Upsilon$ for this case is 
\[ 
\Upsilon = \begin{bmatrix}
1+\cos^2 \alpha & \cos \alpha \cdot \sin \alpha \\
\cos \alpha \cdot \sin \alpha & \sin^2 \alpha
\end{bmatrix},
\]
which is ill-conditioned when $\alpha \ll 1$. In particular, $\lambda_{\max}(\Upsilon)/\lambda_{\min}(\Upsilon) \approx 1/2\alpha^2$. It might seem that such an ill-conditioned $\Upsilon$ is an extreme case that is unlikely to bother our study. However, with the exploration noise chosen in Algorithm~\ref{alg:DYN-LQR}, we give evidence in Lemma~\ref{lemma:bounded_coundition_number} that the condition number of $\Upsilon_{\cI}$ for intervals $\cI$ of interest is concentrated around $\cO(\sqrt{|\cI|})$, while we are trying to get unbiased estimates when the variation of $\Theta_t$ in interval $\cI$ is $\Delta_{\cI} = \cO(|\cI|^{-1/4})$. This precisely corresponds to the problematic setting $\alpha \sim \epsilon \ll 1$ in our toy example above. 

\paragraph{Our proof approach.} We begin by decomposing the problem into bounding the estimation error for each row of the estimate $\hat{\Theta}_{\cI}$. For a given row, $\hat{\theta}_{\cI}$, the key obstacle in the analysis of the estimation error $\norm{\hat{\theta}_{\cI} - \bar{\theta}}^2$ is that while $z_t$ lives in $\RR^{n+d}$, most of its variance is in the $n$-dimensional column space of $[I_n \ K_\cI^\top]^\top$, where $K_\cI$ is the fixed linear feedback controller used during interval $\cI$. This is because the LQR dynamics naturally adds the noise $w_{t-1}$ to arrive at the state $x_t$ allowing efficient exploration/estimation of the component of $\bar{\theta}$ lying in the column space of $[I_n \ K_\cI^\top]^\top$. In particular, the total energy in this column space is $\cO(|\cI|)$ through $w_t$, while the energy in the orthogonal subspace through the exploration noise $\xi_t = \sigma_t \eta_t$ is $\sum_{t\in \cI} \sigma^2_t = \cO(\sqrt{|\cI|})$. Therefore, as our toy example points out, a naive analysis based on a lower bound on the eigenvalues of the matrix $\sum_{t \in \cI} z_t z_t^\top$ fails, because it does not exploit the statistical independence between $\xi_t$ and $x_t$. 

Our approach is to instead to look at one-dimensional OLS problems parameterized by directions $v \in \SSS^{n+d} := \{ v \in \RR^{n+d}, \norm{v}=1\}$:
\[ 
\lambda_v = \argmin_{\lambda} \cL(\bar{\theta} + \lambda \cdot v), 
\]
where $\cL$ is the quadratic loss function for OLS. We argue that $|\lambda_v|$ are small for `enough' directions $v$. That is, in enough directions, the minimizer $\left( \bar{\theta}+\lambda_v \cdot v \right)$ of the 1-dimensional quadratic defined above is close to the candidate $\bar{\theta}$. Furthermore, since the loss function looks very different for $v$ lying close to the column space of $[I_n \ K^\top]^\top$ versus $v$ lying close to its orthogonal subspace, we consider two cases: $v$ lying only in the column space or lying only in its orthogonal subspace, and prove that the geometry of Hessian implies that it is sufficient to look at these two cases. The complete proof is presented in Appendix~\ref{sec:ols_est_proof}.

\paragraph{Results} We state our lemmas for the estimation error for the OLS estimators used in Algorithm~\ref{alg:DYN-LQR}. Lemma~\ref{lemma:ols_esti_new} states it for intervals within exploration blocks $\cB_{i,j}$, while Lemma~\ref{lemma:ols_esti_warmup} states it for warm-up blocks $\cB_{i,0}$. The reason for the two separate results is that within a warm-up block, the controller $K_t^\stab$ is changing, which does not allow a subspace decomposition we mentioned earlier, but the $\cO(1)$ exploration noise still allows us to bound the estimation error. Within an exploration block $\cB_{i,j}$, the exploration noise is of a much smaller magnitude (to control regret due to exploration), but the controller $K_t$ is fixed, which allows the decomposition. 

\begin{lemma}\label{lemma:ols_esti_new} 
Consider an interval $\cI$ in block $\cB_{i,j}$ for some epoch $\cE_i$ in Algorithm~\ref{alg:DYN-LQR}, such that $|\cI| \geq L $ and $\max_{t\in \cI} \norm{x_t} \leq x_u$. Let $\hat{\Theta}_{\cI}$ be the corresponding OLS estimate from observations in $\cI$ and $\bar{\Theta} = \Theta_t$ for some $t \in \cI$. Then, there exists a $T_0$, such that for $T \geq T_0$, with probability at least $1-\varepsilon$:
\*
\norm{\hat{\Theta}_{\cI} - \bar{\Theta}}_F 
& \leq \breve{C}_1 \Delta_\cI + \breve{C}_2 |\cI|^{- \frac{1}{4}},
\*
where
\* 
\breve{C}_1 =  C_{\bias} \sqrt{\ln \frac{1}{\varepsilon} + \ln T} \quad \mbox{and} \qquad
\breve{C}_2 =  C_{\var} \left( \sqrt{\ln T} + \sqrt{\ln \frac{1}{\varepsilon}}\right),
\*
for problem dependent constants $C_{\bias},C_{\var}$ (precise expressions are shown in \eqref{eq:def_breve_C}).
\end{lemma}

\begin{lemma}
\label{lemma:ols_esti_warmup}
Consider a warm-up block $\cB_{i,0}$ in Algorithm~\ref{alg:DYN-LQR} and let $\bar{\Theta}= \Theta_t$ for some $t\in \cB_{i,0}$. There exists a $T_0$, such that for $T \geq T_0$ and the choice of $L$ in Algorithm~\ref{alg:DYN-LQR}, the OLS estimate  $\hat{\Theta}_{\mathcal{B}_{i, 0}}$ of a warm-up block $\cB_{i,0}$ 
satisfies 
\*
\norm{\hat{\Theta}_{\mathcal{B}_{i, 0}} - \bar{\Theta}}_F 
& \leq \breve{C}_{1,\stab} \Delta_{\mathcal{B}_{i, 0}} + \breve{C}_{2,\stab}|\mathcal{B}_{i, 0}|^{- \frac{1}{4}},
\*
with probability at least $1-\varepsilon$ where 
\*
\breve{C}_{1,\stab} = C_{\bias, \stab} \sqrt{\ln T} \quad \mbox{and} \qquad
\breve{C}_{2,\stab} = C_{\var,\stab} \sqrt{ \ln \frac{1}{\varepsilon} + \ln \ln T},
\*
for problem dependent constants $C_{\bias,\stab},C_{\var,\stab}$ (precise expressions are shown in \eqref{eq:def_breve_C_stab}).
\end{lemma}
Applying Lemma~\ref{lemma:ols_esti_new} and Lemma~\ref{lemma:ols_esti_warmup} with $\varepsilon = 1/T^3$ to all the intervals (at most $T^2$) that may be considered during the execution of Algorithm~\ref{alg:DYN-LQR} and a union bound immediately gives the following result.

\begin{lemma} 
\label{lemma:uniform_ols_concentration}
Define \textsc{Event 1} as the event that for each warm-up block $\cB_{i,0}$ in Algorithm \ref{alg:DYN-LQR} it holds that
\*
\norm{\hat{\Theta}_{\mathcal{B}_{i, 0}} - \bar{\Theta}}_F 
& \leq {C}_{\bias,\stab} \sqrt{\ln T} \Delta_{\mathcal{B}_{i, 0}} + 3 {C}_{\var,\stab} \sqrt{\ln T} |\mathcal{B}_{i, 0}|^{- \frac{1}{4}},
\*
and for each phase and non-warmup block, denoted by $\cI=[s,e]$, it holds that 
\*
\norm{\hat{\Theta}_{\cI} - \bar{\Theta}}_F \leq 3  C_{\bias} \sqrt{\ln T} \Delta_\cI + 3 C_{\var} \sqrt{\ln T} |\cI|^{-\frac{1}{4}}.
\*
Then we have that $\prob{\textsc{Event 1}} \geq 1-1/T$. 
\end{lemma}
For succinctness, define 
\#
\label{eq:C_defs}
\bar{C}_{\bias} = \sqrt{\ln T} \max\{ 3 C_{\bias} , C_{\bias, \stab}\} \quad \mbox{and} \quad  \bar{C}_{\var} = \sqrt{\ln T} \max\{ 3 C_{\var}, 3 C_{\var,\stab}\} .
\#

\section{Regret Upper Bound for \dynalg}
\label{sec:analysis}

Our main regret upper bound for \dynalg \ is shown below.
\begin{theorem}
\label{thm:dynalg_regret}
Under Assumption~\ref{assum:stabilization}, the expected regret of \dynalg \ is upper bounded as:
\[ \expct{\Reg^{\dynalg}(T)} = \tilde{\cO}\left( V_T^{2/5} T^{3/5} \right).\]
If the dynamics $\{ \Theta_t\}$ are piecewise constant with at most $S$ switches, then the regret of \dynalg \ is upper bounded as:
\[ \expct{\Reg^{\dynalg}(T)} = \tilde{\cO}\left( \sqrt{ST} \right).
\]
\end{theorem}

Our definition of $\Reg^{\dynalg}(T)$ in \eqref{eqn:regret_defn} measures the regret relative to the benchmark $\sum_{t=1}^T J^*_t$. In the next proposition,  we prove that this benchmark is at most  $\tilde{\cO}(V_T)$ larger than the expected cost of the dynamic optimal policy. This additive error is dominated by the regret $\tilde{\cO}(V_T^{2/5}T^{3/5})$ proved in Theorem~\ref{thm:dynalg_regret}. Proposition~\ref{prop:dynopt_regret} is proved in Appendix~\ref{sec:proof_dynopt_regret}.

\begin{proposition}
\label{prop:dynopt_regret}
 Let $\{u_t\}_{t=1}^T$ be an arbitrary non-anticipative policy for  the non-stationary LQR control problem. Then,
 \[ 
 \expct{ \sum_{t=1}^T x_t^\top Q x_t + u_t^\top R u_t  }  \geq \sum_{t=1}^{T} J^*_t - {\cO}(V_T+ \log T).
 \]
\end{proposition}

We will conduct our analysis under the assumption that \textsc{Event 1} specified in Lemma~\ref{lemma:uniform_ols_concentration} occurs. Since \dynalg \ uses $K_t^{\stab}$ whenever $\norm{x_t} \geq x_u$, outside this event, the total cost is bounded by $\tilde{\cO}(T)$. Note that this happens with probability at most $1/T$.

\subsection{Regret Decomposition}
\label{sec:decomp}
We begin with an informal regret decomposition lemma which highlights the key exploration-exploitation trade-off for non-stationary LQR. 

\textbf{Informal Lemma.}
The expected regret for a policy $\pi$ with $u_t = K_t x_t + \sigma_t \eta_t$ where $K_t, \sigma_t$ are adapted to the filtration $(\cF,\cG)$ is given by:
\begin{align}
\nonumber
\expct{\Reg^{\pi}(T)} & =  \expct{ \sum_{t=1}^T x_t^\top Q x_t + u_t^\top R u_t - J^*_t } \\
\nonumber =& \ \underbrace{\sum_{t=1}^T \expct{ J_t(K_t) - J^*_t }}_{\mbox{\footnotesize{exploitation regret}}}
+ \underbrace{\sum_{t=1}^T \expct{ \sigma_t^2 \Tr\left( R +  B_t^\top P_t(K_t) B_t \right) }}_{\mbox{\footnotesize exploration regret}} \\
& \quad + \underbrace{\sum_{t=1}^{T-1} \expct{x_{t+1}^\top \left( P_{t+1}(K_{t+1}) - P_{t}(K_t) \right) x_{t+1}}}_{\mbox{\footnotesize policy/parameter variation}} 
+ \expct{x_1^\top P_1(K_1) x_1 - x_{T+1}^\top P_T(K_T)x_{T+1}}.
\label{eqn:regret_decomp}
\end{align}

We term the lemma informal because it relies on $J_t(K_t)$ and $P_t(K_t)$ being defined for all $t$. This need not always be true for $\dynalg$  since $K_t$ is the certainty equivalent controller based on an estimate of $\Theta_t$, and therefore the {\it stationary} system corresponding to $\Theta_t$ and $K_t$ need not even be stable, and $J_t(K_t)$ could be unbounded. We shortly address how we handle such time periods, but their contribution to regret will be asymptotically of a smaller order. The decomposition points out that the dominant terms in the analysis will be the exploitation regret and the exploration regret. The policy/parameter variation depends on how much the pair $(\Theta_t, K_t)$ changes during non-warmup blocks of an exploration epoch. By design, the policies $\{K_t\}$ are piece-wise constant with at most $\log T$ changes per epoch, and we will prove that the number of epochs is $\cO(V_T^{4/5}T^{1/5})$. Finally, for a fixed $K$, $\norm{P(\Theta_t, K) - P(\Theta_{t+1},K)} = \cO(\norm{\Theta_t - \Theta_{t+1}})$, and hence this contributes at most $\cO(V_T)$ to the regret across the entire horizon.

To refine the regret decomposition, we recapitulate Algorithm~\ref{alg:DYN-LQR}, and in particular the classification of \textit{exploration epochs}, \textit{stabilization epochs}, \textit{blocks} within exploration epochs, and another concept we define for the purpose of analysis alone -- \textit{bad intervals}.

\textit{(i) Stabilization epochs} -- such epochs begin whenever $\norm{x_t}$ exceeds the upper bound $x_u$, indicating the potential instability of the current controller. We use $\tau^\stab_i$ to denote the start of the $i$-th stabilization epoch. During a stabilization epoch, we use the controller $K_t = K^\stab_t$. The $i$-th stabilization phase ends at $\theta_i^\stab$ (inclusive) where 
\[ 
\theta^\stab_i = \min\{ t \geq \tau^\stab_i +1 \ : \ \norm{x_{t+1}} \leq x_\ell  \}. 
\]
We use $\cS_i$ to denote the interval $[\tau^\stab_i, \theta^\stab_i]$ as well as the $i$-th stabilization epoch symbolically.
    
\textit{(ii)  Exploration epochs} -- such epochs begin either at the end of a stabilization epoch, or at the end of another exploration epoch if sufficient non-stationarity is detected through failure of \textsc{EndOfExplorationTest} or \textsc{EndOfBlockTest}. We will denote the start and end of the $i$-th exploration epoch by $\tau_i$ and $\theta_i$ respectively, and use $\cE_i$ to denote the interval $[\tau_i, \theta_i]$ as well as the epoch symbolically.
    
\textit{(iii) Blocks} -- The $i$-th exploration epoch $\cE_i$ is partitioned into non-overlapping blocks of geometrically increasing duration. Block 0 (also called the \textit{warm-up block}) is the interval $[\tau_i, \tau_i + L -1]$, and the $j$-th block ($j=1,\ldots$) is the interval $[\tau_i +L \cdot 2^{j-1} , \tau_i + L\cdot 2^j - 1 ] \cap \cE_i$ of maximum length $L\cdot 2^{j-1}$. We denote by $\cB_{i,j}$ both the interval as well as the block symbolically. 
The controller used at time $t \in \cB_{i,j}$ ($ j\geq 1$) is given by $K^*(\hat{\Theta}_{i,j-1})$, where $\hat{\Theta}_{i,j-1}$ is the OLS estimator based on the  block $j-1$ of epoch $\cE_i$. For succinctness, we use the notation 
\[ 
\hat{\Theta}_t := \hat{\Theta}_{i,j-1}, \quad \mbox{for } t \in \cB_{i,j}. 
\]
We will use $B_i$ to denote the number of blocks in epoch $\cE_i$.

\textit{(iv) Bad/Good intervals} -- 
It can happen that for some time steps during an exploration epoch, the controller is unstable and therefore $J_t(K_t)$ is undefined, but the $\norm{x_t}$ has not exceeded $x_u$. 
To study the regret due to such $t$, we define the notion of bad intervals within epochs. The $k$-th bad interval of an epoch $i$ begins at $\tau_{i,k}^\bad$ and ends at $\theta^\bad_{i,k}$ where these are defined recursively as:
\begin{align*}
        \tau^\bad_{i,1} &:= \min \left\{ t \in [\tau_i+ L,\theta_i] : \norm{ \hat{\Theta}_t - \Theta_t}_F^2 \geq C_3 \right\},\\
        \tau^\bad_{i,k} &:= \min \left\{ t \in [\theta^{\bad}_{i,k-1},\theta_i]  : \norm{ \hat{\Theta}_t - \Theta_t}_F^2 \geq C_3 \right\}, \\  \theta^\bad_{i,k} & :=  \min \left\{ t \in [\tau^\bad_{i,k}+1, \theta_i] \ : \  \norm{ \hat{\Theta}_{t+1} - \Theta_{t+1}}_F^2 \leq \frac{C_3}{2} \right\},     
\end{align*}
with the constant $C_3$ defined in Assumption~\ref{assum:approx_stabilization}. Note that we do not create bad intervals during the block $\cB_{i,0}$, which is analyzed separately. We denote the $k$-th bad interval of an epoch $i$ as $\cI^{\bad}_{i,k}$. By $\cI^\bad_i$, we denote the union of all bad intervals in $\cE_i$, and by $\cI^\bad$, the union of all bad intervals. All time periods that not in bad intervals, i.e., they are in $\cE_i \setminus \{ \cB_{i,0} \cup \cI^\bad_i \}$, will be called \textit{good} and split into \textit{good intervals}. For analysis purposes, we further split the good time periods based on the blocks. That is, a good interval can end at time $t$ if \textit{(i)} either a bad interval begins at time $t+1$, or \textit{(ii)} a block ends at time $t$ in which case another good interval can begin at time $t+1$. Using a similar notation $\cI^\good_{i,j,k}$ denotes the $k$-th good interval of a block $\cB_{i,j}$ (which must lie entirely inside $[\tau_i + L\cdot 2^{j-1} , \tau_{i}+L \cdot2^j - 1 ]$. We will use $N^\bad_i$ to denote the total number of bad intervals in an epoch $i$ and $N^\good_{i,j}$ to denote the number of good intervals in a block $\cB_{i,j}$. The advantage of defining the good intervals to lie within a block is that for the purposes of analysis, the good intervals within a block $\cB_{i,j}$ are completely defined based on history before the start of the block $\cB_{i,j}$.

We will use $E$ to denote the total number of exploration epochs and $E_S$ to denote the total number of stabilization epochs. Finally, we come to the regret decomposition that we use in the subsequent section:
\begin{align}
\nonumber    \Reg^\dynalg(T) &= \sum_{t=1}^T c_t - J^*_t  \\
\label{eqn:regret_decomposition_T}
    &\leq \underbrace{\sum_{i=1}^{E_S} \sum_{t \in \cS_i} c_t}_{\substack{\text{$T_1$: Stabilization}\\ \text{epochs}}}  + \underbrace{\sum_{i=1}^E \sum_{t \in \cB_{i,0}} c_t}_{\substack{\text{$T_2$: Warm-up}\\ \text{blocks}}} +
    \underbrace{\sum_{i=1}^E \sum_{k=1}^{N^\bad_i} \sum_{t \in \cI^\bad_{i,k}} c_t}_{\substack{\text{$T_3$: Bad}\\ \text{intervals}}}  + \underbrace{\sum_{i=1}^E 
    \sum_{j=1}^{B_i} \sum_{k=1}^{N^\good_{i,j}} \sum_{t \in \cI^\good_{i,j,k}}\left( c_t - J_t^* \right)}_{\substack{\text{$T_4$: Good}\\ \text{intervals}}}.  
\end{align}

\subsection{Regret analysis for \dynalg}
\label{sec:reg_decomp}

The main result of this section is the following lemma, which provides an intermediate characterization of $\expct{\Reg^\dynalg(T)}$ based on \eqref{eqn:regret_decomposition_T}. In particular, the characterization highlights that to bound the regret, it is sufficient to bound \textit{(i)} the number $E$ of exploration epochs (Section~\ref{sec:bound_epochs}) and \textit{(ii)} the total squared norm of the estimation error of dynamics $\Theta_t$ for the good periods (Section~\ref{sec:upper}).

\begin{lemma}
\label{lem:regret_est_error}
The expected regret for \dynalg \ is bounded as follows:
\begin{align}
\label{eqn:regret_est_error}
\nonumber
    \expct{\Reg^\dynalg(T)} & \leq \tilde{\cO}\left( \expct{ \sum_{i=1}^E \sum_{j=1}^{B_i} \sum_{t \in \cB_{i,j}}  \min\left\{  \norm{\hat{\Theta}_{i,j-1} - \Theta_t}_F^2 , C_3 \right\}  + \sqrt{|\cB_{i,j}|} } \right) + \tilde{\cO}(E + V_T), \\
    & \leq \tilde{\cO}\left( \expct{ \sum_{i=1}^E \sum_{j=1}^{B_i} \sum_{t \in \cB_{i,j}}  \min\left\{  \norm{\hat{\Theta}_{i,j-1} - \Theta_t}_F^2 , C_3 \right\}  } \right) + \tilde{\cO}(\sqrt{E\cdot T} + V_T).
\end{align}
\end{lemma}

\begin{proof} We proceed by bounding the terms  in \eqref{eqn:regret_decomposition_T}.\\
\textbf{Upper bound for Term 1.} Since the controllers $\{K^\stab_t\}$ used in a stabilization epoch satisfy sequentially strong stability (Assumption~\ref{assum:stabilization}), in Lemma~\ref{lem:stab_epoch} we prove that the expected total cost per stabilization epoch is $\tilde{\cO}(1)$. Since the number of stabilization epochs is bounded by the number of exploration epochs $E$, the total contribution of Term 1 in \eqref{eqn:regret_decomposition_T} is $\tilde{\cO}(E)$.

\begin{lemma}
\label{lem:stab_epoch}
Let $[\tau^\stab, \theta^\stab]$ be a stabilization epoch. The expected total cost during the stabilization epoch is bounded by 
\[  
\expct{ \left. \sum_{t=\tau^\stab}^{\theta^\stab} c_t \right| \cF_{\tau^\stab-1}, \cG_{\tau^\stab-1} } = \cO \left( \frac{\kappa^2 x_u^2}{1-\rho_0} \right) = {\cO}\left( \frac{ \kappa^2 \beta^2 (n+d + \log T)}{(1-\rho_0)^2} \right) .
\]
\end{lemma}

\textbf{Upper bound for Term 2.} Similar to Lemma~\ref{lem:stab_epoch}, the use of $K^\stab_t$ during warm-up blocks gives a bound of $\tilde{\cO}(1)$ per epoch in Lemma~\ref{lem:warmup}, which gives a $\tilde{\cO}(E)$ contribution due to Term 2.
\begin{lemma}
\label{lem:warmup}
Let $[\tau_i, \tau_i + L - 1]$ denote the warm-up block $\cB_{i,0}$ of an exploration epoch $\cE_i$. The expected total cost during $\cB_{i,0}$, for any $i$, is bounded by 
\[ 
\expct{ \left. \sum_{t=\tau_i}^{\tau_i+L-1} c_t \right| \cF_{\tau_i-1}, \cG_{\tau_i-1} } = {\cO}\left( \frac{\kappa^2 \beta^2(n+d) \log^3 T}{(1-\rho_0)^2}  \right).
\]
\end{lemma}

\textbf{Upper bound for Term 3.}  Since $\norm{x_t}$ is bounded by $x_u = \tilde{\cO}(1)$ for any time period in a bad interval by definition,  the cost is bounded by $\tilde{\cO}(1)$ per time step.  We can bound the number of bad time periods within an  arbitrary interval $\cI$ noting that for $t \in \cI^\bad$,  $\norm{\hat{\Theta}_t - \Theta_t}_F^2 \geq C_3/2$ and thus:
\begin{align}
\label{eqn:size_bad_int}
    \size{\cI^\bad \cap \cI } & = \sum_{t \in \cI^\bad \cap \cI} 1 \leq \frac{2}{C_3} \sum_{t \in \cI} \min\left\{ \norm{ \hat{\Theta}_t - {\Theta}_t}_F^2 , C_3 \right\}.
    \end{align} 
Then the total contribution of Term 3 is ${\cO}\left( \expct{ \sum_{i=1}^E \sum_{j=1}^{B_i} \sum_{t \in \cB_{i,j}}  \min\left\{  \norm{\hat{\Theta}_{i,j-1} - \Theta_t}_F^2 , C_3\right\}}   \right)$.

\textbf{Upper bound for Term 4.} 
\begin{lemma}
\label{lem:expected_regret_goodint}
For some epoch $\cE_i$, a block $\cB_{i,j}$ in epoch $\cE_i$, and a good interval $\cI^\good_{i,j,k} = [\tau, \theta]$ in block $\cB_{i,j}$, the expected regret is bounded as follows:
\begin{align*}
    &\expct{\Reg^\pi(\cI^\good_{i,j,k}) \mid \cF_{\tau-1}, \cG_{\tau-1}} \\
    &\qquad\leq   \sum_{t =\tau}^\theta \left( J_t(K_t) - J^*_t  \right) + \size{\cI^\good_{i,j,k}} \frac{C_0^{1/2}}{L^{3/2}}\cdot \frac{j}{\sqrt{2^j}}  + {\cO}\left( \frac{n+d+\log T}{1-\rho_0} \left(1 + \Delta_{\cI_{i,j,k}} \right) \right) \\
    &\qquad \leq \sum_{t =\tau}^\theta C_4 \norm{ \hat{\Theta}_{i,j-1} - \Theta_t}_F^2 + \size{\cI^\good_{i,j,k}} \frac{C_0^{1/2} C_7}{L^{3/2}}\cdot \frac{j}{\sqrt{2^j}}  + {\cO}\left( \frac{n+d+\log T}{1-\rho_0} \left(1 + \Delta_{\cI_{i,j,k}} \right) \right),
\end{align*}
where the constant $C_7 := \max_{t} \sup\left\{ \Tr\left( R + B_t^\top P_t(K_t) B_t \right) \mid K_t = K^*(\hat{\Theta}), \norm{\hat{\Theta} - \Theta_t}_F^2 \leq C_3 \right\}$.
\end{lemma}

Combining the results above, we can bound the first term in \eqref{eqn:regret_est_error} immediately from Term 3 and the first summand in Lemma~\ref{lem:expected_regret_goodint} for Term 4. Summing the second term in Lemma~\ref{lem:expected_regret_goodint} over all the good intervals within a block $\cB_{i,j}$ (which is of length at most $2^j$) contributes $\tilde{\cO}(\sqrt{|\cB_{i,j}|})$. Since the blocks within an epoch are doubling in length, $\sum_{j} \sqrt{|\cB_{i,j}|} \leq 8 \sqrt{|\cE_i|}$ and $\sum_{i}\sqrt{|\cE_i|} \leq E \sqrt{T/E} = \sqrt{E \cdot T}$. The contribution of the third term in Lemma~\ref{lem:expected_regret_goodint} is proportional to the number of good intervals, which is bounded by $V_t/\sqrt{C_3/2}+E \log T$. To see this, note that without any bad intervals, there would be one good interval per block and there are at most $\log T$ blocks per epoch. For a good interval to begin due to a bad interval ending, the bad interval must `eat up' $\sqrt{C_3/2}$ of the variation due to the criterion chosen for the end of a bad interval. Hence, there can be at most $V_T/\sqrt{C_3/2}$ good intervals created because of the bad intervals.  The last term in Lemma~\ref{lem:expected_regret_goodint} contributes $\tilde{\cO}(V_T)$ to \eqref{eqn:regret_est_error}.
\end{proof}

\subsection{Bounding the Number of Epochs}
\label{sec:bound_epochs}
There are two ways of generating epochs in Algorithm~\ref{alg:DYN-LQR}: 
(1) epochs end due to the detection of non-stationarity
(lines \ref{algline:epochendnonstat1} and \ref{algline:epochendnonstat2}), 
and (2) epochs end due to the detection of instability (line \ref{algline:epochendinstab}).
This section is devoted to bounding the number of epochs from these two sources separately. 

\paragraph{Bounding the number of epochs generated by non-stationarity tests. }
In the subsequent analysis, we will bound the number of 
epochs terminated due to the detection of non-stationarity 
in $\Theta_t$ by $\cO(T^{1/5}V_T^{4/5})$,
which dominates $\cO(V_T)$. Recall that an epoch ends if the non-stationarity
tests in Algorithms~\ref{alg:endofphase} or \ref{alg:endofblock} fail,
which happens if the distance between the new OLS estimate and
the estimate based on the previous block exceeds some threshold. 
The thresholds there are carefully designed according to the concentration
results proved in Section~\ref{sec:ols_esti}, which allow us to prove 
the following lemma characterizing the variation budget needed
for  an epoch to fail the tests in Algorithms \ref{alg:endofphase} and  \ref{alg:endofblock}. 

\begin{lemma}\label{lemma:epoch_bound1}
Assume \textsc{Event 1} holds. Let $\cE_i$ be an epoch with total variation $\Delta_{[\tau_i,t]}\leq (t - \tau_i + 1 )^{-1/4}$, then the epoch does not end because of nonstationarity detection.
\end{lemma}
The following corollary bounds the number of restarts due to detection of non-stationarity.
\begin{corollary}
\label{coro:epoch_bound1}
Assume \textsc{Event 1} holds. The number of epochs that end due to detection of non-stationarity is bounded by $\cO(C_0^{-2/5}T^{1/5} V_T^{4/5})$.
\end{corollary}

\paragraph{Bounding the number of epochs generated by instability tests. }
Lemma~\ref{lem:instab} characterizes the variation budget needed to trigger the end of an epoch due to instability detection, which leads to Corollary~\ref{coro:instab} bounding the number of epochs ended due to instability.
\begin{lemma}
\label{lem:instab}
Let $\cE_i$ be an epoch with total variation $ \Delta_{\cE_i} \leq  \left(\sqrt{\frac{C_3}{4}} - \bar{C}_\var L^{-1/4}\right)/\bar{C}_{bias}$. Then under \textsc{Event 1}, with probability at least $1-\cO(1/T^3)$, the epoch does not end because of instability detection. 
\end{lemma}

\begin{corollary}
\label{coro:instab}
The expected number of epochs that end due to the instability test is bounded by $\cO(V_T \sqrt{\ln T}) $.
\end{corollary}

Combining the two bounds, we get $E = \cO(T^{1/5}V_T^{4/5})$. Therefore, we can bound the $\tilde{\cO}(\sqrt{E\cdot T})$ term in \eqref{eqn:regret_est_error} by $\tilde{\cO}(T^{3/5}V_T^{2/5})$.

\subsection{Bounding the Total Square Norm of the Estimation Error}
\label{sec:upper}

In this section, we analyse the regret due to the estimation error, i.e., the first term in \eqref{eqn:regret_est_error}. For succinctness, define the following \textit{loss} function for an arbitrary interval $\cI$:
\#
\label{eqn:defn_L}
\cL(I) &:= \sum_{t \in \cI} \min\left\{ C_4 \norm{\hat{\Theta}_{i,j-1} - \Theta_t}_F^2 , C_3 \right\}. 
\#
In the sequel, we first focus on an exploration epoch $\cE_i$ and bound $\cL(\cE_i)$. We then combine the regret of epochs to get the requisite regret bound of Theorem~\ref{thm:dynalg_regret}.

Our proof decomposes into three parts. First, we focus on one block, say block $j$, of epoch $i$, and prove a lemma that provides an upper bound for $\cL(\cI)$ for any interval $\cI \subseteq \cB_{i,j}$. Second, we partition a block into intervals with small total variation within each interval. We use the just mentioned bound to bound  $\cL(\cB_{i,j})$ of each block $j$ in an exploration epoch $i$ in terms of the length of the block and the total variation within the block. Finally, we upper bound the total number of blocks within an epoch $i$ and sum up the bound on $\cL(\cB_{i,j})$ for all the blocks in an epoch $\cE_i$ to obtain a bound on $\cL(\cE_i)$.

\begin{lemma}
\label{lem:interval_loss}
For an arbitrary interval $\cI = [s,e]$ that lies in
block $\cB_{i,j}$, define 
$\varepsilon_\cI \coloneqq \norm{ \hat{\Theta}_{i,j-1} -  \Theta_{s} }_F^2$
and $\alpha_\cI \coloneqq \frac{\log{|\cI|}}{\sqrt{\cI}}$.
Then, $\cL(\cI)$ can be bounded as
\*
\cL(\cI) 
 & = \cO \left( |\cI|\alpha_{\cI} + |\cI| \Delta^2_{\cI}+ |\cI|\varepsilon_\cI \mathbbm{1}\{\varepsilon_\cI \geq \alpha_\cI \} \right) .
\*
\end{lemma}
To get a bound for the regret for a block, we need to partition $\cB_{ij}$ into intervals with small variation. Specifically, we have the following lemma adapted from \cite{chen2019new}.
\begin{lemma}\label{lemma:partition}
There is a way to partition any block $\mathcal{B}$ into $\mathcal{I}_{1} \cup \mathcal{I}_{2} \cup \cdots \cup \mathcal{I}_{\Gamma}$ such that  
\[
\Delta_{\mathcal{I}_{k}}^2 \leq \frac{\log _{2} T}{\sqrt{|I_k|}} = \alpha_{I_k}, 
\qquad k \in [\Gamma],
\]
and the number of blocks $\Gamma$ satisfies 
$\Gamma=\mathcal{O}\left(\min \left\{S_{\cB},\left(\log|\cB|\right)^{-\frac{2}{5}} \Delta_{\mathcal{B}}^{\frac{4}{5}}|\mathcal{B}|^{\frac{1}{5}}+1\right\}\right)$.
\end{lemma}
The partition in Lemma~\ref{lemma:partition} is for the analysis only.
The intuition for this partition is to create small enough intervals 
so that their regret can be shown to be small, while at the same time 
not creating too many intervals. Applying Lemma~\ref{lem:interval_loss} to each 
interval of the partition of block $\cJ$:
\#\label{eq:block_upperbound}
 \cL(\cB) &\leq  \tilde{\cO}\Big( \sum_{k=1}^{\Gamma-1}|{\cI_k}|\alpha_{{\cI_k}}+\sum_{k=1}^{\Gamma-1} |{\cI_k}|\varepsilon_{\cI_k} \mathbbm{1}\{\varepsilon_{\cI_k} \geq \alpha_{\cI_k} \}\Big)  +\cL(\cI_\Gamma) .
\#
Plugging in the definition of $\alpha_{\cI_k}$,
we get  $|{\cI_k}|\alpha_{{\cI_k}} = \sqrt{\cI_k}\log{|\cI_k|}$.
Then  by the Cauchy-Schwartz inequality and the upper bound for $\Gamma$ from 
Lemma \ref{lemma:partition}, we have
\*
\sum_{k=1}^{\Gamma-1}  \sqrt{|\cI_k|}\log|\cI_k| \leq \sqrt{ (\Gamma-1) \sum_{k=1}^{\Gamma-1}|\cI_k|\log^2 |\cI_k|} =  \tilde{\cO}\left(\sqrt{(\Gamma-1)\sum_{k=1}^{\Gamma-1}|\cI_k|}\right) = \tilde{\cO}\left(|\mathcal{B}|^{\frac{3}{5}} \Delta_\cB^{\frac{2}{5}}\right).
\*
We defer the bound for the remaining
terms of \eqref{eq:block_upperbound} to Appendix~\ref{sec:proof_loss_function}. 
The following lemma presents the resulting upper bound for the
loss function of a block $\cB$.
\begin{lemma}\label{lemma:block_upperbound}
Let $\cB=\cB_{i,j}$ be a block of some epoch $i$ with $j>0$. 
It holds with high probability that \dynalg \ guarantees
\*
\cL(\cB)\leq \tilde{\cO} \left(  |\mathcal{B}|^{\frac{3}{5}} \Delta_{B}^{\frac{2}{5}}  + \sqrt{|\cB|}\right).
\*
\end{lemma}

From the geometrically increasing size of $\cB_{ij}$, 
we get $\sum_j \sqrt{|\cB_{ij}|} = \cO(|\cE_i|)$. 
From the H\"{o}lder's inequality, we get 
\*
\sum_{j} |\cB_{ij}|^{\frac{3}{5}} \Delta_{B_{i,j}}^{\frac{2}{5}} \leq \left( \sum_{j} |\cB_{ij}|\right)^{\frac{3}{5}} \left( \sum_j \Delta_{B_{i,j}}\right)^{\frac{2}{5}} = |\cE_i|^{\frac{3}{5}} \Delta_{\cE_i}^{\frac{2}{5}};
\*
so that $\cL(\cE_i) = \tilde{\cO}(|\cE_i|^{3/5}\Delta^{2/5}_{\cE_i} + \sqrt{\cE_{i}})$. 
One more application of the H\"{o}lder's inequality gives the
bound of $\tilde{\cO}(T^{\frac{3}{5}}V_T^{\frac{2}{5}})$,
proving Theorem~\ref{thm:dynalg_regret}.

\section{Regret Lower bounds}
\label{sec:lower}

In this section, we prove two lower bounds for 
the regret of the non-stationary LQR problem.
First, in Theorem~\ref{thm:lowerbound} we prove that for any given $V_T = o(T)$,
no learning algorithm can guarantee a regret $o(V_T^{3/5}T^{2/5})$,
showing that the regret of $\dynalg$ is minimax optimal as a function of $V_T$. 
Next, in Theorem~\ref{thm:lowerbound_window_based} we prove that 
a broad class of static-window based online learning algorithms 
are regret suboptimal for non-stationary LQR -- even if the algorithm 
has the knowledge of the variation $V_T$. This rules out several 
popular approaches that have been used in the literature
for learning under non-stationary such as UCB with static restart schedule
or bandit-on-bandit approaches to optimize the window size.

\begin{theorem}
\label{thm:lowerbound}
There exists a $T_0$ such that for any $T \geq T_0$, 
and a total variation $V_T$ of dynamics,
for any randomized online algorithm \textsc{Alg} 
(which knows $T, V_T$),
there exists a non-stationary LQR instance
with regret lower bounded as
\[
\expct{\Reg^{\textsc{Alg}}(T)} = \Omega\left( V_T^{3/5} T^{2/5} \right).
\]
Under switching dynamics with $S$ switches,
for any randomized algorithm $\textsc{Alg}$ (which knows $T, S$), 
there exists an instance with regret lower bounded as
\[
\expct{\Reg^{\textsc{Alg}}(T)} = \Omega\left( \sqrt{ST} \right). 
\]
\end{theorem}

\begin{proof} We build on the lower bound instance from 
\cite{cassel2020logarithmic}. Consider a randomly generated
one dimensional LQR problem instance with dynamics and cost:
\begin{align}
\nonumber    x_{t+1} &= a x_{t} + b u_t + w_t \ , \\
\label{eqn:1D_lower_instance}   c_t &= x_t^2 + u_t^2 \ ,
\end{align}
where $w_t \sim \mathcal{N}(0,1)$. The dynamics are given by $a = 1/\sqrt{5}$
and $b = \chi \sqrt{\epsilon}$, with $\chi$ being 
a Rademacher random variable that
takes values $\pm 1$ with equal probability. Standard results 
show that the optimal linear feedback controller for
the above LQR system is:
\begin{align}
\label{eqn:kstar_1d}
    k^* = - \frac{ab p^*}{1+b^2 p^*} 
\end{align} 
where $p^*$ solves
\begin{align}
\label{eqn:pstar_1d}
  p^* &= 1 + \frac{a^2 p^*}{1+b^2 p^*} .
\end{align} 
In \citet{cassel2020logarithmic}, the authors
prove the following lower bound on the regret of any algorithm.
\begin{theorem}[{\citet[Theorem 13]{cassel2020logarithmic}}] \label{thm:1D_lower_bound}
For $T \geq 12000$ and $\epsilon = \sqrt{T}/4$, the expected
regret of any deterministic learning algorithm for
system \eqref{eqn:1D_lower_instance} satisfies
\[ \expct{\Reg(T)} \geq \frac{\sqrt{T}}{3100} - 4  .\]
\end{theorem}
By Yao's theorem, the above implies that
for any randomized learning algorithm, there is an LQR instance 
with expected regret $\Omega(\sqrt{T})$.

We create a lower bound instance for a
non-stationary LQR problem with the total variation $V_T$ 
by pasting a sequence of these one-dimensional instances.
In particular, we concatenate $\floor{\frac{V_T}{2\sqrt{\epsilon}}}$ instances of
\eqref{eqn:1D_lower_instance} with horizon $\floor{\frac{1}{4\epsilon^2}}$ each, 
where $\epsilon$ satisfies $\frac{V_T}{2\sqrt{\epsilon}} = T \cdot 4 \epsilon^2$,
or equivalently $\epsilon = \left( \frac{V_T}{8T} \right)^{2/5}$. That is,
we re-randomize $\chi$ for every sub-instance. 
To demonstrate a lower bound, we further allow the learner the knowledge 
of the time instants at which a new sub-instance begins,
and the duration of the sub-instance. Theorem~\ref{thm:1D_lower_bound} implies 
that the regret of the learner for each sub-instance 
is $\Omega\left( \frac{1}{2\epsilon} \right)$, for a total regret
over the entire time horizon of 
$\Omega\left( \frac{V_T}{\epsilon^{3/2}} \right) = \Omega\left( V_T^{2/5} T^{3/5} \right)$. 

If, instead of bounded total variation, 
the non-stationary LQR instance has a 
piecewise constant dynamics with $S$ switches,
we create a lower bound instance similarly with $S$ sub-instances 
of horizon $\floor{T/S}$ each, 
and $\epsilon = \frac{\sqrt{T/S}}{4}$. The regret per 
sub-instance for any learner
is $\Omega(\sqrt{T/S})$ for a total 
regret lower bound of $\Omega(\sqrt{ST})$.
\end{proof}

\paragraph{Necessity of Adaptive Restarts.}
A common technique to handle non-stationary learning environments is
to use random restarts or sliding window algorithms to forget the distant history.
In learning problems where the rewards are linear in the unknown parameters 
(e.g., in multi-armed bandit problems), this gives the optimal regret rate
in terms of the total variation of the instance \textit{if} the 
window size is chosen optimally -- in the lower bound instance,
the adversary changes the instance by $\cO\left( V_T^{1/3}T^{-1/3}\right)$ at 
regularly spaced times. In the LQR problem, we instead have that the per-step regret 
$J^*(\Theta)- J(\Theta, K^*(\hat{\Theta}))$ is \textit{quadratic} in
$\norm{\Theta - \hat{\Theta}}_F$. Intuitively,
the adversary can maximally penalize a non-adaptive restart based algorithm 
by changing the instance by as much as $\Theta(1)$ at regularly spaced,
but randomly chosen times. This strategy fails against an adaptive restart 
algorithm such as \dynalg\ because big changes are easy to detect with less exploration effort.
To give a little more formal intuition, we consider the 
one-dimensional LQR problem \eqref{eqn:1D_lower_instance} from 
\cite{cassel2020logarithmic}, but with non-stationary $b_t$,
and a fairly general static window based algorithm for this non-stationary LQR instance. 
We prove that even with optimal tuning of the window size and an
arbitrary exploration strategy, it can incur a regret as large as $\Omega(V_T^{1/3}T^{2/3})$.

We first describe the one-dimensional instance
and the family of sliding window algorithms we consider. 
{\bf Instance:} 
The cost function is $x_t^2 + u_t^2$
and the dynamics are given by:
\[ 
x_{t+1} = a x_t + b_t u_t + w_t,
\]
with $x_1=0$ and $w_t \stackrel{i.i.d.}{\sim} \mathcal{N}(0,1)$. 
The dynamics parameter is time-invariant $a = 1/\sqrt{5}$
and known to the algorithm (therefore, there is no learning needed for $a$). 
The sequence $\{b_t\}$ is random and generated as follows.
Let $\epsilon = 0.05 \cdot \left( V_T/T\right)^{1/6}$.
We choose $b_1 = \epsilon$. For each subsequent $t$, with probability $\frac{V_T}{2T}$, 
$b_t$ is chosen to be $\pm 0.05$ with equal probability,
or, with probability $\left(V_T/4T \right)^{5/6}$,
$b_t$ is chosen to be $\pm \epsilon$ with equal probability,
otherwise $b_t = b_{t-1}$. The key feature of the instance is that 
while most of the time $b_t$ is small of size $\epsilon$
and most of the changes in $b_t$ are of order $\epsilon$ as well,
there are much rarer changes in $b_t$ of $\mathcal{O}(1)$ size. 
These two scales of changes make any fixed window size suboptimal for the regret.

{\bf Non-adaptive Restart with Exploration (RestartLQR($W$)) Algorithm.} 
We consider a family of algorithms parametrized by a window size $W$.
Let $\eta_t \stackrel{i.i.d.}{\sim} \mathcal{N}(0,1)$. The algorithm splits
the horizon $T$ into non-overlapping phases of duration $W$ each, 
and for time $t$ in phase $i$, the algorithm plays $u_t = \hat{k}_{(i)} x_t + \sigma_t \eta_t$, 
where $\hat{k}_{(i)}$ is a linear feedback controller estimated by the algorithm 
based \textit{only} on the trajectory observed in phase $(i-1)$,
and $\sigma_t$ is an arbitrary adapted sequence of exploration noise (energy)
injected by the algorithm. To emphasize, the algorithm is restricted in two senses.
First, it is restricted to playing a fixed linear feedback controller within each 
phase with Gaussian exploration noise. Second, at the beginning of each phase, the algorithm
forgets the entire history and restarts the estimation of the dynamics.

\begin{theorem}
\label{thm:lowerbound_window_based}
The expected regret of RestartLQR under optimally tuned window size $W$ and 
exploration strategy is at least $\Omega\left( V_T^{1/3} T^{2/3} \right)$.
\end{theorem}

\section{Concluding Thoughts}
\label{sec:conclusion}

In this paper, we have tried to fill an obvious gap in the literature -- the absence 
of any low dynamic regret algorithm for the control of a non-stationary LQR system under 
stochastic noise. We discuss the possibility of wider applicability of our results and
some open questions.

\paragraph{A Queueing Application.}
While in the paper we focused on the LQR problem, 
the key motif of the LQR problem that drove our results was
that \textit{(i)} given the state and action, the feedback we receive was a linear
function (i.e., linear feedback); and 
\textit{(ii)} given an $\epsilon$ error in the parameter estimates,
the optimal controller for the estimated parameters
has an $\mathcal{O}(\epsilon^2)$ additive suboptimality (i.e., quadratic cost).
Similar motif shows up in numerous other applications where 
we believe a similar regret trade-off would show up.
Here we mention a queueing example. Consider the following discrete time queueing 
system with a configurable server: the arrivals per period are i.i.d.~Bernoulli with 
a known mean $\lambda < 1$. The server has two resources (say CPU and memory) 
and the operator can choose a configuration $(x, y) \in \{ x^2 + y^2 \leq 1 ; x, y \geq 0 \}$ 
of the two resources. Given the configuration, the number 
of departures per period is also a Bernoulli random variable with mean, 
\[ \mu_t = \alpha_t x + \beta_t y, \]
where $\alpha_t, \beta_t \geq 0, \lambda < \alpha_t^2 + \beta_t^2 \leq 1$ represent
the resource requirements of the jobs, are non-stationary, and unknown to the operator.
Assume a job that arrives in time step $t$ can not be served before time step $t+1$. 
The cost at time step $t$ is $N_t$, the number of jobs in the system.
This system fits the motif of linear feedback and quadratic cost. 
The linear feedback can be seen by noting that the feedback at time step $t$ 
is the Bernoulli random variable for the number of departures,
which can be written as $ \alpha_t x_t + \beta_t y_t + \eta_t$,
where $\eta_t$ is a mean 0 bounded random variable (independent across time periods). 
To see the quadratic cost part, consider the steady-state problem with 
stationary $(\alpha, \beta)$, and a stationary action $(x,y)$ giving
$\mu_t = \mu = \alpha x + \beta y$. The steady-state average cost
would be $N(\mu,\lambda) = \frac{\lambda(1-\lambda)}{\mu - \lambda}$. In this case, 
the optimal action is to choose $(x_t, y_t)$ in the direction $(\alpha, \beta)$ under 
which $\mu^* = \alpha^2 + \beta^2$ with optimal cost $N^* = N_{\mu^*,\lambda}$. Consider 
an estimate 
$(\hat{\alpha},\hat{\beta})$ such that $|\alpha - \hat{\alpha}|+|\beta - \hat{\beta}| = \epsilon$.
If $\lambda \leq \alpha^2 + \beta^2 - \frac{1}{100}$, then the controller based
on the estimated $\hat{\alpha},\hat{\beta}$ gives cost $N^* + \Theta(\epsilon^2)$, which is
what we mean by a quadratic cost. We therefore expect that our 
results for the LQR problem would extend to the control of such queueing systems.  

\paragraph{Open Questions}
We believe both our algorithm and the regret analysis can be tightened, e.g., 
using sequential hypothesis testing to detect instability instead of
our current threshold based approach, and made parameter free.
An algorithm with a bound on regret  of the following 
flavor would be desirable: There exist constant
$\epsilon_0, T_0$ such that for a non-stationary LQR problem with 
variation $V_T = \epsilon T$,
where $\epsilon \leq \epsilon_0$ and $T \geq T_0$, the regret attained 
is at most $\epsilon^{2/5}T + o(T)$. It is also desirable to develop a notion of
instance-optimal regret -- instead of using the summary $V_T$ and 
presenting minimax optimal guarantees.

Yet another challenging direction is that there seem
to be two prevalent approaches to studying robustness for 
online control of LQR systems -- one with non-stochastic/adversarial noise and 
another with unknown non-stationary dynamics. This leaves an open problem of finding
a controller which achieves both types of robustness simultaneously
or proving the impossibility of doing so. A second open problem is to 
consider more general convex cost functions. Many of the elegant results in LQR theory,
and indeed the regret bounds in our paper, depend on the quadratic objective function.
A starting point would be to study a bandit problem with linear feedback,
but a general convex cost function. Finally, a notoriously hard problem is to study 
the robust control where the action set may depend on the state, 
which touches upon the theme of \textit{safe exploration}.
Doing so in the context of LQR could be fruitful.

\newpage 

{\small
\begin{spacing}{1.2}
\bibliographystyle{abbrvnat}
\bibliography{NonStyLQR}
\end{spacing}
}

\begin{appendix}
\clearpage
\section{Basic lemmas}
\label{sec:basic}

\begin{lemma}[Laurent-Massart Bound \cite{laurent2000adaptive}]
\label{lem:laurent-massart}
Let $a_1,\ldots, a_n$ be non-negative, and $X_1,\ldots,X_n$ be \textit{i.i.d.}~$\chi^2$ random variables. Let 
\[ |a|_\infty := \max_{i \in [n]} a_i  \quad \mbox{and} \quad |a|^2_2 := \sum_{ i \in [n]} a_i^2. \]
Then,
\begin{align*}
    \prob{\sum_{i \in [n]} a_i (X_i-1) \geq 2 |a|_2 \sqrt{x} + 2 |a|_{\infty} x} & \leq e^{-x},\\
    \prob{\sum_{i \in [n]} a_i (X_i-1) \leq - 2 |a|_2 \sqrt{x} } & \leq e^{-x}.
\end{align*}
\end{lemma}

\begin{lemma}
 \label{lem:bound_max_chi_squared}
 Let $Y_1, \ldots, Y_T$ be \textit{i.i.d.}~$\chi^2_k$ random variables. Then,
 \begin{align*}
     \expct{\max_{t \in T} Y_t} & \leq k + \max\{12 k , 3 \ln T\} +3,\\
     \expct{\max_{t \in T} \sqrt{Y_t}} & \leq \sqrt{k} + \sqrt{8 \ln T} + \sqrt{\frac{\pi}{2}}.
 \end{align*}
\end{lemma}
\begin{proof}
 By Laurent-Massart bound (Lemma~\ref{lem:laurent-massart}),
 \*
 \prob{Y_t \geq k + 2\sqrt{kx} + 2x} & \leq e^{-x}.
 \*
 For $y \geq 12k$, we have the following sequence of implications
  \*
 \prob{Y_t \geq k + y} & \leq e^{-\frac{y}{3}},\\
 \implies  \prob{Y_t \leq k + y} & \geq 1 - e^{-\frac{y}{3}} \\
 \implies  \prob{\max_{t \in [T]} Y_t \leq k + y} & \geq \left( 1 - e^{-\frac{y}{3}}\right)^T \\
 \implies \prob{\max_{t \in [T]} Y_t \geq k + y} & \leq 1 - \left( 1 - e^{-\frac{y}{3}}\right)^T.
 \*
 Let $y = \max\{12k,3\ln T \} + z$ for $z \geq 0$. Then,
\*
\prob{\max_{t \in [T]} Y_t \geq k + y} & \leq 1 - \left( 1 - e^{-\frac{\max\{12k,3\ln T\} + z}{3}}\right)^T \\
& \leq  1 - \left( 1 - \frac{1}{T}e^{-\frac{z}{3}}\right)^T \\
& \leq e^{-\frac{z}{3}}.
\*
From the above,
\*
\expct{ \max_{t \in T} Y_t} &\leq k + \max\{12 k , 3 \ln T\} + \int_{z=0}^{\infty} e^{\frac{z}{3}} dz \leq k + \max\{12 k , 3 \ln T\} +3.
\*

For the second part, we again begin from the Laurent-Massart bound. For any $x \geq 0$,
 \*
 \prob{\sqrt{Y_t} \geq \sqrt{k} + \sqrt{2x}} & \leq e^{-x},
 \*
 which in turn implies for $y\geq 0$,
 \*
 \prob{\sqrt{Y_t} \geq \sqrt{k} + \sqrt{2 \ln T} + y} & \leq e^{-\frac{y^2 + 2 y \sqrt{2 \ln T}}{2}} \leq e^{-\frac{y^2}{2}}.
 \*
 Further substituting $y = \sqrt{2 \ln T} + z$ for $z \geq 0$, 
 \*
 \prob{\max_{t} \sqrt{Y_t} \geq \sqrt{k} + 2 \sqrt{2 \ln T} + z} &  \leq 1 - \left(1- \frac{1}{T}e^{-\frac{z^2 + 2z\sqrt{2 \ln T}}{2}} \right)^T  \leq  e^{-\frac{z^2}{2}}.
 \*
 Finally, 
 \*
 \expct{ \max_{t} \sqrt{Y_t} } & \leq \sqrt{k} + \sqrt{8 \ln T} + \int_{0}^\infty e^{-\frac{z^2}{2}}dz \\
 &= \sqrt{k} + \sqrt{8 \ln T} + \sqrt{\frac{\pi}{2}}.
 \*
\end{proof}

The following lemma is adapted from \cite{hajek1982hitting}, but we prove it here for completeness.
\begin{lemma}
\label{lem:hitting_time}
Let $W_1, W_2, \ldots$ be a non-negative stochastic process, and $\left(\cW_t\right)_{t \in \NN}$ be the induced filtration.
Let $Y_0, Y_1, \ldots$ be a non-negative stochastic process adapted to $\cW_t$ such that for some $0 < \rho < 1$, for all $t \geq 0$,
\[ Y_{t+1} \leq \rho Y_{t} + W_{t+1}, \qquad \qquad \mbox{almost surely.}\]
Let $a \geq 0$ and $\rho \leq \hat{\rho} < 1$ be such that for all $t \geq 1$, 
\[ \expct{ \left.  \rho + \frac{W_{t+1}}{a} \ \right|\  \cW_t} \leq \hat{\rho}. \]
Define the $a$-hitting time of process $\{Y_t\}$ as:
\[ \tau_a = \min_{ k \geq 1 }\left\{ Y_k \leq a \right\}. \]
Then,
\begin{enumerate}
    \item $\prob{\tau_a \geq k \mid \cW_0} \leq \frac{Y_0}{a} \hat{\rho}^t$,
    \item $\expct{ \sum_{k=0}^{\tau_a} Y^2_k \mid \cW_0} \leq \frac{Y^2_0}{1-\hat{\rho}^2}\leq \frac{Y^2_0}{1-\hat{\rho}}.$
\end{enumerate}
\end{lemma}
\begin{proof}
Conditioning on the event $\{Y_t \geq a\}$ and using the definition of $\hat{\rho}$ above, 
\begin{align*}
    \expct{Y_{t+1} \mid \cW_t, Y_t \geq a} &\leq \expct{ \rho Y_t + W_{t+1} \mid \cW_t, Y_t \geq a}  \\
    & = \expct{ \left. Y_t \left(\rho + \frac{W_{t+1}}{Y_t}\right) \ \right| \ \cW_t , Y_t \geq a  } \\
    & \leq \hat{\rho} \cdot Y_t.
\end{align*}
Therefore, the stopped process $Y_{t \wedge \tau_a}/\hat{\rho}^{t \wedge \tau_a}$ is a non-negative supermartingale, and hence
\begin{align*}
    Y_0 &\geq \expct{\left. \frac{Y_{k \wedge \tau_a}}{\hat{\rho}^{k \wedge \tau_a}}\ \right| \ \cW_0 } 
    \ \geq \  \expct{\left. \frac{Y_{k \wedge \tau_a}}{\hat{\rho}^{k \wedge \tau_a}} \mathbbm{1}\{\tau_a \geq k\} \ \right| \ \cW_0 } 
    \ \geq \ \frac{a}{\hat{\rho}^{k}} \prob{\tau_a \geq k \mid \cW_0}. 
\end{align*}
That is, $\prob{\tau_a \geq k \mid \cW_0} \leq \frac{Y_0}{a} \hat{\rho}^k$, proving the first part of the lemma. For the second part,
\begin{align*}
    \sum_{k=0}^{\tau_a} Y^2_k &= \sum_{k=0}^{\infty} Y^2_k \cdot \mathbbm{1}\{ \tau_a \geq k \}.
\end{align*}
Taking the expectation and using the supermartingale result from above, 
\begin{align*}
        \expct{\sum_{k=0}^{\tau_a} Y^2_k \mid \cW_0} &= \sum_{k=0}^{\infty} \expct{ Y^2_k \cdot \mathbbm{1}\{ \tau_a \geq k \} \mid \cW_0} 
        \  \leq \ \sum_{k=0}^{\infty} Y^2_0 \hat{\rho}^{2k} 
        \ = \ \frac{Y^2_0}{1-\hat{\rho}^2}.
\end{align*}
\end{proof}

The following lemma on hitting times of exponentially ergodic random walks will be helpful for bounding the number of epochs that end because of instability detection through $\norm{x_t}$ becoming large.

\begin{lemma}
\label{lem:hitting_upper}
Let $Y_0, Y_1, \ldots$ be a non-negative stochastic process satisfying 
\[  Y_{t+1} \leq \rho Y_t + \sum_{i=1}^{m} \beta_{i,t+1} \mid W_{i,t+1}| \]
where $\rho < 1$, and $W_{i,t}$ are \textit{i.i.d}~$\mathcal{N}(0,1)$ random variables. Furthermore, let $\max_{i,t} \beta_{i,t} \leq B$, and $\bar{a} = \left( \frac{\sqrt{8m}B}{\sqrt{1-\rho}} \sqrt{\log T} + \frac{mB}{1-\rho} \right)$. Then,
\[ \prob{\max_{t \in [T]} Y_t \geq Y_0 + \bar{a}} \leq \frac{1}{T^3}. \]
\end{lemma}
\begin{proof}
Extending the sequence of random variables for $t \leq 0$, we get the following upper bound on $Y_t$:
\[ Y_{t} \leq \rho^t Y_0 + \sum_{k=0}^{\infty} \rho^{k} B \sum_{i=1}^m |W_{i,t-k}|.  \]
Let 
\[ S_t := \sum_{k=0}^{\infty} \rho^{k} B \sum_{i=1}^m |W_{i,t-k}|. \]
Therefore, for $a \geq 0$,
\*
\prob{ Y_t \geq Y_0 + a} & \leq \prob{ \rho^t Y_0 + S_t \leq Y_0 + a } \ \leq \prob{S_t \leq a}.
\*
Furthermore,
\*
S_t^2 &= B^2\left(  \sum_{k=0}^\infty  \sum_{i=1}^m \rho^{2k} W^{2}_{i,t-k} + \sum_{ (k_1, i_1)\neq (k_2,i_2) } \rho^{k_1+k_2} |W_{i_1,t-k_1}|\cdot |W_{i_2,t-k_2}| \right) \\
& \leq B^2\left(  \sum_{k=0}^\infty \sum_{i=1}^m \rho^{2k}  W^{2}_{i,t-k} + \frac{1}{2}\sum_{ (k_1, i_1)\neq (k_2,i_2) } \rho^{k_1+k_2} \left( W^2_{i_1,t-k_1}  + W^2_{i_2,t-k_2}\right) \right) \\
&= m B^2 \sum_{k=0}^\infty \sum_{i=1}^m \frac{\rho^{k}}{1-\rho} W^2_{i,k}. 
\*
Applying Laurent-Massart bound from Lemma~\ref{lem:laurent-massart},
\*
\prob{ S^2_t \geq \frac{m^2B^2}{(1-\rho)^2} + \frac{2m^{3/2}B^2}{(1-\rho)\sqrt{1-\rho^2} } \sqrt{x} +  \frac{2mB^2}{1-\rho} x } &\leq e^{-x}.
\*
A simple upper bound on the right hand side within $\prob{\cdot}$ gives,
\#
\label{eqn:random_walk_excursion}
\prob{ S_t \geq \frac{mB}{1-\rho} + \frac{\sqrt{2m}B}{\sqrt{1-\rho}} \sqrt{x} } & \leq e^{-x}.
\intertext{Substituting $x = 4 \ln T$,}
\nonumber
\prob{ Y_t \geq Y_0 + \frac{mB}{1-\rho} + \frac{\sqrt{8m}B}{\sqrt{1-\rho}} \sqrt{\ln T} } & \leq \frac{1}{T^4}.
\#
A union bound completes the final argument.
\end{proof}

For reference, we note some basic matrix norm inequalities:
\begin{enumerate}
    \item $ \frac{1}{2} \norm{\Theta - \hat{\Theta}} \leq \max\left\{\norm{A - \hat{A}}, \norm{B - \hat{B}} \right\} \leq \norm{\Theta - \hat{\Theta}}$;
    \item $ \frac{1}{\sqrt{n}} \norm{\Theta - \hat{\Theta}}_F \leq \norm{\Theta - \hat{\Theta}} \leq  \norm{\Theta - \hat{\Theta}}_F $.
\end{enumerate}
Lemma~\ref{lem:quadratic} states that $\norm{\Theta - \hat{\Theta}}_F \leq C_3$ implies $J^*(\Theta) - J(\Theta,K^*(\hat{\Theta})) \leq C_4 \norm{\Theta - \hat{\Theta}}_F^2$.
\section{Proof of Sequential Strong Stability}
\label{sec:proof_ss}

\begin{lemma}[\cite{gahinet1990sensitivity}]
\label{lemma:sensitivity}
Let $X$ be the solution to the Lyapunov equation 
\*
X-F^{\top} X F=M.
\*
Let $X+\Delta X$ be the solution to the perturbed problem
\*
Z-(F+\Delta F)^{\top} Z (F+\Delta F)=M.
\*
Then the following inequality holds for the spectral norm:
\*
\frac{\|\Delta X\|}{\|X+\Delta X\|} \leq 2 \left\|\sum_{k=0}^{+\infty}\left(F^{\top}\right)^{k} F^{k}\right\|\cdot (2\|F\|+\|\Delta F\|) \cdot {\|\Delta F\|}.
\*
\end{lemma}

\paragraph{Proof of Lemma~\ref{lem:nonstat_seq_stg_stab}}

Let $P_t := P(\Theta_t, \hat{K})$ and  $P_{t+1}:=P(\Theta_{t+1}, \hat{K})$ be the solutions to the following Lyapunov equations, respectively:
\*
P_t &= Q+\hat{K}^{\top} R \hat{K}+(A_t+B_t \hat{K})^{\top} P_t (A_t+B_t \hat{K}),\\
P_{t+1} &= Q+\hat{K}^{\top} R \hat{K}+(A_{t+1}+B_{t+1} \hat{K})^{\top} P_{t+1}(A_{t+1}+B_{t+1} \hat{K}).
\*
Taking $X = P_t$, $X+\Delta X  = P_{t+1}$, $F = A_t+B_t\hat{K}$, $F +\Delta F = A_{t+1}+B_{t+1} \hat{K}$, and applying Lemma~\ref{lemma:sensitivity}, we get the following Lemma as a corollary. 
\begin{lemma}
\label{lemma:PP}
 It holds that 
 \*
    P_t  \preceq P_{t+1}\cdot\left( 1+ \frac{2(1-\gamma)^2}{1-(1-\gamma)^2}(2(1-\gamma)+ (\kappa+1)\|\Theta_{t+1} -\Theta_{t}\|)\right)\|\Theta_{t+1} -\Theta_{t}\|.
    \*
 \end{lemma}
\begin{proof}
Applying Lemma~\ref{lemma:sensitivity} with $X = P_t$, $X+\Delta X  = P_{t+1}$, $F = A_t+B_t\hat{K}$, and $\Delta F = A_{t+1}+B_{t+1} \hat{K} -  (A_t+B_t\hat{K}) = (A_{t+1} - A_{t}) + (B_{t+1} - B_{t})\hat{K} $, we have 
\*
\frac{\|P_{t+1} - P_t\| }{ \|P_{t+1}\|}&\leq 2 \left\|\sum_{k=0}^{+\infty}\left((A_t+B_t \hat{K})^{\top}\right)^{k} (A_t+B_t \hat{K})^{k}\right\|\\
&\quad\cdot (2\|(A_t+B_t \hat{K})\|+\|(A_{t+1} - A_{t}) + (B_{t+1} - B_{t})\hat{K}\|) \cdot {\|(A_{t+1} - A_{t}) + (B_{t+1} - B_{t})\hat{K}\|}\\
&\leq \frac{2(1-\gamma)^2}{1-(1-\gamma)^2} \cdot(2(1-\gamma)+(1+\kappa)\|\Theta_{t+1} -\Theta_{t}\|)\|\Theta_{t+1} -\Theta_{t}\|,
\*
where in the last inequality we use $\|A_t+B_t \hat{K}\|\leq 1-\gamma$ and $\|\hat{K}\|\leq \kappa $. Then by direct computation, we have 
\*
\|P_t\|&\leq \|P_{t+1}\| + \|P_{t+1} - P_t\| \\
&\leq  \|P_{t+1}\|\cdot\left( 1+ \frac{2(1-\gamma)^2}{1-(1-\gamma)^2}(2(1-\gamma)+ (\kappa+1)\|\Theta_{t+1} -\Theta_{t}\|)\|\Theta_{t+1} -\Theta_{t}\|\right).
\*
 \end{proof}

In the sequel, we first prove that $K$ is $(\kappa,\gamma)$-strongly stable for $A_{t}+B_{t} K=H_{t} L_{t} H_{t}^{-1}$. Note that by our assumption,  
\*
  J^*(\Theta_t, \hat{K} )  \leq J^*(\Theta_t) + C_4  \left\|\Theta_{t}-\hat{\Theta}\right\|_{F} ^{2}\leq  J^*(\Theta_t) + C_4 C_3\leq J_\cI^* + C_4 C_3 \coloneqq \tilde{J}_\cI^*.
\* 
We have  $\lambda_{\max}(P_t) \leq \tilde{J}_\cI^*/\psi^{2}$ and $\|H_t\|\leq \sqrt{\tilde{J}_\cI^*}/\psi\eqqcolon B_0$. By definition, we have 
\*
P_t &= Q+K^{\top} R K+(A_t+B_t K)^{\top} P_t(A_t+B_t K)\\
&\succeq  q_{\min} I +   r_{\min} K^{\top}K +  (A_t+B_t K)^{\top} P_t(A_t+B_t K)\\
&\succeq   q_{\min} I +  (A_t+B_t K)^{\top} P_t(A_t+B_t K).
\*
Specifically, we have $P_t \succeq q_{\min} I $. Hence $\|H_t^{-1}\| \leq q_{\min}^{-1/2} \eqqcolon 1/b_0$. Then setting $\kappa = B_0/b_0 = \sqrt{\frac{\tilde{J}_\cI^*}{\psi^2 q_{\min}}}$ will suffice. 
By $P_t \succeq r_{\min} K^\top K$, we have
\*
\|{K}\|\leq \sqrt{\frac{\|P_t\|}{r_{\min}}}\leq   \sqrt{\frac{\tilde{J}_\cI^*}{\psi^{2} r_{\min}}}\eqqcolon \kappa.
\*
Moreover, 
\*
L_t^\top L_t  &= P_{t}^{-1 / 2}\left(A_{t}+B_{t} K\right)^\top P_{t}\left(A_{t}+B_{t} K\right)P_{t}^{-1 / 2}\\
&\preceq  P_{t}^{-1 / 2} \left(P_{t} -  q_{\min} I \right) P_{t}^{-1 / 2}\\
&\preceq I -  q_{\min} P_{t}^{-1 }.
\*
Then
\*
\| L_t\|^2\leq 1-  \frac{q_{\min}\psi^{2}}{\tilde{J}_\cI^*}
\*
and
\*
\|L_t\| \leq \sqrt{ 1-  \frac{q_{\min}\psi^{2}}{J^*}} \leq 1- \frac{q_{\min}\psi^{2}}{2\tilde{J}_\cI^*}.
\*

In the sequel,  we prove the  $(\kappa,\gamma)$-sequentially strongly stability.
By direct computation, we have
\*
\|H_{t+1}^{-1} H_{t}\|^{2}&=\|P_{t+1}^{-1 / 2} P_{t}^{1 / 2}\|^{2}\\
&= \|P_{t+1}^{-1 / 2} P_{t} P_{t+1}^{-1 / 2}\|\\
&\leq\left( 1+ \frac{2(1-\gamma)^2}{1-(1-\gamma)^2}(2(1-\gamma)+ (\kappa+1)\|\Theta_{t+1} -\Theta_{t}\|)\|\Theta_{t+1} -\Theta_{t}\|\right),
\*
where in the inequality we apply Lemma~\ref{lemma:PP}. By the fact that $\sqrt{1+x}\leq 1+\frac{1}{2}x$ for $x\geq0$, we have 
\*
\|H_{t+1}^{-1} H_{t}\| 
&\leq  1+ \frac{(1-\gamma)^2}{1-(1-\gamma)^2}\left(2(1-\gamma)+ (\kappa+1)\|\Theta_{t+1} -\Theta_{t}\|\right)\|\Theta_{t+1} -\Theta_{t}\|\\
&\leq  1+ \frac{2(1-\gamma)^2}{1-(1-\gamma)^2}\left((1-\gamma)+ (\kappa+1)\right)\|\Theta_{t+1} -\Theta_{t}\|,
\*
where in the last step we use that $\norm{\Theta_{t+1}-\Theta_t} \leq 2 \sqrt{C_3} \leq 2$ by our assumption that $C_3 \leq 1$.

 \paragraph{Proof of Lemma~\ref{lem:nonstat_seq_stg_stab_bdd_norm}}
 Without loss of generality, we let $s_\cI = 1$. Since $x_{t+1}=\left(A_{t}+B_{t} K\right) x_{t}+ \sigma_t B_t \eta_t + w_{t}$, we have 
 \*
 x_{t}=M_{1} x_{1}+\sum_{s=1}^{t-1} M_{s+1} ( \sigma_s B_s \eta_s + w_s),
 \*
 where we define $M_{t}=I$ and $M_s = \prod_{j=s}^{t-1}\left(A_{j}+B_{j} K_{j}\right)$. Moreover, 
 \*
 \left\|M_{s}\right\|&=\left\|\prod_{j=s}^{t-1} H_{j} L_{j}^{\top} H_{j}^{-1}\right\|\\
 &\leq\left\|H_{t-1}\right\|\left(\prod_{j=s}^{t-1}\left\|L_{j}\right\|\right)\left(\prod_{j=s}^{t-2}\left\|H_{j+1}^{-1} H_{j}\right\|\right)\left\|H_{s}^{-1}\right\|\\
 &\leq  B_{0}(1-\gamma)^{t-s}\left(\prod_{j=s}^{t-2}\left\|H_{j+1}^{-1} H_{j}\right\|\right)\left(1 / b_{0}\right)\\
 &\leq \kappa (1-\gamma)^{t-s}\left(\prod_{j=s}^{t-2}\left\|H_{j+1}^{-1} H_{j}\right\|\right).
 \*
 Using the fact that $1+x\leq e^x$, we have 
 \*
\prod_{j=s}^{t-2}\left\|H_{j+1}^{-1} H_{j}\right\| &\leq e^{\sum_{j=s}^{t-2}\frac{2(1-\gamma)^2}{1-(1-\gamma)^2}((1-\gamma)+ (\kappa+1))\|\Theta_{t+1} -\Theta_{t}\|}\\
&\leq e^{C_{ss} V_{[s,t-1]}}
 \*
for some constant $C_{ss}$. Then it holds that
 \*
 \|M_{s}\|&\leq \kappa (1-\gamma)^{t-s}e^{C_{ss} V_{[s,t-1]}}\\
 &\leq \kappa e^{-\gamma(t-s)} e^{C_{ss} V_{[s,t-1]}}.
 \*
 Then we can bound the norm of $x_t$ as 
 \*
 \left\|x_{t}\right\| &\leq\left\|M_{1}\right\|\left\|x_{1}\right\|+\sum_{s=1}^{t-1}\left\|M_{s+1}\right\|\left\| \sigma_s B_s \eta_s + w_{s}\right\|\\
 &\leq \kappa e^{-\gamma(t-1)} e^{C_{ss} V_{[1,t-1]}}\|x_1\| +  \kappa e^{-\gamma(t-s)} e^{C_{ss} V_{[s,t-1]}}\sum_{s=1}^{t-1}(1-\gamma )^{t-s-1}\left\|\sigma_s B_s \eta_s + w_{s}\right\|\\
 &\leq\kappa e^{-\gamma(t-1)} e^{C_{ss} V_{[1,t-1]}} \|x_1\| +  \kappa e^{-\gamma(t-s)} e^{C_{ss} V_{[s,t-1]}} \max _{1<s<t}\left\|\sigma_s B_s \eta_s + w_{s}\right\|  \sum_{t=1}^{\infty}(1-\gamma )^{t}\\
 &=  \kappa e^{-\gamma(t-1)+C_{ss} V_{[1,t-1]}} \|x_1\| +
 \frac{\kappa e^{-\gamma(t-s) + C_{ss} V_{[s,t-1]}} }{\gamma} \max _{1<s<t}\left\|\sigma_s B_s \eta_s + w_{s}\right\|.  
 \*

\section{Estimation error bounds for OLS with non-stationary dynamics}
\label{sec:ols_est_proof}

\subsection{Proof of error bound for $\cI \subseteq \cB_{ij}$ (Lemma~\ref{lemma:ols_esti_new})}

Given the OLS estimator for interval $\cI=[s,e]$, 
\[ \hat{\Theta}_{\cI} = \argmin_{\Theta} \sum_{t\in \cI} \norm{ x_{t+1} - \Theta [ x_t^\top \  u_t^\top ]^\top  }_F^2, \]
our goal is to bound the estimation error $\norm{\hat{\Theta}_\cI - \bar{\Theta}}_F$ where $\bar{\Theta}$ is a `representative' $\Theta$ for $\{\Theta_t\}_{t \in \cI}$, for example $\Theta_e$. We will assume that during the entire interval $\cI$, the controller $K_t=K$ is stationary. Let $M\coloneqq  \begin{bmatrix} I_n \\ K \end{bmatrix}$.
We will use the notation 
\[  y_t = M x_t  \quad \mbox{and} \quad \xi_t  =  {\sigma_t} \tilde{\eta}_t  = {\sigma_t} \begin{bmatrix} 0 \\ I_d \end{bmatrix}  \eta_t, \quad \mbox{so that:}\quad z_t = \begin{bmatrix} x_t \\  u_t \end{bmatrix} = y_t + \xi_t \quad  \mbox{and} \quad  x_{t+1} = \Theta_t z_t  + w_t. \]
By our choice of $\sigma_t$, we have $\sigma_L^2  := \nu_{1} =  \sqrt{\frac{C_0}{L}}  \geq \sigma_t^2 \geq \sigma_\cI^2 := \sqrt{\frac{C_0}{|\cI|}} $ for all $t\in \cI $. With these notations, we can write the OLS loss function and estimator as:
\[ \hat{\Theta}_{\cI} = \argmin_{\Theta} \cL(\Theta), \qquad \mbox{where} \quad \cL(\Theta) =  \sum_{t\in \cI} \norm{ x_{t+1}  - \Theta z_t  }_F^2 = \sum_{t\in \cI} \norm{ \Theta_t z_t + w_t  - \Theta z_t  }_F^2. \]

Due to the OLS objective function, we can decompose this problem and estimate each row of $\hat{\Theta}_\cI$ separately. Towards that end, let us fix a row $i$.  With abuse of notation, denote the $i$th rows of $\Theta_t, \Theta, \hat{\Theta}_I, \bar{\Theta}$ by $\theta_t, \theta, \hat{\theta}, \bar{\theta}$, respectively. Let us also use $\omega_t$ to denote the $i$th entry of $w_t$. The OLS estimation problem for the row $i$ becomes:
\[ \hat{\theta} = \argmin_{\theta} \cL(\theta), \qquad \mbox{where,} \quad \cL(\theta) = \sum_{t\in \cI} \left( \inner{\theta_t}{z_t} + \omega_t  - \inner{\theta}{z_t}  \right)^2. \]
The solution for this OLS estimation problem is given by the solution of the following linear system:
\[  \hat{\theta} \left( \sum_{t \in \cI} z_t z_t^\top \right)  = \left( \sum_{t\in \cI} \theta_t z_t z_t^\top  \right) + \sum_{t\in \cI} \omega_t z_t^\top, \]
or
\[  \hat{\theta}   = \left( \sum_{t\in \cI} \theta_t z_t z_t^\top  \right) \left( \sum_{t \in \cI} z_t z_t^\top \right)^{-1} + \left( \sum_{t\in \cI} \omega_t z_t^\top \right) \left( \sum_{t \in \cI} z_t z_t^\top \right)^{-1}. \]
The second term above is a martingale sum, since $\omega_t$ is zero mean and independent of $z_t$, and contributes to the variance of the estimator. However, the first term which contributes to the `bias' is non-trivial. In the stationary case, $\theta_t = \theta$ and the first term becomes $\theta$, which implies that the OLS estimator is unbiased. However, in the non-stationary case, the first term can be far from $\bar{\theta}$ even when all the $\theta_{t}$ are close to each other. This necessitates a fresh analysis of the OLS estimator in the non-stationary setting.

The key obstacle in the analysis of the estimation error $\norm{\hat{\theta} - \bar{\theta}}^2$, is that while $z_t$ lives in $\RR^{n+d}$, most of its variance is in the $n$-dimensional column space of $[I_n \ K^\top]^\top$. This is because the LQR dynamics naturally adds the noise $w_{t-1}$ to arrive at the state $x_t$. In fact, this is precisely the reason we add the exploration noise $\xi_t$: to be able to distinguish changes in $\Theta_t = [A_t\ B_t]$ that are orthogonal to the column space of $[I_n\ K^\top]^\top$. However, this also means that we can not use a naive analysis based on a lower bound on the eigenvalues of the matrix $\sum_{t \in \cI} z_t z_t^\top$. 

Our approach to bounding the estimation error of the OLS estimator is to begin by looking at the one dimensional OLS problems parametrized by $v \in \SSS^{n+d} := \{ v \in \RR^{n+d}, \norm{v}=1\}$:
\[ \lambda_v = \argmin_{\lambda} \cL(\bar{\theta} + \lambda \cdot v), \]
and argue that $|\lambda_v|$ are small for `enough' directions $v$. That is, in enough directions, the minimizer of the 1-dimensional quadratic defined above is close to the candidate $\bar{\theta}$. Finally, we will show via an $\epsilon$-net argument that this implies that the true OLS estimator $\hat{\theta}$ is also close to $\bar{\theta}$.

{\bf Step 1: Decomposing the problem into orthogonal subspaces.} Fixing a direction $v$, the first order conditions for the minimizer $\lambda_v$ of $\cL(\bar{\theta} + \lambda \cdot v)$ gives:
\begin{align}
\label{eqn:ols_1d_foc}
    \lambda_v \sum_{t} \inner{v}{z_t}^2 =  \sum_t \inner{\theta_t - \bar{\theta}}{z_t} \cdot \inner{v}{z_t} + \sum_t \omega_t \inner{v}{z_t}. 
\end{align}
For a vector $u \in \RR^{n+d}$, let $u^\parallel$ and $u^\perp$ denote the projections onto the column space of $[I_n \ K^\top]^\top$ and its orthogonal space, respectively. That is,
\[ u^\parallel = \begin{bmatrix} I_n \\ K\end{bmatrix} (I + K^\top K)^{-1}  \begin{bmatrix} I_n \\ K\end{bmatrix}^\top u \quad \mbox{and} \quad u^\perp = u - u^\parallel.\]
Similarly, let $\hat{u}^\parallel$ and $\hat{u}^\perp$ denote the unit vectors in the direction $u^\parallel$ and $u^\perp$, respectively.

For analysis, it will be convenient to generalize the one dimensional problem of finding the minimizer on the line $\bar{\theta} + \lambda \cdot v$ to instead finding the minimizer in the plane $\bar{\theta}+\lambda^\parallel \hat{v}^\parallel + \lambda^\perp \hat{v}^\perp$, where we seek the optimal values of $\lambda^\parallel$ and $\lambda^\perp$. From \eqref{eqn:ols_1d_foc}, denoting $V := \sum_t z_t z_t^\top$, the Hessian of the corresponding quadratic loss function is given by
\begin{align}
    H_{\hat{v}^\parallel, \hat{v}^\perp} &= 
    \begin{bmatrix}
 \left( \hat{v}^\parallel \right)^\top V \hat{v}^\parallel & 
  \left( \hat{v}^\parallel \right)^\top V \hat{v}^\perp \\
   \left( \hat{v}^\perp \right)^\top V \hat{v}^\parallel & 
  \left( \hat{v}^\perp \right)^\top V \hat{v}^\perp \\
    \end{bmatrix} \ 
    = 
    \begin{bmatrix}
    \sum_t \inner{\hat{v}^\parallel}{z_t^\parallel}^2 & 
  \sum_{t} \inner{\hat{v}^\parallel}{z_t^\parallel}\inner{\hat{v}^\perp}{\xi_t^\perp}  \\
\sum_{t} \inner{\hat{v}^\parallel}{z_t^\parallel}\inner{\hat{v}^\perp}{\xi_t^\perp} & 
    \sum_t \inner{\hat{v}^\perp}{\xi_t^\perp}^2 \\
    \end{bmatrix}.
\end{align}
A careful analysis on $H_{\hat{v}^\parallel, \hat{v}^\perp}$ later yields the following lemma that indicates it suffices to consider the following two simpler cases: $v =\hat{v}^\parallel $ and $v = \hat{v}^\perp$. 

\begin{lemma}\label{lemma:quadratic_lemma_final}
Let $\lambda_{\hat{v}^\parallel} = \argmin_{\lambda} \cL(\bar{\theta} + \lambda \cdot \hat{v}^\parallel)$ and $\lambda_{\hat{v}^\perp} = \argmin_{\lambda} \cL(\bar{\theta} + \lambda \cdot \hat{v}^\perp)$. It holds with probability at least $1-11\delta$ that 
\*
\lambda_v^2 &\leq 2\lambda_{\hat{v}^\parallel}^2 +2 \lambda_{\hat{v}^\perp}^2.
\*
\end{lemma}
\begin{proof}\label{pf:quadratic_lemma_final}
Combining Lemma~\ref{lem:quadratic_lemma} and Lemma~\ref{lem:quadratic_lemma_helper}, it suffices to prove that 
\*
\max \left\{ 
\frac{\sum_t \left|\inner{\hat{v}^\parallel}{z_t^\parallel}\inner{\hat{v}^\perp}{\xi_t^\perp}\right|}{ \sum_t \inner{\hat{v}^\parallel}{z_t^\parallel}^2},
\frac{\sum_t \left|\inner{\hat{v}^\parallel}{z_t^\parallel}\inner{\hat{v}^\perp}{\xi_t^\perp}\right|}{\sum_t \inner{\hat{v}^\perp}{\xi_t^\perp}^2 }
\right\} \leq \frac{1}{33}
\*
holds with probability at least $1- 11\delta $.

Note that $z_t  = y_t + \xi_t =  M x_{t-1} + \xi_t = M \Theta_{t-1}  z_{t- 1} +  Mw_{t-1}  + \xi_t$. We have
\*
z_t^{\parallel} = M \Theta_{t-1}  z_{t- 1} +  Mw_{t-1}  + \xi_t^{\parallel}, \quad z_t^{\perp} = \xi^{\perp}_t.
\*

{\bf Step 0: Bound on $\sum_{t \in \cI} \norm{z_t}^2$.}
We begin with the following corollary of Lemma~\ref{lem:zu_bound}: For $|\cI| \geq 16 \ln \frac{1}{\delta}$, conditioned on $\max_{t \in \cI} \norm{x_t} 
\leq x_u$, it holds with probability at least $1-\delta$ that 
\*
\sum_{t\in \cI} \norm{z_t}^2 \leq 2 |\cI|\left((1+K_u^2) x_u^2+ 2\sigma_L^2\right) =: |\cI| z_u^2.
\*

{\bf Step 1: Upper bound on the numerator.} Recall $\xi_t = \sigma_t \tilde{\eta}_t$ and $ \sigma_t \leq \nu_1 =: \sigma_L$. Let 
\[ \sigma^2_{\hat{v}^{\perp}} = \expct{(\hat{v}^{\perp}) ^\top \begin{pmatrix} 0_n \\ \eta_t\end{pmatrix}  \begin{pmatrix} 0_n \\ \eta_t\end{pmatrix}^\top\hat{v}^{\perp}}\]
denote the variance of $\inner{\hat{v}^{\perp}}{\tilde{\eta}_t^\perp}$. Write $\hat{v}^{\perp} $ as  $\hat{v}^{\perp} = [(\hat{v}^{\perp}_1)^\top  \ (\hat{v}^{\perp}_2)^\top]^\top $, where $\hat{v}^{\perp}_1 \in \mathbb{R}^{n},  \hat{v}^{\perp}_2 \in \mathbb{R}^{d}$, and $\norm{\hat{v}^{\perp}_1} + \norm{\hat{v}^{\perp}_2} = 1 $. Since $\hat{v}^{\perp}$ is a unit vector in the orthogonal space of the columns space of $[I_n \ K^\top]^\top$, we must have $\hat{v}^{\perp}_1 + K^\top \hat{v}^{\perp}_2 = 0$. Then $\norm{\hat{v}^{\perp}_1}  = \norm {-K^\top \hat{v}^{\perp}_2 }\leq \norm{K} \norm{\hat{v}^{\perp}_2}   $ and hence $\norm{\hat{v}^{\perp}_2} \geq \frac{1}{1+ \norm{K}  }$. We have $\sigma^2_{\hat{v}^{\perp}} = \expct{(\hat{v}^{\perp}_2) ^\top \eta_t \eta_t^\top \hat{v}^{\perp}_2}  = (\hat{v}^{\perp}_2) ^\top I_d \hat{v}^{\perp}_2  = \norm{\hat{v}^{\perp}_2}^2\geq \frac{1}{(1 + \norm{K})^2}$. Also, $\sigma^2_{\hat{v}^{\perp}} \leq 1$.

Applying a supermartingale argument, we get
\#\label{eq:paraller_perp_interaction_term}
\prob{ \left| \sum_{t} \inner{\hat{v}^\parallel}{z_t^\parallel}\inner{\hat{v}^\perp}{\xi_t^\perp}\right| \geq  \sigma_{\hat{v}^{\perp}}\sigma_L \sqrt{2 \ln \frac{1}{\delta}\sum_{t}  \inner{\hat{v}^\parallel}{z_t^\parallel}^2 }} &\leq 2\delta.
\#

Next, we lower bound the denominator.

{\bf Step 2:  Lower bound for $ \sum_t \inner{\hat{v}^\parallel}{z_t^\parallel}^2$.} By direct computation, 
\*
\sum_t \inner{\hat{v}^\parallel}{z_t^\parallel}^2  &= \inner{\hat{v}^\parallel}{M \Theta_{t-1}  z_{t- 1} +  Mw_{t-1}  + \xi_t^{\parallel}}^2  \\
&\geq \sum_t \inner{\hat{v}^\parallel}{Mw_{t-1} } ^2 
+ 2\inner{\hat{v}^\parallel}{Mw_{t-1} } \inner{\hat{v}^\parallel}{M \Theta_{t-1}  z_{t- 1}} \\
 &\quad +  2\inner{\hat{v}^\parallel}{Mw_{t-1} } \inner{\hat{v}^\parallel}{\xi_t^\parallel }   + 
2\inner{\hat{v}^\parallel}{\xi_t^\parallel }\inner{\hat{v}^\parallel}{M \Theta_{t-1}  z_{t- 1} } .
\*
Let $\sigma^2_1 = (\hat{v}^\parallel) ^\top M W M^\top \hat{v}^\parallel$ denote the variance of $\inner{\hat{v}^\parallel}{M w_{t-1}}$. Write $  \hat{v}^\parallel = M x_v $, where $1 = \norm{\hat{v}^\parallel}^2 = \norm{[x_v^\top \ x_v^\top K^\top  ]}^2  = \norm{x_v}^2 + \norm{K x_v}^2  $. 
 Recalling that $W  \succcurlyeq \psi^{2} I_{n}$, we have 
 \*
\sigma^2_1 =(\hat{v}^\parallel) ^\top M W M^\top \hat{v}^\parallel &\geq \psi^2 \cdot  x_v^\top M^\top M M ^\top M x_v \\
& =  \psi^{2} \cdot  x_v^\top(I + K^\top K ) (I + K^\top K )   x_v\\
 & =  \psi^{2}  \left (x_v^\top x_v  + 2 x_v^\top K^\top Kx_v + x_v^\top  K^\top K  K^\top K   x_v  \right ) \\
 & =  \psi^{2} \left ( \norm{x_v}^2 + 2\norm{ Kx_v}^2 + \norm{ K^\top K   x_v} ^2   \right)\\
 & \geq \psi^{2} (\norm{x_v}^2  + \norm{ Kx_v}^2)\\
 & =\psi^{2}.
 \*
 We also have $W \preccurlyeq \Psi^{2} I_{n}$. Then
  \*
\sigma^2_1 = (\hat{v}^\parallel) ^\top M W M^\top \hat{v}^\parallel &\leq \Psi^2  \cdot x_v^\top M^\top M M ^\top M x_v \\
& =  \Psi^{2}  \cdot  x_v^\top(I + K^\top K ) (I + K^\top K )   x_v\\
 & =  \Psi^{2}  \left (x_v^\top x_v  + 2 x_v^\top K^\top Kx_v + x_v^\top  K^\top K  K^\top K   x_v  \right ) \\
 & =  \Psi^{2} \left ( \norm{x_v}^2 + 2\norm{ Kx_v}^2 + \norm{ K^\top K   x_v} ^2   \right)\\
& \leq \Psi^{2}  \left ( \norm{x_v}^2 + \norm{ Kx_v}^2 + \norm{K}^2\norm{ x_v}^2 + \norm{K^\top}^2 \norm{  K   x_v} ^2   \right)\\
&\leq \Psi^{2}(1+\norm{K}^2).
 \*
By standard Laurent-Massart bounds, we get
\#
\prob{\sum_t \inner{\hat{v}^\parallel}{Mw_{t-1} } ^2  \geq  \Psi^2(1+\norm{K}^2) \left(|\cI| + 2 \sqrt{|\cI| \ln(\frac{1}{\delta})}+2  \ln(\frac{1}{\delta})\right)} \leq \delta,\label{eq:LM_bound_1}
\\
\prob{\sum_t \inner{\hat{v}^\parallel}{Mw_{t-1} } ^2  \leq  \psi^2 \left(|\cI| - 2 \sqrt{|\cI| \ln(\frac{1}{\delta})}\right)} \leq \delta.\label{eq:LM_bound_2}
\#
Note that
\*
\sigma_L^2 \sum_t \inner{\hat{v}^{\parallel}}{\tilde{\eta}_t^\parallel } ^2  \geq \sum_t \inner{\hat{v}^{\parallel}}{\xi_t^\parallel } ^2 = \sum_t \sigma_t^2 \inner{\hat{v}^{\parallel}}{\tilde{\eta}_t^\parallel } ^2 \geq \sigma_\cI^2 \sum_t \inner{\hat{v}^{\parallel}}{\tilde{\eta}_t^\parallel } ^2 .
\*
Let  $\sigma^2_{\hat{v}^{\parallel}} = \expct{(\hat{v}^{\parallel}) ^\top \begin{pmatrix} 0_n \\ \eta_t\end{pmatrix}  \begin{pmatrix} 0_n \\ \eta_t\end{pmatrix}^\top\hat{v}^{\parallel}}$ denote the variance of $\inner{\hat{v}^{\parallel}}{\tilde{\eta}_t^\perp}$. Write $\hat{v}^{\parallel} = [\hat{v}^{\parallel}_1 \ \hat{v}^{\parallel}_2  ] $, where $\hat{v}^{\parallel}_1 \in \mathbb{R}^n, \hat{v}^{\parallel}_2 \in \mathbb{R}^d$ and $\norm{\hat{v}^{\parallel}_1}, \norm{\hat{v}^{\parallel}_2}\leq 1 $. We have  $\sigma^2_{\hat{v}^{\parallel}} = \expct{(\hat{v}^{\parallel}_2) ^\top  \eta_t \eta_t^\top \hat{v}^{\parallel}_2} = (\hat{v}^{\parallel}_2) ^\top  I_d   (\hat{v}^{\parallel}_2)  = \norm{\hat{v}^{\parallel}_2}^2 \leq 1$. 

Again, by standard Laurent-Massart bounds, we have 
\*
\prob{\sum_t \inner{\hat{v}^{\parallel}}{\xi_t^\parallel } ^2  \geq \sigma_L^2} \left(|\cI| + 2 \sqrt{|\cI| \ln(\frac{1}{\delta})}+2  \ln(\frac{1}{\delta})\right) \leq \delta,
\\
\prob{\sum_t \inner{\hat{v}^{\parallel}}{\xi_t^\parallel } ^2  \leq  \sigma_\cI^2 \sigma^2_{\hat{v}^{\parallel}} \left(|\cI| - 2 \sqrt{|\cI| \ln(\frac{1}{\delta})}\right)} \leq \delta.
\*
Applying a supermartingale argument, we get
\#
\prob{ \left| \sum_{t} \inner{\hat{v}^\parallel}{Mw_{t-1}}\inner{\hat{v}^\parallel}{M\Theta_{t-1}z_{t-1}}\right| \geq  \Psi(1+\norm{K}^2)^{\frac{1}{2}} \sqrt{2 \ln \frac{1}{\delta}\sum_{t}  \inner{\hat{v}^\parallel}{M\Theta_{t-1}z_{t-1}}^2 }} &\leq 2\delta,\label{eq:martingale_term1}\\
\prob{ \left| \sum_{t} \inner{\hat{v}^\parallel}{\xi_t^\parallel }\inner{\hat{v}^\parallel}{M\Theta_{t-1}z_{t-1}}\right| \geq \sigma_{L} \sqrt{2 \ln \frac{1}{\delta}\sum_{t}  \inner{\hat{v}^\parallel}{M\Theta_{t-1}z_{t-1}}^2 }} &\leq 2\delta,\label{eq:martingale_term2}\\
\prob{ \left| \sum_{t} \inner{\hat{v}^\parallel}{Mw_{t-1} } \inner{\hat{v}^\parallel}{\xi_t^\parallel }  \right| \geq  \sigma_{L} \sqrt{2 \ln \frac{1}{\delta}\sum_{t}  \inner{\hat{v}^\parallel}{M w_{t - 1 }}^2 }} &\leq 2\delta. \label{eq:martingale_term3}
\#
Combining \eqref{eq:LM_bound_1} and \eqref{eq:martingale_term3},  we have
\*
\prob{ \left| \sum_{t} \inner{\hat{v}^\parallel}{Mw_{t-1} } \inner{\hat{v}^\parallel}{\xi_t^\parallel }  \right| \geq  \sigma_{L}  \Psi(1+\norm{K}^2)^{\frac{1}{2}}\sqrt{2 \ln \frac{1}{\delta} \left(|\cI| + 2 \sqrt{|\cI| \ln(\frac{1}{\delta})}+2  \ln(\frac{1}{\delta})\right)}} &\leq 3\delta.
\*
 Combining the inequalities above, it holds with probability at least $1-8\delta$ that
 \#
 \sum_t \inner{\hat{v}^\parallel}{z_t^\parallel}^2 &\geq \sum_t \inner{\hat{v}^\parallel}{Mw_{t-1} } ^2 
+ 2\inner{\hat{v}^\parallel}{Mw_{t-1} } \inner{\hat{v}^\parallel}{M \Theta_{t-1}  z_{t- 1}} \nonumber\\
 &\quad +  2\inner{\hat{v}^\parallel}{Mw_{t-1} } \inner{\hat{v}^\parallel}{\xi_t^\parallel }   + 
2\inner{\hat{v}^\parallel}{\xi_t^\parallel }\inner{\hat{v}^\parallel}{M \Theta_{t-1}  z_{t- 1} }\nonumber \\
& \hspace{-0.2in} \geq  \psi^2 \left(|\cI| - 2 \sqrt{|\cI| \ln(\frac{1}{\delta})}\right)  - 2\Psi(1+\norm{K}^2)^{\frac{1}{2}} \sqrt{2 \ln \frac{1}{\delta}\sum_{t}  \inner{\hat{v}^\parallel}{M\Theta_{t-1}z_{t-1}}^2 }  \nonumber\\
& \hspace{-0.2in} \quad - 2  \sigma_{L}\sqrt{2 \ln \frac{1}{\delta}\sum_{t}  \inner{\hat{v}^\parallel}{M\Theta_{t-1}z_{t-1}}^2 } \nonumber\\
& \hspace{-0.2in} \quad - 2 \sigma_{L}  \Psi(1+\norm{K}^2)^{\frac{1}{2}}\sqrt{2 \ln \frac{1}{\delta} \left(|\cI| + 2 \sqrt{|\cI| \ln(\frac{1}{\delta})}+2  \ln(\frac{1}{\delta})\right)} \nonumber\\
& \hspace{-0.2in} \geq  \psi^2 \left(|\cI| - 2 \sqrt{|\cI| \ln(\frac{1}{\delta})}\right)  - 2\Psi(1+\norm{K}^2)^{\frac{1}{2}} \norm{M} \Theta_u z_u  |\cI|^{\frac{1}{2}} \sqrt{2 \ln \frac{1}{\delta}} \nonumber\\
& \hspace{-0.2in} \quad - 2 \sigma_L\norm{M} \Theta_u z_u  |\cI|^{\frac{1}{2}}  \sqrt{2 \ln \frac{1}{\delta}}  \nonumber\\
& \hspace{-0.2in} \quad - 2 \sigma_{L}  \Psi(1+\norm{K}^2)^{\frac{1}{2}}\sqrt{2 \ln \frac{1}{\delta} \left(|\cI| + 2 \sqrt{|\cI| \ln(\frac{1}{\delta})}+2  \ln(\frac{1}{\delta})\right)}\nonumber\\
& \hspace{-0.2in} \geq  \psi^2 \left(|\cI| - 2 \sqrt{|\cI| \ln(\frac{1}{\delta})}\right)  - 2\Psi(1+\norm{K}^2)^{\frac{1}{2}} \norm{M} \Theta_u z_u  |\cI|^{\frac{1}{2}} \sqrt{2 \ln \frac{1}{\delta}} \nonumber\\
& \hspace{-0.2in} \quad - 2 \sigma_L\norm{M} \Theta_u z_u  |\cI|^{\frac{1}{2}}  \sqrt{2 \ln \frac{1}{\delta}}  - 2 \sigma_{L}  \Psi(1+\norm{K}^2)^{\frac{1}{2}}\sqrt{2 \ln \frac{1}{\delta} 2I}\nonumber\\
& \hspace{-0.2in}  =  \psi^2  |\cI| - \sqrt{2|\cI| \ln \frac{1}{\delta} } \left( 2 \psi^2  + 2\Psi(1+\norm{K}^2)^{\frac{1}{2}} \norm{M} \Theta_u z_u \right) \nonumber\\
& \hspace{-0.2in} \qquad\qquad\qquad \left. + 2 \sigma_L\norm{M} \Theta_u z_u + 2\sqrt{2} \sigma_{L}  \Psi(1+\norm{K}^2)^{\frac{1}{2}}  \right) \eqqcolon \Lambda_1 .
\label{eq:parallel_denominator_bound1}
 \#
In arriving at \eqref{eq:parallel_denominator_bound1} we have used the assumption $|\cI| \geq 16 \ln \frac{1}{\delta}$, which implies 
\[ 2 \ln \frac{1}{\delta} \left(|\cI| + 2 \sqrt{|\cI| \ln(\frac{1}{\delta})}+2  \ln(\frac{1}{\delta})\right)\leq 2|\cI|.\]
 To get a further cleaner expression, we further assume
 \#
 \label{eq:cond_L_1}
 |\cI|\geq 32 \psi^{-4} \ln \frac{1}{\delta} \left(  \psi^2  + \Psi(1+\norm{K}^2)^{\frac{1}{2}} \norm{M} \Theta_u z_u +  \sigma_L\norm{M} \Theta_u z_u + \sqrt{2} \sigma_{L}  \Psi(1+\norm{K}^2)^{\frac{1}{2}}  \right)^2,
 \#
 which in turn implies $|\cI| \geq 16 \ln \frac{1}{\delta}$, under which the bound simplifies to
 \#
 \label{eq:parallel_denominator_bound}
 \Lambda_1 & \geq \frac{\psi^2}{2} |\cI|.
 \#
Combining \eqref{eq:paraller_perp_interaction_term} and \eqref{eq:parallel_denominator_bound1}, it holds with probability at least $1- 10\delta$ that 
\#
\frac{\sum_t \left|\inner{\hat{v}^\parallel}{z_t^\parallel}\inner{\hat{v}^\perp}{\xi_t^\perp}\right|}{\sum_t \inner{\hat{v}^\parallel}{z_t^\parallel}^2}\leq \frac{\sigma_{\hat{v}^{\perp}}\sigma_L \sqrt{2 \ln \frac{1}{\delta} }}{\sqrt{\sum_t \inner{\hat{v}^\parallel}{z_t^\parallel}^2}} \leq \sigma_L 2\sqrt{\ln \frac{1}{\delta} } \psi^{-1} |\cI|^{-\frac{1}{2}} \leq \frac{1}{33},
\#
provided that 
\# 
\label{eq:cond_L_2}
|\cI|& \geq 4 \cdot 33^2  \sigma_L^2  \ln \frac{1}{\delta}  \psi^{-2}.
\#

{\bf Step 3: Lower bound on $\sum_t \inner{\hat{v}^\perp}{\xi_t^\perp}^2$.}  Recall that $\xi_t = \sigma_t \tilde{\eta}_t$ and $ \sigma_t \leq \sigma_L$.  Let 
\[ \sigma^2_{\hat{v}^{\perp}} = \expct{(\hat{v}^{\perp}) ^\top \begin{pmatrix} 0_n \\ \eta_t\end{pmatrix}  \begin{pmatrix} 0_n \\ \eta_t\end{pmatrix}^\top\hat{v}^{\perp}}\]
denote the variance of $\inner{\hat{v}^{\perp}}{\tilde{\eta}_t^\perp}$.
Write $\hat{v}^{\perp} $ as  $\hat{v}^{\perp} = [(\hat{v}^{\perp}_1)^\top  \ (\hat{v}^{\perp}_2)^\top]^\top $, where $\hat{v}^{\perp}_1 \in \mathbb{R}^{n},  \hat{v}^{\perp}_2 \in \mathbb{R}^{d}$ and $\norm{\hat{v}^{\perp}_1} + \norm{\hat{v}^{\perp}_2} = 1 $. Since $\hat{v}^{\perp}$ is a unit vector in the orthogonal space of the columns space of $[I_n \ K^\top]^\top$, we must have $\hat{v}^{\perp}_1 + K^\top \hat{v}^{\perp}_2 = 0$. Then $\norm{\hat{v}^{\perp}_1}  = \norm {-K^\top \hat{v}^{\perp}_2 }\leq \norm{K} \norm{\hat{v}^{\perp}_2}   $ and hence $\norm{\hat{v}^{\perp}_2} \geq \frac{1}{1+ \norm{K}  }$. We have $\sigma^2_{\hat{v}^{\perp}} = \expct{(\hat{v}^{\perp}_2) ^\top \eta_t \eta_t^\top \hat{v}^{\perp}_2}  = (\hat{v}^{\perp}_2) ^\top I_d \hat{v}^{\perp}_2  = \norm{\hat{v}^{\perp}_2}^2\geq \frac{1}{(1 + \norm{K})^2}$. Also, $\sigma^2_{\hat{v}^{\perp}} \leq 1$.

By standard Laurent-Massart bounds, the denominator can be bounded from below by 
\*
&\prob{\sum_t \inner{\hat{v}^{\perp}}{\xi_t^\perp} ^2  \leq  \frac{\sigma_\cI^2}{(1 + \norm{K} )^2}   \left(|\cI| - 2 \sqrt{|\cI| \ln(\frac{1}{\delta})}\right)} \leq \delta.
\*
Plugging in our choice of $\sigma_\cI^2$,  with probability at least $1-2\delta$ we have 
\*
\sum_t \inner{\hat{v}^{\perp}}{\xi_t^\perp} ^2   \geq  \frac{1}{(1 + \norm{K} )^2} \sqrt{\frac{C_0}{|\cI|}} \left(|\cI| - 2 \sqrt{|\cI| \ln(\frac{1}{\delta})}\right) = \frac{1}{(1 + \norm{K} )^2} C_0^{\frac{1}{2}}   (|\cI|^{\frac{1}{2}} - 2 \sqrt{ \ln \frac{1}{\delta}})\eqqcolon \Lambda_2.
\*
Under the assumption that $|\cI| \geq 16 \ln \frac{1}{\delta}$, we get:
\#
\label{eq:perp_denominator_bound}
\Lambda_2 &\geq  \frac{1}{2} |I|^{\frac{1}{2}} \frac{1}{(1+\norm{K})^2} C_0^{\frac{1}{2}}.
\#
Combining with \eqref{eq:paraller_perp_interaction_term}, it holds with probability at least $1- 4\delta$ that 
\*
\frac{\sum_t \left|\inner{\hat{v}^\parallel}{z_t^\parallel}\inner{\hat{v}^\perp}{\xi_t^\perp}\right|}{\sum_t \inner{\hat{v}^{\perp}}{\xi_t^\perp}^2}&\leq \frac{\sigma_{\hat{v}^{\perp}}\sigma_L \sqrt{2 \ln \frac{1}{\delta}\sum_{t}  \inner{\hat{v}^\parallel}{z_t^\parallel}^2 }}{\sum_t \inner{\hat{v}^{\perp}}{\xi_t^\perp}^2}\\
&\leq \frac{\sigma_{\hat{v}^{\perp}}\sigma_L \sqrt{2 \ln \frac{1}{\delta}\sum_{t}  \norm{z_t^\parallel}^2 }}{\sum_t \inner{\hat{v}^{\perp}}{\xi_t^\perp}^2 }\\
&\leq \frac{\sigma_L \sqrt{2 \ln \frac{1}{\delta}  }|\cI|^{\frac{1}{2}}  z_u }{\sum_t \inner{\hat{v}^{\perp}}{\xi_t^\perp}^2 }\\
&\leq \frac{\sigma_L \sqrt{2 \ln \frac{1}{\delta}  }|\cI|^{\frac{1}{2}}  z_u }{\frac{1}{2} |\cI|^{\frac{1}{2}} \frac{1}{(1+\norm{K})^2} C_0^{\frac{1}{2}} }\\
& = 2\sqrt{2} \sqrt{\ln \frac{1}{\delta} } \sigma_L z_u (1+\norm{K})^2 C_0^{- \frac{1}{2}}\\
& = 2\sqrt{2} \sqrt{\ln \frac{1}{\delta} }z_u (1+\norm{K})^2 L^{-\frac{1}{2}}\\
& \leq \frac{1}{33},
\*
provided $L$ satisfies:
\#
\label{eq:cond_L_3}
L &\geq 66^2 \cdot 2 \cdot \ln \frac{1}{\delta} \cdot z_u^2 (1+\norm{K})^4.
\#
\end{proof}


\textbf{Step 2: Bounding $\lambda_v$ when $v = \hat{v}^\parallel$.} 
Noting that $z_t^\parallel = y_t + \xi_t^\parallel$, we can rewrite the left hand side of  \eqref{eqn:ols_1d_foc} in this case as $\lambda_{\hat{v}^\parallel} \sum_{t} \inner{v}{z_t^\parallel}^2  $,
and the right hand side of \eqref{eqn:ols_1d_foc} as
\begin{align*}
    & \sum_t \omega_t \inner{v}{z_t^\parallel} + \sum_t \inner{\theta_t - \bar{\theta}}{z_t} \cdot \inner{v}{z_t^\parallel} .
\end{align*}
Therefore,
\#\label{eq:lambda_parallel}
|\lambda_{\hat{v}^\parallel}| & \leq  \frac{\left| \sum_t \inner{\theta_t - \bar{\theta}}{z_t} \cdot \inner{v}{z_t^\parallel}  \right|}{\sum_{t}  \inner{v}{z_t^\parallel}^2 } +  \frac{ \left| \sum_t \omega_t \inner{v}{z_t^\parallel} \right|}{ \sum_{t}  \inner{v}{z_t^\parallel}^2 } .
\#
By \eqref{eq:parallel_denominator_bound1} and \eqref{eq:parallel_denominator_bound}  it holds with probability at least $1-8\delta$ that
\*
 \sum_t \inner{\hat{v}^\parallel}{z_t^\parallel}^2 \geq  \Lambda_1 \geq \frac{\psi^2}{2} |\cI|, 
 \*
 if we have 
 \*
 |\cI|\geq 32 \psi^{-4} \ln \frac{1}{\delta} \left(  \psi^2  + \Psi(1+\norm{K}^2)^{\frac{1}{2}} \norm{M} \Theta_u z_u +  \sigma_L\norm{M} \Theta_u z_u + \sqrt{2} \sigma_{L}  \Psi(1+\norm{K}^2)^{\frac{1}{2}}  \right)^2.
 \*
 
It remains to upper bound the numerators of the two terms in \eqref{eq:lambda_parallel}. 
For the first term, Cauchy-Schwartz inequality gives:
\#\label{eq:parallel_bias_numerator}
\left| \sum_t \inner{\theta_t - \bar{\theta}}{z_t} \cdot \inner{v}{z_t^\parallel}  \right|
& \leq  \sqrt{  \sum_t \inner{\theta_t - \bar{\theta}}{z_t}^2} \sqrt{ \sum_t \inner{v}{z_t^\parallel}^2 } \nonumber \\
&\leq \max_t \left|\theta_t - \bar{\theta} \right| \cdot \sqrt{\sum_t \norm{z_t}^2}  \sqrt{ \sum_t \inner{v}{z_t^\parallel}^2 }\nonumber\\
& \leq \Delta_\cI \sqrt{z_u^2 |\cI|} \sqrt{ \sum_t \inner{v}{z_t^\parallel}^2 } \nonumber\\
&\leq  \Delta_\cI z_u |\cI|^{\frac{1}{2}} \sqrt{ \sum_t \inner{v}{z_t^\parallel}^2 }. 
\#
Plugging this into \eqref{eq:lambda_parallel} gives
\#\label{eq:parallel_variance_numerator}
\frac{\left| \sum_t \inner{\theta_t - \bar{\theta}}{z_t} \cdot \inner{v}{z_t^\parallel}  \right|}{\sum_{t}  \inner{v}{z_t^\parallel}^2 } & \leq \frac{\Delta_\cI z_u |\cI|^{\frac{1}{2}} }{\sqrt{\sum_{t}  \inner{v}{z_t^\parallel}^2}}\leq\Delta_\cI z_u |\cI|^{\frac{1}{2}}\Lambda_1^{-\frac{1}{2}}  .
\# 
For the second term, let $\omega_t \stackrel{d}{=}\mathcal{N}(0,\psi_i^2)$. Note that assuming $w_t(i)$ is $\psi_i^2$ sub-Gaussian suffices.  By the assumption on $w_t$,  a standard supermartingale argument implies that
\[ \prob{ \left| \sum_{t} \omega_t \inner{v}{z_t^\parallel} \right| \geq  \psi_i \sqrt{2 \ln \frac{1}{\delta}\sum_{t}  \inner{v}{z_t^\parallel}^2 }} \leq 2\delta. \]
Plugging this into \eqref{eq:lambda_parallel} gives
\#\label{eq:parallel_variance_numerator}
\frac{ \left| \sum_t \omega_t \inner{v}{z_t^\parallel} \right|}{ \sum_{t}  \inner{v}{z_t^\parallel}^2 } & \leq \frac{ \psi_i \sqrt{2 \ln \frac{1}{\delta} }}{\sqrt{\sum_{t}  \inner{v}{z_t^\parallel}^2}}\leq \psi_i \sqrt{2 \ln \frac{1}{\delta}}\Lambda_1^{-\frac{1}{2}}.
\#

Finally, we conclude  it holds with probability at least $1-10\delta$ that  
\#\label{eq:parallel_lambda_bound}
|\lambda_{\hat{v}^\parallel}|   &\leq  \Delta_\cI z_u |\cI|^{\frac{1}{2}}\Lambda^{-\frac{1}{2}}_1 + \psi_i \sqrt{2 \ln \frac{1}{\delta}}\Lambda_1^{-\frac{1}{2}} \nonumber \\
& \leq  \sqrt{2} \psi^{-1}  \Delta_\cI z_u  + 2 \psi^{-1}  \psi_i \sqrt{ \ln \frac{1}{\delta}} |\cI|^{-\frac{1}{2}} .
\#

\textbf{Step 3: Bounding $\lambda_v$ when $v = \hat{v}^\perp$.} 
Noting that $z_t^\perp = \xi_t^\perp$, we can rewrite the left hand side of  \eqref{eqn:ols_1d_foc} as $
    \lambda_{\hat{v}^\perp} \sum_{t}  \inner{v}{\xi_t^\perp}^2$,
and the right hand side of \eqref{eqn:ols_1d_foc} as
\begin{align*}
    & \sum_t \omega_t \inner{v}{\xi_t^\perp} +  \sum_t \inner{\theta_t - \bar{\theta}}{z_t} \cdot \inner{v}{\xi_t^\perp}.
\end{align*}
Therefore,
\#\label{eq:lambda_perp} 
    |\lambda_{\hat{v}^\perp}| & \leq   \frac{\left| \sum_t \inner{\theta_t - \bar{\theta}}{z_t} \cdot \inner{v}{\xi_t^\perp} \right|}{\sum_{t}  \inner{v}{\xi_t^\perp}^2} +\frac{\left| \sum_t \omega_t \inner{v}{\xi_t^\perp}  \right|}{\sum_{t}  \inner{v}{\xi_t^\perp}^2 }.
\#

For the first term, we observe that $\xi_t^\perp$ is normally distributed and is independent of $z_t$. Applying a supermartingale argument, we get
\begin{align*}
\prob{ \left| \sum_{t} \inner{\theta_t-\bar{\theta}}{z_t}\cdot \inner{v}{\xi_t^\perp} \right| \geq  \sigma_L \sqrt{2 \ln \frac{1}{\delta}\sum_{t}  \inner{\theta_t - \theta}{z_t}^2 }} &\leq 2\delta,
\end{align*}
and hence 
\*
\prob{ \left| \sum_{t} \inner{\theta_t-\bar{\theta}}{z_t}\cdot \inner{v}{\xi_t^\perp} \right| \geq  \sigma_L \Delta_\cI z_u|\cI|^{\frac{1}{2}} \sqrt{2 \ln \frac{1}{\delta} }} &\leq 3\delta.
\*
Then the first term is upper bounded as 
\*
\frac{\left| \sum_t \inner{\theta_t - \bar{\theta}}{z_t} \cdot \inner{v}{\xi_t^\perp} \right|}{\sum_{t}  \inner{v}{\xi_t^\perp}^2} \leq \frac{\sqrt{\sum_t \inner{\theta_t - \bar{\theta}}{z_t}^2}\sqrt{\sum_t \inner{v}{\xi_t^\perp}^2 }}{\sum_{t}  \inner{v}{\xi_t^\perp}^2} \leq \sigma_L  \Delta_\cI z_u |\cI|^{\frac{1}{2}}\sqrt{2 \ln \frac{1}{\delta}    }\Lambda_2^{-1}.
\*

For the second term, a supermartingale argument implies that
\[ \prob{ \left| \sum_{t} \omega_t \inner{v}{\xi_t^\perp} \right| \geq  \psi_i \sqrt{2 \ln \frac{1}{\delta}\sum_{t} \inner{v}{\xi_t^\perp}^2 }} \leq 2\delta. \]
Then 
\*
\frac{\left| \sum_t \omega_t \inner{v}{\xi_t^\perp}  \right|}{\sum_{t}  \inner{v}{\xi_t^\perp}^2 } \leq \frac{\psi_i \sqrt{2 \ln \frac{1}{\delta}\sum_{t} \inner{v}{\xi_t^\perp}^2}}{\sum_{t}   \inner{v}{\xi_t^\perp}^2} \leq  \psi_i \sqrt{2 \ln \frac{1}{\delta}}  \Lambda_2^{ - \frac{1}{2}} .
\*

Finally, we conclude  it holds with probability at least $1-6\delta$ that
\*
    |\lambda_{\hat{v}^\perp}| & \leq \frac{\left| \sum_t \inner{\theta_t - \bar{\theta}}{z_t} \cdot \inner{v}{\xi_t^\perp} \right|}{\sum_{t}  \inner{v}{\xi_t^\perp}^2} + \frac{\left| \sum_t \omega_t \inner{v}{\xi_t^\perp}  \right|}{\sum_{t}  \inner{v}{\xi_t^\perp}^2 }\nonumber\\
    &\leq   \sigma_L  \Delta_\cI z_u |\cI|^{\frac{1}{2}}\sqrt{2 \ln \frac{1}{\delta}    }\Lambda_2^{-1} + \psi_i \sqrt{2 \ln \frac{1}{\delta}}  \Lambda_2^{ - \frac{1}{2}} .
\*
Assuming $|\cI|\geq 16 \ln \frac{1}{\delta}$ and using the bound on $\Lambda_2$ from \eqref{eq:perp_denominator_bound}, we have
\#
\label{eq:perp_lambda_bound}
|\lambda_{\hat{v}^\perp}| 
    &\leq   2\sigma_L  \Delta_\cI z_u \sqrt{2 \ln \frac{1}{\delta}    } (1+ \norm{K})^2 C_0^{-\frac{1}{2}}  
    +2 \psi_i \sqrt{ \ln \frac{1}{\delta}}  (1+\norm{K})C_0^{-\frac{1}{4}} |\cI|^{-\frac{1}{4}} .
\#
Combining  Lemma~\ref{lemma:quadratic_lemma_final}, \eqref{eq:parallel_lambda_bound}  and \eqref{eq:perp_lambda_bound}, we have that
\*
\lambda_v^2 &\leq 2\lambda_{\hat{v}^\parallel}^2 +2 \lambda_{\hat{v}^\perp}^2\\
&\leq  4 \Delta_\cI^2 z_u^2|\cI|\Lambda_1^{-1} +  8 \psi_i ^2  \ln \frac{1}{\delta} \Lambda_1^{-1} +  8  \sigma_L^2 \Delta_\cI^2 (1+\norm{K})^2 x_u^2  \ln\frac{1}{\delta} |\cI|\Lambda_2^{-2} +  8\psi_i^2  \ln \frac{1}{\delta} \Lambda_2^{ - 1}  \\
& =  4 \Delta_\cI^2 z_u^2 \left( |\cI|\Lambda_1^{-1} + 2  C_0^{\frac{1}{2}}   \ln \frac{1}{\delta}    |\cI|^{\frac{1}{2}} \Lambda_2^{-2} \right) +8 \psi_i ^2  \ln \frac{1}{\delta} \left( \Lambda_1^{-1} + \Lambda_2^{ - 1} \right) \\
& =  4  \Delta_\cI^2 z_u^2 |\cI|\left( \Lambda_1^{-1} + 2  \sigma_L^2     \ln\frac{1}{\delta} \Lambda_2^{-2} \right) +8 \psi_i ^2  \ln \frac{1}{\delta} \left( \Lambda_1^{-1} + \Lambda_2^{ - 1} \right) \\ 
&\leq 8\psi^{-2}\Delta_\cI^2 z_u^2 + 16\psi^{-2} \psi_i^2 \ln \frac{1}{\delta} |\cI|^{-1} + 32\sigma_L^2  \Delta_\cI^2 z_u^2  \ln \frac{1}{\delta}     (1+ \norm{K})^4 C_0^{-1}  
    +16 \psi_i^2  \ln \frac{1}{\delta}  (1+\norm{K})^2 C_0^{-\frac{1}{2}} |\cI|^{-\frac{1}{2}} \\
 &= 8 \Delta_\cI^2 z_u^2 \left( \psi^{-2} + 4\sigma_L^2 \ln \frac{1}{\delta} (1+\norm{K})^4 C_0^{-1} \right) + 16 \psi_i^2 \ln \frac{1}{\delta} \left(   \psi^{-2} |\cI|^{-1}  +    (1+\norm{K})^2 C_0^{-\frac{1}{2}} |\cI|^{-\frac{1}{2}}  \right)   
\*
holds with probability at least $1 -27\delta$.

Hence, we conclude  that
\#\label{eq:lambda_final_bound_vector}
\nonumber
|\lambda_v| & \leq 2\sqrt{2} \Delta_\cI z_u \sqrt{ \psi^{-2} + 4\sigma_L^2 \ln \frac{1}{\delta} (1+\norm{K})^4 C_0^{-1}}  + 4\psi_i \sqrt{\ln \frac{1}{\delta}}\sqrt{   \psi^{-2} |\cI|^{-1}  +    (1+\norm{K})^2 C_0^{-\frac{1}{2}} |\cI|^{-\frac{1}{2}} } \\
\nonumber
&\leq   2\sqrt{2} \Delta_\cI z_u \sqrt{ \psi^{-2} + 4\sigma_L^2 \ln \frac{1}{\delta} (1+\norm{K})^4 C_0^{-1}}  + 4\psi_i \sqrt{\ln \frac{1}{\delta}}\sqrt{   \psi^{-2}   +    (1+\norm{K})^2 C_0^{-\frac{1}{2}}  } |\cI|^{-\frac{1}{4}} \\ 
& \eqqcolon \tilde{C}_1 \Delta_\cI  + \tilde{C}_2 |\cI|^{- \frac{1}{4}},
\#
holds with probability at least $1 -27\delta$, where we define 
\# 
\label{eq:def_tilde_C}
\nonumber
\tilde{C}_1 &=  2\sqrt{2}  z_u \sqrt{ \psi^{-2} + 4\sigma_L^2 \ln \frac{1}{\delta} (1+\norm{K})^4 C_0^{-1}} ,  \\
\tilde{C}_2 &=  4\psi_i \sqrt{\ln \frac{1}{\delta}}\sqrt{   \psi^{-2}   +    (1+\norm{K})^2 C_0^{-\frac{1}{2}}  } .
\#

{\bf Step 4: An $\epsilon$-net argument.}
To summarize, thus far we have shown that for any fixed direction $v \in \RR^{n+d}$, and for any row $\theta_i$ of the parameter matrix $\Theta$, the minimizer of the one-dimensional quadratic loss function satisfies \eqref{eq:lambda_final_bound_vector}. We next invoke Lemma~\ref{lem:eps_net_helper}, which implies that if this statement holds for all $v$ in an $\epsilon$-net of the $(n+d)$-dimensional unit sphere (where $\epsilon$ depends on the condition number $\kappa_u$ of the Hessian $\sum_t z_t z_t^\top$ as $\epsilon \leq  \frac{1}{5(1+\kappa_u)}$), then the Frobenius norm of the OLS estimator $\hat{\theta}_i$ and $\bar{\theta}_i$ is upper bounded by $\frac{5}{3} \bar{\lambda}$. The bound on the condition number $\kappa_u$ of the Hessian is proved in Lemma~\ref{lemma:bounded_coundition_number}. We make this more formal next.

We fix $\varepsilon$ as the confidence level. First, substituting $\delta = \varepsilon/6$ in Lemma~\ref{lemma:bounded_coundition_number} gives that with probability at least $1-\varepsilon/6$, the condition number of the Hessian is bounded from above as $\kappa_{\cI} \leq C_9 \sqrt{|\cI|}$ provided $\varepsilon \leq 18/100$ and
\#
\label{eq:cond_L_4}
|\cI| & \geq \frac{2000}{9}\left(2 
(n+d) \log\frac{6}{\varepsilon} +(n+d) \log \frac{\bar{x}^2(1+\norm{K}^2)+\sigma_L^2}{{ \sigma}^2_\cI \min \left\{\frac{1}{2}, \frac{\psi^2}{{ \sigma}^2_L+2\norm{K}^2}\right\}}\right).
\#
We will thus choose $\epsilon = (5(1+C_9 \sqrt{I}))^{-1}$ in the $\epsilon$-net result of Lemma~\ref{lem:eps_net_helper} and Lemma~\ref{lem:cardinality_eps_net}. This gives an upper bound on the cardinality of the $\epsilon$-net of $\left(1 + \frac{4}{\epsilon} \right)^{n+d} \leq (10(1+C_9\sqrt{|\cI|}))^{n+d}$.

Applying \eqref{eq:lambda_final_bound_vector} by substituting $ \delta = \frac{\varepsilon}{54n (10(1+C_9\sqrt{|\cI|}))^{n+d}}$, it holds with  probability at least $1-\varepsilon$ that for every row of $\Theta$ we have
\*
\norm{\hat{\theta}_{\cI} - \bar{\theta}}_F & =  \norm{\hat{\theta}_{\cI} - \bar{\theta}} \ \leq \ 
\frac{5}{3}\left( \tilde{C}_1  \Delta_\cI  + \tilde{C}_2  |\cI|^{- \frac{1}{4}} \right).
\*
Combining $n$ rows, we have 
\*
\norm{\hat{\Theta}_{\cI} - \bar{\Theta}}_F 
& \leq \frac{5\sqrt{n}}{3}\left( \tilde{C}_1  \Delta_\cI  + \tilde{C}_2  |\cI|^{- \frac{1}{4}} \right) \\
& \leq \breve{C}_1 \Delta_\cI + \breve{C}_2 |\cI|^{- \frac{1}{4}},
\*
where 
\# 
\label{eq:def_breve_C}
\nonumber
\breve{C}_1 &=  5  z_u \sqrt{n} \left( \psi^{-1} + \sqrt{4\sigma_L^2  (1+\norm{K})^4 C_0^{-1}} \sqrt{\ln \frac{1}{\varepsilon} + \ln 54n + (n+d)\ln \left( 10 (1+C_9 \sqrt{|\cI|})\right)} \right),  \\
\breve{C}_2 &=  7 \psi_i \sqrt{n} \sqrt{   \psi^{-2}   +    (1+\norm{K})^2 C_0^{-\frac{1}{2}}} \sqrt{\ln \frac{1}{\varepsilon} + \ln 54n + (n+d)\ln \left( 10 (1+C_9 \sqrt{|\cI|})\right)} .
\#
Under our choice of $C_0 = \cO(\log T)$, $L = \cO(\log^3 T)$, $\sigma^2_L = \sqrt{C_0/L}$, $\varepsilon = \cO(T^{-3})$, and assuming that $T$ is large enough so that $L$ satisfies conditions \eqref{eq:cond_L_1}, \eqref{eq:cond_L_2}, \eqref{eq:cond_L_3}, and \eqref{eq:cond_L_4}, both $\breve{C_1},\breve{C_2}$ are $\cO(\sqrt{\log T})$.

\subsection{Proof of OLS Concentration for $\cB_{i,0}$ (Lemma~\ref{lemma:ols_esti_warmup}) }

To bound the estimation error of the OLS estimator of the warm-up block $\cI = \cB_{i,0}$, we look at the one dimensional problem \eqref{eqn:ols_1d_foc} again:
\*
\lambda_v \sum_{t} \inner{v}{z_t}^2 = \sum_t \omega_t \inner{v}{z_t} + \sum_t \inner{\theta_t - \bar{\theta}}{z_t} \cdot \inner{v}{z_t}. 
\*
Since we are using a sequence of sequentially strongly stable policies $\{K_t^{\stab}\}$ instead of a fixed policy, we do not have a fixed column space anymore. Nevertheless, the $O(1)$ exploration noise $\xi_t = \nu_0 \tilde{\eta}_t$  enables us to bound the error similar to the case $v = v^\parallel$. Recall that we set $\nu_0=1$ in Algorithm~\ref{alg:DYN-LQR}.

Let $M_t \coloneqq  \begin{bmatrix} I_n \\ K^\stab_t \end{bmatrix}$. Note that
\*
\sum_t \inner{v}{z_t}^2  &= \inner{v}{M_t \Theta_{t-1}  z_{t- 1} +  M_t w_{t-1}  + \xi_t}^2  \\
&\geq \sum_t \inner{v}{  M_t w_{t-1} + \xi_t } ^2 
+ 2\inner{v}{M_t w_{t-1} } \inner{v}{M_t \Theta_{t-1}  z_{t- 1}} \\
 &\quad +  2\inner{v}{M_t w_{t-1} } \inner{v}{\xi_t}   + 
2\inner{v}{\xi_t }\inner{v}{M_t \Theta_{t-1}  z_{t- 1} } .
\*
Let $v = v^\parallel_{t} + v^\perp_{t} $, where $v^\parallel_{t}$  is the projection of $v$ onto the column space generated by $M_{t}$. We have 
\*
 \inner{v}{  M_t w_{t-1} + \xi_t } ^2  \geq  \inner{v_t^\parallel}{M_t w_{t-1}} ^2 +\inner{v^\perp_t}{\tilde{\eta}_t}^2.
\*
 Let $\sigma^2_{1,t} = ({v}_t^\parallel) ^\top M_t W M_t^\top {v}_t^\parallel$ denote the variance of $\inner{v_t^\parallel}{M_t w_{t-1}}$. Write $  {v}^\parallel_t = M_t x_{v,t} $, where $ \norm{v_t^\parallel}^2 = \norm{[x_{v,t}^\top \ x_{v,t}^\top K_t^\top  ]}^2  = \norm{x_{v,t}}^2 + \norm{(K_t^\stab) x_{v,t}}^2  $. 
 Recall that $W  \succcurlyeq \psi^{2} I_{n}$. We have 
 \#
\sigma^2_{1,t} &=(v_t^\parallel) ^\top M_t W M_t^\top v_t^\parallel \nonumber\\
&\geq \psi^2 \cdot  x_{v,t}^\top M_t^\top M_t M_t ^\top M_t x_{v,t} \nonumber\\
& =  \psi^{2} \cdot  x_{v,t}^\top(I + (K_t^\stab)^\top (K_t^\stab) ) (I + (K_t^\stab)^\top (K_t^\stab) )   x_{v,t}\nonumber\\
 & =  \psi^{2}  \left (x_{v,t}^\top x_{v,t}  + 2 x_{v,t}^\top (K_t^\stab)^\top_t  (K_t^\stab) x_v + x_{v,t}^\top  (K_t^\stab)^\top (K_t^\stab)   (K_t^\stab)^\top (K_t^\stab)   x_{v,t}  \right )\nonumber \\
 & =  \psi^{2} \left ( \norm{x_{v,t}}^2 + 2\norm{ (K_t^\stab) x_{v,t}}^2 + \norm{ (K_t^\stab)^\top (K_t^\stab)   x_{v,t}} ^2   \right)\nonumber \\
 & \geq \psi^{2} (\norm{x_{v,t}}^2  + \norm{ (K_t^\stab) x_{v,t}}^2)\nonumber \\
 & =\psi^{2}\norm{v_t^\parallel}^2.\label{eq:sigma_lower_bound_1}
 \#
Let $\sigma^2_{2,t} $ denote the variance of $\inner{v^\perp_t}{\tilde{\eta}_t}$. Write $v_t^{\perp} $ as  $v_t^{\perp} = [(v_{t,1}^{\perp})^\top  \ (v_{t,2}^{\perp})^\top]^\top $, where $v_{t,1}^{\perp} \in \mathbb{R}^{n}$, $v_{t,2}^{\perp} \in \mathbb{R}^{d}$, and $\norm{v_{t,1}^{\perp}} + \norm{v_{t,2}^{\perp}} = \norm{v_{t}^{\perp}} $. Since $\hat{v}^{\perp}$ is in the orthogonal space of the columns space of $[I_n \ K^\top_t]^\top$, we must have $v_{t,1}^{\perp} + K_t^\top v_{t,2}^{\perp} = 0$. Then $\norm{v_{t,1}^{\perp}}  = \norm {-K^\top v_{t,2}^{\perp} }\leq \norm{K_t^\stab} \norm{v_{t,2}^{\perp}}   $ and hence $\norm{v_{t,2}^{\perp}} \geq \frac{1}{1+ \norm{K_t^\stab}  }$. We have 
\#\label{eq:sigma_lower_bound_2}
\sigma^2_{2,t} = \expct{(v_{t,2}^{\perp}) ^\top \eta_t \eta_t^\top v_{t,2}^{\perp}}  = (v_{t,2}^{\perp}) ^\top I_d v_{t,2}^{\perp}  = \norm{v_{t,2}^{\perp}}^2\geq \frac{1}{(1 + \norm{K_t^\stab})^2} \norm{v_{t}^{\perp}}^2\geq \frac{1}{(1 + {K_u})^2} \norm{v_{t}^{\perp}}^2.
\# 
Recall that $ \norm{v_{t}^{\parallel}}^2 +  \norm{v_{t}^{\perp}}^2  = 1$.  Combing \eqref{eq:sigma_lower_bound_1} and \eqref{eq:sigma_lower_bound_2}, we have 
\*
\expct{\inner{v}{\xi_t +  M_t w_{t-1}}^2} 
&\geq \expct{ \inner{v_t^\parallel}{M_t w_{t-1}} ^2 +\inner{v^\perp_t}{\xi_t}^2}\\
&\geq \psi^{2}\norm{v_t^\parallel}^2 + \frac{\nu_0^2 }{(1 + {K_u})^2} \norm{v_{t}^{\perp}}^2\\
&\geq \min\left\{\psi^{2},\frac{\nu_0^2}{(1 + {K_u})^2}\right\} \\
&\eqqcolon \sigma_v^2.
\*

Using the standard Laurent-Massart bound implies
\*
\prob{\sum_t\inner{v}{\xi_t +  M_t w_{t-1}}^2   \leq   \sigma^2_{v} \left(|\cI| - 2 \sqrt{|\cI| \ln(\frac{1}{\delta})}\right)} \leq \delta.
\*

Let $v = [v_1^\top \ v_2^\top]^\top$, where $v_1 \in\mathbb{R}^{n}$ and $v_2 \in \mathbb{R}^{d} $.  Then
\*
\expct{\inner{v}{M_tw_{t-1} }^2} &= \expct{\inner{v_1 +  (K_t^\stab)^\top v_2} {w_t}^2}\\ 
&\leq \expct{2 \inner{v_1}{w_t}^2 + 2\inner{(K_t^\stab)^\top v_2}{w_t}^2}\\
&\leq 2(\norm{v_1}^2+ K_u^2\norm{v_2}^2) \expct{\norm{w_t}^2}\\
&\leq 2(1+K_u^2) \expct{\norm{w_t}^2} \\
&\eqqcolon \sigma^2_{\stab}.
\*
By the standard Laurent-Massart bound, we have 
\*
\prob{\sum_t \inner{v}{M_tw_{t-1} } ^2  \geq  \sigma^2_\stab \left(|\cI| + 2 \sqrt{|\cI| \ln(\frac{1}{\delta})}+2  \ln(\frac{1}{\delta})\right)} \leq \delta.
\*
Applying a supermartingale argument, we get
\*
\prob{ \left| \sum_{t} \inner{v}{M_t w_{t-1}}\inner{v}{M\Theta_{t-1}z_{t-1}}\right| \geq  \sigma_\stab \sqrt{2 \ln \frac{1}{\delta}\sum_{t}  \inner{v}{M_t \Theta_{t-1}z_{t-1}}^2 }} &\leq 2\delta,\\
\prob{ \left| \sum_{t} \inner{v}{\xi_t }\inner{v}{M_t \Theta_{t-1}z_{t-1}}\right| \geq  \sigma_{v}\nu_{0} \sqrt{2 \ln \frac{1}{\delta}\sum_{t}  \inner{v}{M_t \Theta_{t-1}z_{t-1}}^2 }} &\leq 2\delta,\\
\prob{ \left| \sum_{t} \inner{v}{M_t w_{t-1} } \inner{v}{\xi_t }  \right| \geq  \sigma_{v}\nu_{0} \sqrt{2 \ln \frac{1}{\delta}\sum_{t}  \inner{v}{M_t w_{t - 1 }}^2 }} &\leq 2\delta. 
\*
By direct computation, we have that
\*
\sum_t \inner{v}{z_t}^2  &\geq \sum_t \inner{v^\parallel_t}{ M_t w_{t-1}}^2 + \sum_t \inner{v^\perp_t}{\xi_t } ^2  
+ 2\inner{v}{M_t w_{t-1} } \inner{v}{M_t \Theta_{t-1}  z_{t- 1}} \\
 &\quad +  2\inner{v}{M_t w_{t-1} } \inner{v}{\xi_t}   + 
2\inner{v}{\xi_t }\inner{v}{M_t \Theta_{t-1}  z_{t- 1} } \\
&\geq  \sigma_v^2 \left(|\cI| - 2 \sqrt{|\cI| \ln(\frac{1}{\delta})}\right)  - 2\sigma_{\stab} \sqrt{2 \ln \frac{1}{\delta}\sum_{t}  \inner{v}{M_t \Theta_{t-1}z_{t-1}}^2 }  \\
& \quad - 2 \sigma_{v} \nu_{0}\sqrt{2 \ln \frac{1}{\delta}\sum_{t}  \inner{v}{M_t \Theta_{t-1}z_{t-1}}^2 } \\
& \quad - 2 \sigma_{v}\nu_{0}   \sqrt{2 \ln \frac{1}{\delta} \left(|\cI| + 2 \sqrt{|\cI| \ln(\frac{1}{\delta})}+2  \ln(\frac{1}{\delta})\right)} \\
&\geq  \sigma_v^2 \left(|\cI| - 2 \sqrt{|\cI| \ln(\frac{1}{\delta})}\right)  - 2\sigma_{\stab} M_u \Theta_u z_u  |\cI|^{\frac{1}{2}} \sqrt{2 \ln \frac{1}{\delta}} \\
&\quad - 2 \sigma_{v} \nu_0 M_u \Theta_u z_u  |\cI|^{\frac{1}{2}}  \sqrt{2 \ln \frac{1}{\delta}}  - 2 \sigma_{v}\nu_{0}  \sqrt{2 \ln \frac{1}{\delta} \left(|\cI| + 2 \sqrt{|\cI| \ln(\frac{1}{\delta})}+2  \ln(\frac{1}{\delta})\right)}\\
& \geq \sigma_v^2 \left(|\cI| - 2 \sqrt{|\cI| \ln(\frac{1}{\delta})}\right)  - 2\sigma_{\stab} M_u \Theta_u z_u  |\cI|^{\frac{1}{2}} \sqrt{2 \ln \frac{1}{\delta}} \\
&\quad - 2 \sigma_{v} \nu_0 M_u \Theta_u z_u  |\cI|^{\frac{1}{2}}  \sqrt{2 \ln \frac{1}{\delta}}  - 2 \sigma_{v}\nu_{0}  \sqrt{2 \ln \frac{1}{\delta} 2|\cI| }\\
& = \sigma_v^2 |\cI| - |\cI|^{\frac{1}{2}}  \sqrt{2 \ln \frac{1}{\delta}} \left(\sqrt{2} \sigma_v^2 + 2\sigma_{\stab} M_u \Theta_u z_u + 2 \sigma_{v} \nu_0 M_u \Theta_u z_u  + 2\sqrt{2} \sigma_{v}\nu_{0}   \right) \\
& \eqqcolon \Lambda_\stab,
\*
holds with probability at least $1-8\delta$, where $M_u := \max_t \norm{M_t}$.  
We can simplify the bound to 
\*
\Lambda_\stab \geq \frac{1}{2}\sigma_v^2 L
\*
given the warm-up block $\cI$ is long enough:
\#
\label{eq:cond_L_5}
|\cI| = L  & \geq 8 \sigma_v^{-4}   \ln \frac{1}{\delta} \left(\sqrt{2} \sigma_v^2 + 2\sigma_{\stab} M_u \Theta_u z_u + 2 \sigma_{v} \nu_0 M_u \Theta_u z_u  + 2\sqrt{2} \sigma_{v}\nu_{0}   \right)^2.
\#
Note that this condition is stricter than $L\geq 16\ln \frac{1}{\delta}$.

By a supermartingale argument, we have 
\#
\label{eq:martingale_stab_1}
\prob{ \left| \sum_{t} \omega_t \inner{v}{z_t} \right| \geq  \psi_i \sqrt{2 \ln \frac{1}{\delta}\sum_{t}  \inner{v}{z_t}^2 }} \leq 2\delta.
\#
Hence, it holds with probability at least $1-9\delta$ that 
\#
|\lambda_v |&\leq \frac{\left| \sum_t \inner{\theta_t - \bar{\theta}}{z_t} \cdot \inner{v}{z_t}  \right|}{\sum_{t}  \inner{v}{z_t}^2 } +  \frac{ \left| \sum_t \omega_t \inner{v}{z_t} \right|}{ \sum_{t}  \inner{v}{z_t}^2 }  \nonumber \\
&\leq   \Delta_\cI z_u|\cI|^{\frac{1}{2}}\Lambda_\stab^{-\frac{1}{2}}\nonumber + \psi_i \sqrt{2 \ln \frac{1}{\delta} }\Lambda_\stab^{-\frac{1}{2}} \\
&  \leq \sqrt{2} \sigma_v^{-1} z_u \Delta_\cI + 2\psi_i  \sigma_v^{-1} \sqrt{ \ln \frac{1}{\delta} }|\cI|^{-\frac{1}{2}}  \nonumber\\
&\eqqcolon \tilde{C}_{1,\stab}\Delta_\cI + \tilde{C}_{2,\stab}|\cI|^{-\frac{1}{2}}.
\label{eq:lambda_final_bound_vector_B0}
\#
Here we have used Cauchy-Schwartz inequality, \eqref{eq:martingale_stab_1}, and the upper bounds  $|\theta_t - \bar\theta|\leq \Delta_\cI$, $ \norm{K_{t}^\stab}\leq K_u $ and the definition of $z_u$. 

Finally, we combine \eqref{eq:lambda_final_bound_vector_B0} with an $\epsilon$-net argument as in the proof of Lemma~\ref{lemma:ols_esti_new}. We let $\varepsilon$ be the confidence parameter. Choosing $\delta = \frac{\varepsilon}{6}$ in Lemma~\ref{lemma:bounded_coundition_number} we get an upper bound on the condition number of the Hessian as $\kappa_0 = \cO(1)$. Setting $\epsilon = 5(1+\kappa_0)$, and applying \eqref{eq:lambda_final_bound_vector_B0} with $\delta= \frac{\varepsilon}{18n (5(1+\kappa_0))^{n+d}}$,  it holds with  probability at least $1-\varepsilon$ that for every row we  have,
\*
\norm{\hat{\theta} - \bar{\theta}}_F & =  \norm{\hat{\theta}_{\mathcal{B}_{i, 0}} - \bar{\theta}} \ \leq\ \frac{5}{3} \left( \tilde{C}_{1,\stab} \Delta_{\mathcal{B}_{i, 0}}  + \tilde{C}_{2,\stab}|\mathcal{B}_{i, 0}|^{- \frac{1}{4}} \right).
\*
Combining the $n$ rows, we have 
\*
\norm{\hat{\Theta}_{\mathcal{B}_{i, 0}} - \bar{\Theta}}_F 
& \leq \frac{5\sqrt{n}}{3} \left( \tilde{C}_{1,\stab} \Delta_{\mathcal{B}_{i, 0}}  + \tilde{C}_{2,\stab}|\mathcal{B}_{i, 0}|^{- \frac{1}{4}} \right) \\
& \leq \breve{C}_{1,\stab} \Delta_{\mathcal{B}_{i, 0}} + \breve{C}_{2,\stab}|\mathcal{B}_{i, 0}|^{- \frac{1}{4}},
\*
where 
\#
\nonumber
\label{eq:def_breve_C_stab}
\breve{C}_{1,\stab} &= 3 \sqrt{n} \sigma_v^{-1} z_u, \\
\breve{C}_{2,\stab} &= 4 \psi_i  \sigma_v^{-1} \sqrt{n}\sqrt{ \ln \frac{1}{\varepsilon} + \ln 18 n + (n+d) \ln 5(1+\kappa_0)}.
\#
For our choice of $x_u, \nu_0, L$, assuming that $T$ is large enough so that $L$ satisfies \eqref{eq:cond_L_5}, and setting $\varepsilon = O(T^{-3})$, both $\breve{C}_{1,\stab},\breve{C}_{2,\stab}$ are $\cO(\sqrt{\ln T})$.

\subsection{Lemmas on the geometry of the Hessian}

The following lemma gives a sufficient condition under which to upper bound the distance of a point $p'=(x',y')$ from the minimizer of a quadratic form $f(x,y)$, it suffices to upper bound the distance of $p'$ from the minimizers of the one-dimensional functions $h(x) = f(x,y')$ and $g(y) = f(x',y)$. In a nut shell, the lemma states that if the level sets of $f$ are ``almost axis-parallel'' (the precise requirement being given by the condition number of the Hessian), then it suffices to obtain upper bounds on the one-dimensional minimizers.

\begin{figure}[t]
    \centering
    \subfigure[A level set of $f(x,y)$]{
    \includegraphics[width=1.8in]{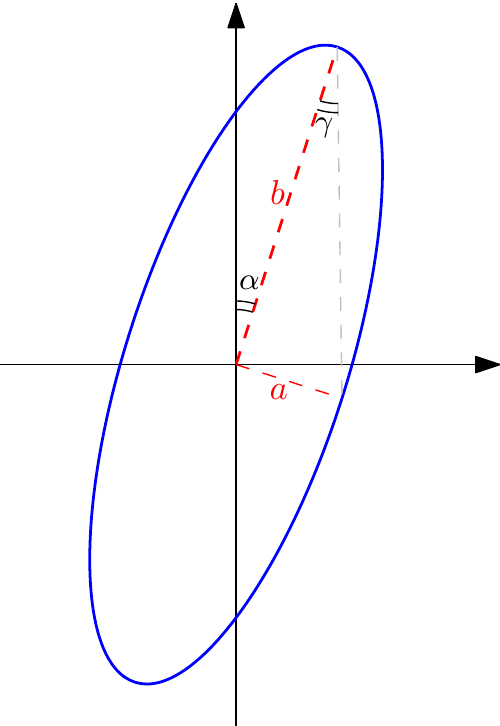}
    }
    \hspace{0.5in}
    \subfigure[Illustration of 1-d minimizers in Lemma~\ref{lem:quadratic_lemma}]{
    \includegraphics[width=1.9in]{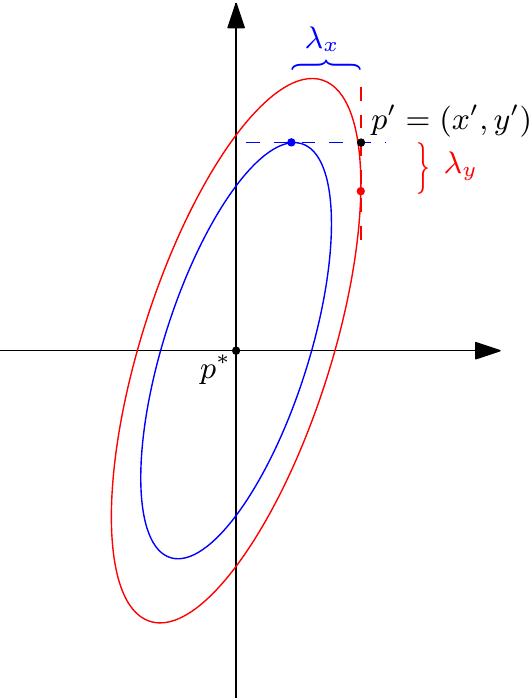}
    \label{fig:quadratic_lemma_case}
    }
    \caption{Illustration of the setup for Lemma~\ref{lem:quadratic_lemma}. The figure on the left shows one level set of the quadratic form $f(x,y)$; a $\alpha$ rotation of an axis-parallel ellipse with principal axes of lengths $a$ and $b$. The figure on the right is a visual illustration of $\lambda_x, \lambda_y$ in the Lemma statement. For example, the blue ellipse denotes the level set on which the minimizer $(x'',y')$ (blue dot) lies, giving $\lambda_x = x' - x''$. } 
    \label{fig:quadratic_lemma}
\end{figure}

\begin{lemma}\label{lem:quadratic_lemma}
Let $f(x,y)$ be a quadratic form with Hessian $H = \begin{bmatrix} A^2 & C \\ C & B^2 \end{bmatrix} \succ 0$. Let the level sets of $f(x,y)$ be given by ellipses that are clockwise rotation by an angle $ \alpha \in \left( -\frac{\pi}{4},  \frac{\pi}{4} \right) $ of axis parallel ellipses: $\frac{x^2}{a^2} + \frac{y^2}{b^2}=r^2$. Define $\gamma$ such that $\tan \gamma = \frac{\min\{a,b\}}{\max\{a,b\} }\leq 1$. 

For a given point $p'=(x',y')$, define 
\begin{align*}
x'' = \argmin_{x} f(x,y'), \qquad y'' = \argmin_{x} f(x',y),
\end{align*}
and $\lambda_x = x' - x''$, $\lambda_y = y' - y''$. Let $p^* = (x^*,y^*) = \argmin_{(x,y)} f(x,y)$ be the true minimizer of $f(\cdot)$. If \textit{(i)} $\tan \gamma \geq \frac{5}{8}$, or \textit{(ii)} $ |\tan \alpha| \leq \frac{\tan^2 \gamma}{4}$, then 
\[  \lambda_x^2 + \lambda_y^2 \geq \frac{1}{2} \norm{p' - p^*}^2 .\]
\end{lemma}
\begin{proof}
We will assume that the true minimizer $p^* = {0}$ without loss of generality, so that we need to prove that $\lambda_x^2 + \lambda_y^2 \geq \frac{1}{2}\cdot(x'^2 + y'^2)$. We will also assume without loss of generality that $a \leq b$. Furthermore, in the proof we will focus on the case $\alpha \in [0, \pi/4]$ and $x',y' \geq 0$, as illustrated in Figure~\ref{fig:quadratic_lemma_case}. This is indeed the hardest case: if $x' \leq 0, y' \geq 0, 0 \leq \alpha \leq \pi/4$, then $ |\lambda_y| \geq |y'|, |\lambda_x | \geq x'$ and the Lemma follows. The other cases are symmetric to one of the above.

We begin by finding an expression for $y''$. Observe that at the point $(x',y'')$, the tangent to the level set is parallel to the $y$-axis. If we now imagine rotating the level set, along with the point $(x',y'')$ counter-clockwise by an angle $\alpha$, so that the level set becomes axis-parallel and the point $(x',y'')$ moves to $(\hat{x},\hat{y})$, then the tangent at $(\hat{x},\hat{y})$ to this axis-parallel ellipse has a slope of $m = -1/\tan \alpha$. We can now obtain one relationship between $\hat{x},\hat{y}$ by differentiating the equation for the level set with respect to $x$:
\[ \frac{d}{dx}\left.\left( \frac{x^2}{a^2} + \frac{y^2}{b^2} \right)\right|_{(\hat{x},\hat{y})} = 0 \ \implies \ \hat{y} = \hat{x} \frac{b^2}{a^2}\tan \alpha.  \]
Let $\tan \beta := \frac{b^2}{a^2}\tan \alpha$, so that $\hat{y} = \hat{x} \tan \beta$. Since $(x',y'')$ is obtained by clockwise rotation of $(\hat{x},\hat{y})$ by $\alpha$, we have
\[ y'' = x' \tan (\beta - \alpha) = x' \frac{\tan \beta - \tan \alpha}{1+\tan \beta \tan \alpha}.\]
Substituting $\tan \beta = \frac{b^2}{a^2} \tan \alpha = \tan \alpha / \tan^2 \gamma$:
\begin{align}
    y'' &= x' \tan\alpha \frac{ 1 - \tan^2 \gamma}{\tan^2 \gamma +\tan^2 \alpha}.
\end{align} 
A similar analysis gives,
\begin{align}
    x'' &= y' \tan\alpha \frac{1 - \tan^2 \gamma}{1 + \tan^2 \gamma \tan^2 \alpha}.
\end{align} 
Writing $(x',y')$ in polar coordinates $(r,\theta)$, we get
\begin{align}
    \lambda_x &= x' - x'' = r \left( \cos \theta - \sin \theta \cdot \tan \alpha \frac{1 - \tan^2 \gamma}{1 + \tan^2 \gamma \tan^2 \alpha} \right) ,\\
    \lambda_y &= y' - y'' = r \left( \sin \theta - \cos \theta \cdot \tan \alpha  \frac{ 1 - \tan^2 \gamma}{\tan^2 \gamma +\tan^2 \alpha} \right).
\end{align}
By our assumptions, $\tan \alpha \geq 0$ and $0 \leq \tan \gamma \leq 1$. One can further verify that under the condition $\tan \alpha \leq \frac{\tan^2 \gamma}{4}$, we have
\[ 0 \leq \tan \alpha \frac{1 - \tan^2 \gamma}{1 + \tan^2 \gamma \tan^2 \alpha}  \leq 1/4 ,\]
and 
\[ 0 \leq \tan \alpha  \frac{ 1 - \tan^2 \gamma}{\tan^2 \gamma +\tan^2 \alpha} \leq 1/4. \]
To see why, the above inequalities can be rearranged into the following quadratic inequalities in $\tan \alpha$ for any fixed value $\tan \gamma$:
\begin{align*}
    \left(\frac{\tan^2 \gamma}{1-\tan^2 \gamma} \right) \tan^2 \alpha  - 4\tan \alpha + \frac{1}{1-\tan^2 \gamma} & \geq 0, \\
    \left(\frac{1}{1-\tan^2 \gamma} \right) \tan^2 \alpha  - 4\tan \alpha + \frac{\tan^2 \gamma}{1-\tan^2 \gamma} & \geq 0.
\end{align*}
It can then be shown that $\frac{\tan^2 \gamma }{4}$ is a lower bound on the smaller roots of the quadratic expressions on the left hand side above. In fact, the second condition above is stricter than the first (when $\tan \gamma \leq 1$), and $\frac{\tan^2 \gamma}{4}$ is a linear approximation to the smaller root (= $2(1-\tan^2 \gamma)-\sqrt{4\tan^2\gamma - 9 \tan \gamma +4}$) in the vicinity of $\tan^2 \gamma = 0$ for the second inequality above. In fact, if $\tan \gamma \geq \frac{5}{8}$, then the two inequalities are always true.

Finally, 
\begin{align*}
    \lambda_x^2 + \lambda_y^2 &= (x'-x'')^2 + (y'-y'')^2 \\
    & \geq r^2 \left[ \sin^2 \theta + \cos^2 \theta - 2 \sin \theta \cos \theta \tan \alpha \left(\frac{1 - \tan^2 \gamma}{1 + \tan^2 \gamma \tan^2 \alpha} +  \frac{ 1 - \tan^2 \gamma}{\tan^2 \gamma +\tan^2 \alpha} \right) \right] \\
    & \geq r^2 \left( 1 - \frac{1}{2} \sin 2\theta \right) \  \geq \  \frac{r^2}{2}.
\end{align*}
\end{proof}

The Hessian for the quadratic described in Lemma~\ref{lem:quadratic_lemma} is given by:
\begin{align}
\label{eqn:hessian}
    H & = \begin{bmatrix}
        A^2 & C \\
        C & B^2            
    \end{bmatrix}
     \ =  \begin{bmatrix}
     \frac{\cos^2 \alpha}{a^2} + \frac{\sin^2 \alpha}{b^2} & \sin \alpha \cos \alpha \left( \frac{1}{b^2} - \frac{1}{a^2} \right) \\  
     \sin \alpha \cos \alpha \left( \frac{1}{b^2} - \frac{1}{a^2} \right)  &      \frac{\sin^2 \alpha}{a^2} + \frac{\cos^2 \alpha}{b^2}
     \end{bmatrix}.
\end{align}

\begin{lemma}
\label{lem:quadratic_lemma_helper}
The Hessian in \eqref{eqn:hessian} satisfies the conditions of Lemma~\ref{lem:quadratic_lemma} if $\frac{|C|}{\min\{ A^2, B^2\}} \leq \frac{1}{33}$.
\end{lemma}
\begin{proof}
Without loss of generality, assume $a \leq b$, so that with $\alpha \in [ -\pi/4, \pi/4 ]$, we have $B^2 \leq A^2$. To neaten the exposition, we will further focus on $\alpha \in [0,\pi/4]$, since only $|\sin \alpha|$ and $|\cos \alpha|$ are involved in verifying the condition.

It suffices to prove that
\begin{align}
\label{eqn:hess_cond}
    \frac{C}{B^2} &= \frac{1}{2} \cdot \frac{ \sin 2\alpha }{a^2/b^2} \cdot \frac{1-\frac{a^2}{b^2}}{1+\left( \frac{b^2}{a^2}  - 1\right)\sin^2 \alpha  } \ \leq \ \frac{1}{33} 
\end{align}
implies $\tan \alpha \leq \frac{a^2}{4 b^2}$, \textit{under the assumption} that $\frac{a^2}{b^2} \leq 25/64$, since otherwise $\tan \gamma \geq 5/8$ and the first condition in      Lemma~\ref{lem:quadratic_lemma} is satisfied. Since under this assumption $1-\frac{a^2}{b^2} \geq \frac{39}{64}$, \eqref{eqn:hess_cond} implies
\[ \frac{ \sin 2\alpha }{a^2/b^2} \cdot \frac{1}{1+\left( \frac{b^2}{a^2}  - 1\right)\sin^2 \alpha  } \ \leq \ \frac{128}{39 \cdot 33} \leq \frac{1}{10}.   \]
Rearranging,
\begin{align} 
\label{eqn:hess_cond_2}
    \frac{b^2}{a^2} \leq \frac{1 - \sin^2 \alpha}{10 \sin 2\alpha - \sin^2 \alpha} \leq \frac{1 - \sin^2 \alpha}{ 10 \sqrt{2} \sin \alpha - \sin^2 \alpha} ,
\end{align}
since in the interval $\alpha \in [0, \pi/4]$ we have $\sin 2\alpha \geq \sqrt{2} \sin \alpha$. Since $\frac{b^2}{a^2} \geq \frac{64}{25} \geq 2$, we get the quadratic inequality
\[  \sin^2 \alpha - 20\sqrt{2} \sin \alpha + 1 \geq 0, \]
which implies $\sin \alpha \leq 0.04$ and therefore $\tan \alpha = \frac{\sin \alpha}{\cos \alpha} \leq 1.01 \sin \alpha$.

Starting from \eqref{eqn:hess_cond_2} again, 
\[  \left( \frac{b^2}{a^2} - 1 \right) \sin^2 \alpha - 10\sqrt{2}\frac{b^2}{a^2} \sin \alpha + 1 \geq 0, \]
which has roots $\frac{10\sqrt{2} b^2/a^2 \pm \sqrt{ 200b^4/a^4 - 4 b^2/a^4 + 4 } }{2(b^2/a^2-1)}$. Since $b^2/a^2 \geq 64/25$, we observe that the larger root is greater than $5\sqrt{2}$, and hence the smaller root is bounded above by  $\frac{1}{5\sqrt{2}} \cdot \frac{1}{\frac{b^2}{a^2}-1} \leq \frac{64/39}{5\sqrt{2}} \cdot \frac{a^2}{b^2}$, which is also an upper bound on $\sin \alpha$. In the last inequality we have again used $b^2/a^2 \geq 64/25$. Finally, $\tan \alpha \leq 1.01 \sin \alpha \leq 1.01 \cdot \frac{64/39}{5\sqrt{2}} \cdot \frac{a^2}{b^2} \leq \frac{a^2 }{4b^2}$ as needed.
\end{proof}

The following lemma adapted from the volume argument of $\epsilon$-net w.r.t.~Euclidean norm \citep{vershynin2010introduction}  gives an upper bound for the covering numbers of the sphere using $\epsilon$-net w.r.t. $\tan$. 

\begin{lemma}
\label{lem:cardinality_eps_net}
Let $N(\epsilon,\mathbb{S}^{n-1}) $ be the minimal cardinality of an $\epsilon$-net of $\mathbb{S}^{n-1}$ such that for every unit vector $v \in \mathbb{S}^{n-1}$, there exists a $v_\epsilon \in \cS_\epsilon$ such that the $\tan$ of the angle between $v$ and $v_{\epsilon}$ is in $[-\epsilon, \epsilon]$. If $\epsilon \leq \frac{1}{5}$, we have that 
\*
N(\epsilon,\mathbb{S}^{n-1}) \leq \left( 1+\frac{4}{\epsilon} \right)^n.  
\*
\end{lemma}

\begin{proof}
Choose $\mathcal{N}_\epsilon$  to be the maximal subset of $\mathbb{S}^{n-1}$ such that the tan of the angle between two arbitrary vectors $v_1,v_2$ is larger than $\epsilon$. By the maximality property, $\mathcal{N}_\epsilon$ is an $\epsilon$-net. Moreover, using the fact that $x \geq \frac{1}{2} 
\tan(x)$ if $x \leq \frac{1}{5}$, the balls of
radii $\frac{\epsilon}{4}$ centered at the points in $\mathcal{N}_\epsilon$  are disjoint. Let $\cB_{n, 2}$ denote the unit Euclidean ball in $\mathbb{R}^n$ centered at the origin. By comparing the volumes, it holds that
\*
N(\epsilon,\mathbb{S}^{n-1}) \cdot \left(\frac{\epsilon}{4}\right)^n vol(\cB_{n})  = N(\epsilon,\mathbb{S}^{n-1}) \cdot vol\left(\frac{\epsilon}{4}\cB_{n}\right) \leq vol\left(\left(1+\frac{\epsilon}{4}\right)\cB_{n}\right) = \left(1+\frac{\epsilon}{4}\right)^n\cdot vol(\cB_{n}). 
\*
Hence we conclude
\*
N(\epsilon,\mathbb{S}^{n-1}) \leq \frac{(1+\frac{\epsilon}{4})^n}{\frac{\epsilon}{4}^n } = \left(1+\frac{4}{\epsilon}\right)^n.
\*
\end{proof}

\begin{lemma}
\label{lem:eps_net_helper}
Let $\mathcal{L}(\theta): \RR^n \to \RR$ be a quadratic form loss function with positive definite Hessian $H$ and minimizer $\theta^*$. Let $\kappa \geq 1$, the condition number, denote the ratio of the largest to the smallest eigenvalue of $H$. Let $\cS_\epsilon \subset \mathbb{S}^{n-1}$ be an `$\epsilon$-net' of the $n$-dimensional unit sphere so that for every unit vector $v \in \mathbb{S}^{n-1}$, there exists a $v_\epsilon \in \cS_\epsilon$ such that the $\tan$ of the angle between $v$ and $v_{\epsilon}$ is in $[-\epsilon, \epsilon]$ and $\epsilon \leq \frac{1}{5+2\kappa}$. Let $\bar{\theta}$ be an approximate minimizer of $\mathcal{L}$, and let $\lambda_v$ denote the minimizer of the scalar quadratic function $\cL_{v}(\lambda) = \cL(\bar{\theta} + \lambda v)$. If for all $v_\epsilon \in \cS_{\epsilon}$, $|\lambda_{v_\epsilon}| \leq \overline{\lambda}$, then $|\bar{\theta} - \theta^*| \leq \frac{\bar{\lambda}}{1 - 2 (1+\kappa) \epsilon}$.
\end{lemma}
\begin{proof}
Let $v=(v_x,v_y)$ denote the unit vector in the direction $\bar{\theta}-\theta^*$, and let $u = (u_x,u_y) \in \cS_\epsilon$ satisfy the condition in the Lemma statement with respect to $v$. That is, if $m_v = \frac{v_y}{v_x}= \tan \alpha $ and $m_u = \frac{u_y}{u_x}=\tan \beta$, then $\tan (\beta -\alpha) = \epsilon'$ with $\epsilon' \in [-\epsilon, \epsilon]$. Let $P$ denote the point $\bar{\theta} + \lambda_{u} u$. The points $\bar{\theta},\theta^*, P$ define a plane and the subsequent analysis will be restricted to this plane. Without loss of generality, let us translate and rotate our co-ordinate system so that $\theta^*$ is at the origin, the level sets of the loss  function $\cL$ restricted to the plane of interest have the form $\frac{x^2}{a^2}+\frac{y^2}{b^2} = r^2$ (with $\frac{1}{\kappa} \leq \frac{a^2}{b^2}\leq \kappa$), and $\bar{\theta} = (\bar{\theta}_x , \bar{\theta}_y) = r_{\bar{\theta}} \cdot v$ lie in the positive quadrant.     

Using the fact that the point $P = \bar{\theta}+\lambda_u u$ is tangent to the level set, we get
\begin{align*}
    -\lambda_u = \frac{\bar{\theta}_x b^2 + \bar{\theta}_y m_u a^2}{a^2 m_u^2 + b^2 } \cdot \sqrt{1+m_u^2} = r_{\bar{\theta}} .\frac{v_x b^2 + v_y m_u a^2}{a^2 m_u^2 + b^2 } \cdot \sqrt{1+m_u^2}.
\end{align*}
Since $m_u = \tan \beta = \frac{\tan \alpha + \epsilon'}{1-\epsilon' \tan \alpha} = \frac{m_v + \epsilon'}{1 - \epsilon' m_v} $, some calculations give,
\begin{align*}
    |\lambda_u| &= r_{\bar{\theta}} .\frac{v_x b^2 + v_y m_u a^2}{a^2 m_u^2 + b^2 } \cdot \sqrt{1+m_u^2}\\
    &= r_{\bar{\theta}} \sqrt{1+\epsilon'^2}\left[  1 - \epsilon' 
    \underbrace{\frac{(m_v+\epsilon')a^2 - m_v (1 -m_v \epsilon')b^2}{(m_v+\epsilon')^2 a^2 + (1-\mu \epsilon')^2 b^2}}_{D} \right].
\end{align*}
We can bound $D$ as:
\begin{align*}
    D &\leq \underbrace{\frac{(m_v+\epsilon')a^2 }{(m_v+\epsilon')^2 a^2 + (1-\mu \epsilon')^2 b^2}}_{D_1} + \underbrace{\frac{ m_v|1-m_v \epsilon'|b^2}{(m_v+\epsilon')^2 a^2 + (1-\mu \epsilon')^2 b^2}}_{D_2}.
\end{align*}
Assuming $|\epsilon'| \leq \epsilon \leq 1/5$, if $m_v \geq 1$, then $D_1 \leq 1$ and $D_2 \leq \frac{2b^2}{a^2} \leq 2 \kappa$. If $m_v \leq 1$, then $D_1 \leq \frac{2a^2}{b^2} \leq 2 \kappa$ and $D_2 \leq 2$. Therefore, $D \leq 2(1+\kappa)$, finally giving:
\begin{align*}
    |\bar{\theta} - \theta^*| = r_{\bar{\theta}} \leq \frac{|\lambda_u|}{\sqrt{1+\epsilon'^2}}\cdot \frac{1}{1-2(1+\kappa)\epsilon'} \leq \frac{\bar{\lambda}}{1 - 2 (1+\kappa) \epsilon}.
\end{align*}
\end{proof}

\subsection{Bound on $\sum_{t\in \cI} \norm{z_t}^2 $}
\begin{lemma}
\label{lem:zu_bound}
For a $\delta \in (0,1)$ and an interval $\cI$ lying within some block $\cB_{ij}$ with $j \geq 1$ and $|\cI| \geq 16 \ln \frac{1}{\delta}$, it holds with probability at least $1-\delta$ that \[ \sum_{t\in \cI} \norm{z_t}^2 \leq |\cI|\left(2\left((1+K_u^2) \max_{t\in \cI}\norm{x_t}^2+ 2\sigma_L^2\right)  \right).\]
In particular, for any $\bar{x} \geq 0$, conditioned on $\max_{t\in \cI} \norm{x_t} \leq \bar{x}$, we have $\sum_{t \in \cI}\norm{z_t}^2 \leq |\cI| \bar{z}$, where $\bar{z}:= \sqrt{2\left((1+K_u^2) \bar{x}^2+ 2\sigma_L^2\right) }$. Here $\sigma^2_L := \nu^2_1 = \sqrt{C_0/2L}$.
\end{lemma}

\begin{proof}
Recall that $z_t = y_t + \xi_t $ and $\norm{K_t}, \norm{K_t^\stab}  \leq K_u$. We have
\*
\sum_{t \in \cI} \norm{z_t}^2 &= \sum_{t \in \cI} \norm{y_t + \xi_t}^2 \\
& \leq 2 \sum_{t \in \cI} \norm{y_t}^2 + 2\sum_{t \in \cI}\norm{\xi_t}^2\\
& \leq  2(1+K_u^2)|\cI| \max_{t\in \cI}\norm{x_t}^2 + 2\sum_{t \in \cI}\norm{\xi_t}^2.
\*
By a standard Laurent-Massart bound, we have:
\*
\prob{\sum_{t \in \cI}\norm{\xi_t} ^2  \geq \sigma_L^2  \left(|\cI| + 2 \sqrt{|\cI| \ln\left(\frac{1}{\delta}\right)}+2  \ln\left(\frac{1}{\delta}\right)\right)} \leq \delta.
\*
Using the fact that $ \left(|\cI| + 2 \sqrt{|\cI| \ln(\frac{1}{\delta})}+2  \ln(\frac{1}{\delta})\right) \leq 2|\cI| $ when $|\cI|\geq 16\ln \frac{1}{\delta}$, plugging the above in the bound derived above for $\sum_t \norm{z_t}^2$ indicates that
\*
\sum_{t \in \cI} \norm{z_t}^2& \leq 2\left((1+K_u^2) \max_{t\in \cI}\norm{x_t}^2+ 2\sigma_L^2\right)  |\cI|
\*
holds with probability at least $1-\delta$. The right hand side of the bound given above is a random variable since it involves $\max_{t \in \cI}\norm{x_t}$, and can instead be interpreted as saying that for any $\bar{x}$, conditioned on the event $\max_{t\in \cI}\norm{x_t} \leq \bar{x}$, $\sum_{t \in \cI} \norm{z_t}^2 \leq 2\left((1+K_u^2) \bar{x}^2+ 2\sigma_L^2\right)  |\cI| =: |\cI|\bar{z}$.
\end{proof}

\begin{definition}
Define $z_u := \sqrt{2\left((1+K_u^2) x_u^2+ 2\sigma_L^2\right) } $, where $x_u$ is defined in Algorithm~\ref{alg:DYN-LQR}.
\end{definition}

\subsection{Bound on the condition number of $\sum_{t\in \cI} z_t z_t^\top$}

\begin{lemma}\label{lemma:bounded_coundition_number}
For an arbitrary interval $\cI$, denote design matrix $\Upsilon_\cI = \sum_{t\in \cI} z_t z_t^\top $ and its condition number $\kappa = \lambda_{max}(\Upsilon_\cI)/\lambda_{min}(\Upsilon_{\cI})$.

\textit{(i)} Let $\cI$ be an interval within a block $\cB_{i,j}$ in Algorithm \ref{alg:DYN-LQR}. Define $\bar{x} = \max_{t \in \cI}\norm{x_t}$, and $\bar{z}$ as in Lemma~\ref{lem:zu_bound}. If we have 
\*
|\cI| &\geq \frac{2000}{9}\left(2 
(n+d) \log\frac{1}{\delta} +(n+d) \log \frac{\bar{x}^2(1+\norm{K}^2)+\sigma_L^2}{{ \sigma}^2_\cI \min \left\{\frac{1}{2}, \frac{\psi^2}{{ \sigma}^2_L+2\norm{K}^2}\right\}}\right),
\*
then for $\delta \leq 3/100$, it holds with  probability at least $1- 3\delta$ that the condition number is upper bounded as 
\*
\kappa \leq \frac{\bar{z}^2 |\cI| }{ \frac{9|\cI|}{1600} { \sigma}^2_\cI \min \left\{\frac{1}{2}, \frac{\psi^2}{{ \sigma}^2_L+2\norm{K}^2}\right\}}  = \frac{1600\bar{z}^2}{9 { \sigma}^2_\cI \min \left\{\frac{1}{2}, \frac{\psi^2}{{ \sigma}^2_L+2\norm{K}^2}\right\}}.
\*
Define $\kappa_\cI$ be the bound above when $\bar{x}=x_u$:
\[  \kappa_{\cI} \coloneqq \frac{1600z_u^2}{9 { \sigma}^2_\cI \min \left\{\frac{1}{2}, \frac{\psi^2}{{ \sigma}^2_L+2\norm{K}^2}\right\}}, \]
whence it follows that $\kappa_{\cI} \leq C_9 \sqrt{|\cI|}$ for some problem-dependent constant (independent of $T$) $C_9$.

\textit{(ii)} For any warm-up block $\cB_{i, 0}$ in in Algorithm \ref{alg:DYN-LQR}, with sequentially strongly stablizing policies $\{K_t^\stab\} $ such that $\norm{K_t^\stab} \leq K_u $, if 
\*
|\cB_{i, 0} | &\geq \frac{2000}{9}\left(2 
(n+d) \log \frac{1}{\delta} +(n+d) \log \frac{\bar{x}^2(1+K_u^2)+\sigma_0^2}{{ \sigma}^2_0 \min \left\{\frac{1}{2}, \frac{\psi^2}{{ \sigma}^2_0+2(K_l^\stab)^2}\right\}}\right),
\*
we have for $\delta \leq 3/100$, 
\*
\kappa \leq \frac{\bar{z}^2 |\cB_{i, 0}| }{ \frac{9|\cB_{i, 0}|}{1600} { \sigma}^2_0 \min \left\{\frac{1}{2}, \frac{\psi^2}{{ \nu}^2_0+2\norm{K}^2}\right\}}  = \frac{1600\bar{z}^2}{9 { \nu}^2_0 \min \left\{\frac{1}{2}, \frac{\psi^2}{{ \nu}^2_0+2\norm{K}^2}\right\}}.
\*
Define $\kappa_0$ to be the bound above when $\bar{x}=x_u$:
\[ \kappa_0 \coloneqq \frac{1600 z_u^2}{9 { \nu}^2_0 \min \left\{\frac{1}{2}, \frac{\psi^2}{{ \nu}^2_0+2\norm{K}^2}\right\}}, \]
from where it follows that $\kappa_{0} \leq C_{10} \ln T$ for some problem dependent constant $C_{10}$. 

\end{lemma}
\begin{proof}
We need to bound $\lambda_{\min}(\Upsilon_\cI)$  and $\lambda_{\max}( \Upsilon_\cI)$ separately. By direct computation and Lemma~\ref{lem:zu_bound}, it holds with probability at least $1-\delta$ that
\*
\lambda_{\max}( \Upsilon_\cI) \leq \Tr(\Upsilon_\cI)  = \sum_{t\in \cI} z_t z_t^\top = \sum_{t\in \cI} \norm{z_t}^2 \leq \bar{z}^2 |\cI| .
\*

In the sequel, we bound $\lambda_{\min}(\Upsilon_\cI)$ from below by specifying the choice of $\Upsilon_{0}$ such that $\Upsilon_{0}\preceq \Upsilon_\cI$  with high probability using Lemma~\ref{lemma:cov_lower_bound}. Note that $z_t \mid \cF_{t-1} \sim \mathcal{N}\left(\overline{z}_{t}, \Sigma_{t}\right)$, where $\overline{z}_{t}$ and $ \Sigma_{t}$ are measurable and 
\*
\Sigma_{t}\succeq \left[\begin{array}{cc}
\psi^2 I_n & \psi^2 I_n K^{\top} \\
 \psi^2 K I_n & \psi^2 K  I_d K^{\top}+{ \sigma}^2_t I_d
\end{array}\right].
\*
By  \citet[Lemma F. 6]{dean2018regret}, we have 
\*
\lambda_{\min}\left(\Sigma_{t}\right) \geq { \sigma}^2_t \min \left\{\frac{1}{2}, \frac{\psi^2}{{ \sigma}^2_t+2\norm{K}^2}\right\}\geq  { \sigma}^2_\cI \min \left\{\frac{1}{2}, \frac{\psi^2}{{ \sigma}^2_L+2\norm{K}^2}\right\} .
\*
Moreover, we have 
\*
\Tr\left(\expct{\Upsilon_\cI } \right)&= \expct{\sum_{t\in \cI} \left\|{x}_{t}\right\|^{2}+\left\|{u}_{t}\right\|^{2} }\\
&\leq \expct{\sum_{t\in \cI} \left\|{x}_{t}\right\|^{2}+\|K_t\|^2\left\|{x}_{t}\right\|^{2} + \sigma_L^2}\\
&\leq |\cI| \left(\bar{x}^2(1+\norm{K}^2)+\sigma_L^2\right).
\*
Setting $\cE = \Omega$ (the probability space) and 
\*
\Upsilon_0 = \frac{9|\cI|}{1600} { \sigma}^2_\cI \min \left\{\frac{1}{2}, \frac{\psi^2}{{ \sigma}^2_L+2\norm{K}^2}\right\}  I_{d+n},
\* 
Lemma~\ref{lemma:cov_lower_bound} implies  if 
\#\label{eq:suff_long_I}
|\cI| &\geq  \frac{2000}{9}\left(2 
(n+d) \log \frac{100}{3}+(n+d) \log \frac{\bar{x}^2(1+\norm{K}^2)+\sigma_L^2}{{ \sigma}^2_\cI \min \left\{\frac{1}{2}, \frac{\psi^2}{{ \sigma}^2_L+2\norm{K}^2}\right\}}\right) \\
& \geq \frac{2000}{9}\left(2 
(n+d) \log \frac{1}{\delta}  +(n+d) \log \frac{\bar{x}^2(1+\norm{K}^2)+\sigma_L^2}{{ \sigma}^2_\cI \min \left\{\frac{1}{2}, \frac{\psi^2}{{ \sigma}^2_L+2\norm{K}^2}\right\}}\right)
\#
with $\delta \leq 3/100$, we have
\*
\mathbb{P}\left[ \Upsilon_T \nsucceq \Upsilon_0 \right] &\leq 2\exp(-\frac{9}{2000((n+d)+1)} |\cI|) \\
& \leq 2\exp\left(-\frac{9}{2000((n+d)+1)} \frac{2000}{9}\left( 2 
(n+d) \log \frac{1}{\delta} \right)\right)\\
&\leq   2\exp\left(-\frac{9}{2000((n+d)+1)} \frac{2000}{9}\left( 
(n+d+1) \log \frac{1}{\delta}\right)\right)\\
& = 2\delta.
\*
Then it holds with probability at least $1 -2\delta-\delta = 1-3 \delta$  that 
\#\label{eq:conditional_number_upper_bound}
\kappa \leq \frac{\bar{z}^2 |\cI| }{ \frac{9|\cI|}{1600} { \sigma}^2_\cI \min \left\{\frac{1}{2}, \frac{\psi^2}{{ \sigma}^2_L+2\norm{K}^2}\right\}}  = \frac{1600\bar{z}^2}{9 { \sigma}^2_\cI \min \left\{\frac{1}{2}, \frac{\psi^2}{{ \sigma}^2_L+2\norm{K}^2}\right\}}.
\#

For a warm-up block $\cB_{i, 0}$, note that the exploration noise is fixed at $\nu_0^2=1$, and we have $\left\|K_{t}^{\text {stab }}\right\| \leq K_{u}$. Plugging  these parameters into \eqref{eq:suff_long_I} and \eqref{eq:conditional_number_upper_bound} yields the corresponding results. 
\end{proof}

\subsection{Supporting Lemmas}

\begin{lemma}[Lemma E.4 in \cite{simchowitz2020naive}]\label{lemma:cov_lower_bound}
Suppose $z_t \mid \cF_{t-1} \sim \mathcal{N}\left(\overline{z}_{t}, \Sigma_{t}\right)$, where $\overline{z}_{t}\in \RR^{\tilde{d}}$ and $\Sigma_{t} \in \RR^{\tilde{d}\times \tilde{d}}$ are $\cF_{t-1}$-measurable, and $\Sigma_{t} \succeq \Sigma \succ 0$. Suppose $\cE$ is an arbitrary event and suppose $\Tr\left(\expct{V_\cI \mathbbm{1}\{\cE\}}\right)\leq \Lambda T$
 for some  constant $\Lambda\geq 0$. Then for 
\*
T\geq \frac{2000}{9}(2\tilde{d}\log(\frac{100}{3})+\tilde{d}\log \frac{\Lambda}{\lambda_{\min}(\Sigma)}),
\*
let $V_0 \coloneqq \frac{9T}{1600} \Sigma $, it holds that
\*
\prob{\{V_T \nsucceq V_0 \} \cap \cE } \leq 2\exp\left( -\frac{9}{2000(\tilde{d}+1)} T \right).
\*
\end{lemma}

\begin{lemma}[Self-Normalized Tail Bound \citep{abbasi2011improved}]
\label{lem:self_normalized_tail_bound} Let $\{\eta_t\}_{t\geq 1}$ be a $\cF_t$-adapted sequence such that $\eta_t \mid \cF_{t-1}$ is $\sigma^2$-sub-Gaussian. Define $V_T\coloneqq \sum_{t=1}^{T} z_{t} z_{t}^{\top}$. Fix $V_0\succ 0 $, it holds with probability $1-\delta$ that
\*
\left\|\sum_{t=1}^{T} \mathbf{x}_{t} \eta_{t}\right\|_{\left(V_{0}+V_{T}\right)^{-1}}^{2} \leq 2 \sigma^{2} \log \left\{\frac{1}{\delta} \operatorname{det}\left(V_{0}^{-1 / 2}\left(V_{0}+V_{T}\right) V_{0}^{-1 / 2}\right)\right\}.
\*
\end{lemma}

\section{Proof of Proposition~\ref{prop:dynopt_regret}}
\label{sec:proof_dynopt_regret}

Consider a non-stationary Markov decision process on state space $\cS$ and action space $\cA$, with a time-invariant cost function $c(\cdot, \cdot):\cS \times \cA \to \RR$, and time-dependent transition kernel parametrized by $\{\Theta_t\}_{t\in [T]}$. Let $J^*_t$ denote the optimal (minimum) average cost of the MDP corresponding to $\Theta_t$, and let $h_t(\cdot) : \cS \to \RR$ denote the relative value (bias) function. Then for an arbitrary state $s_t \in \cS$ and action $a_t\in \cA$, we have the inequality:
\begin{align}
 c(s_t, a_t) \geq J^*_t + h_t(s_t) - \expctsub{\Theta_t}{h(s_{t+1}) \mid s_t, a_t},
\end{align}
where $\expctsub{\Theta_t}{X}$ denotes the expectation of random variable $X$ under transition kernel parametrized by $\Theta_t$. Summing the above inequality from $t=1$ to $T$:
\begin{align*}
    \sum_{t=1}^T c(s_t, a_t) &\geq \sum_{t=1}^T J^*_t + h_t(s_t) - \expctsub{\Theta_t}{h(s_{t+1}) \mid s_t, a_t} \\
    & = \sum_{t=1}^T J^*_t + h_1(s_1) - \expctsub{\Theta_T}{h_T(s_{T+1}) \mid s_T, a_T} \\
    & \qquad + \sum_{t=1}^{T-1} h_{t+1}(s_{t+1}) - \expctsub{\Theta_t}{h_t(s_{t+1}) \mid s_t, a_t} \\
    & = \sum_{t=1}^T J^*_t + h_1(s_1) - \expctsub{\Theta_T}{h_T(s_{T+1}) \mid s_T, a_T} \\
    & \qquad + \sum_{t=1}^{T-1} h_{t+1}(s_{t+1}) - \expctsub{\Theta_t}{h_{t+1}(s_{t+1}) \mid s_t, a_t} \\
    & \qquad + \sum_{t=1}^{T-1} \expctsub{\Theta_t}{h_{t+1}(s_{t+1}) - h_{t}(s_{t+1})\mid s_t, a_t}. 
\end{align*}
Taking expectation with respect to the randomization of the policy and the evolution of the non-stationary MDP,
\begin{align*}
    \expct{ \sum_{t=1}^T c(s_t, a_t)} - \sum_{t=1}^T J^*_t & \geq h_1(s_1) - \expct{ h_T(s_{T+1})} + \sum_{t=1}^{T-1} \expct{h_{t+1}(s_{t+1}) - h_{t}(s_{t+1}) }
    \intertext{or}
         \sum_{t=1}^T J^*_t - \expct{ \sum_{t=1}^T c(s_t, a_t)}  & \leq - h_1(s_1) + \expct{ h_T(s_{T+1})} + \sum_{t=1}^{T-1} \expct{h_{t}(s_{t+1}) - h_{t+1}(s_{t+1}) }.
\end{align*}
Specializing to the non-stationary LQR setting, $s_t \equiv x_t$, $a_t \equiv u_t$, $c(x_t, u_t) = x_t^\top Q x_t + u_t^\top R u_t$, $h_t(s_t) \equiv x_t^\top P^*_t x_t$:
\begin{align*}
         \sum_{t=1}^T J^*_t - \expct{ \sum_{t=1}^T c(x_t, u_t)} 
         & \leq  \expct{ x_{T+1}^\top P_T^* x_{T+1}} + \sum_{t=1}^{T-1} \expct{x_{t+1}^\top \left( P_{t}^* - P_{t+1}^*\right) x_{t+1} } \\
         & \leq  \expct{ \norm{x_{T+1}}^2} \norm{P_T^*} + \sum_{t=1}^{T-1} \expct{ \norm{x_{t+1}}^2} \norm{ P_{t}^* - P_{t+1}^*}.
\end{align*}
In Lemma~\ref{lem:bounded_xnorm}, we prove that under the optimal dynamic policy, $\expct{\norm{x_t}^2}$ is bounded from above by a constant depending only on the cost parameters and the sequential stability parameters $\kappa, \gamma$. The perturbation result for the solution of Discrete Algebraic Riccati equation gives $\norm{P_t^* - P_{t+1}^*} \leq \min\{ 2C_4 \norm{\Theta_t - \Theta_{t+1}}^2, 2 P_u\} = \cO(\Delta^2_{t+1})$ \cite[Theorem 5]{simchowitz2020naive}. Lemma~\ref{lem:bounded_xnorm} does not bound $\expct{\norm{x_{T+1}}^2}$, however, following a similar argument as in the Lemma, we can create another policy that has logarithmic regret compared to the optimal policy and has bounded $\expct{\norm{x_{T+1}}^2}$. Combining these, we get the desired bound on the additional regret of $\cO(V_T + \log T)$ with respect to the dynamic optimal policy.
\hfill $\Box$

\begin{lemma}\label{lem:bounded_xnorm}
Under the optimal dynamic policy for the non-stationary LQR problem, 
\[ q_{\min} \expct{\norm{x_t}^2} \leq M_\Gamma \frac{2\kappa^2}{\gamma} M_x + M_P \frac{2\kappa^2}{\gamma} + \left( \kappa^2 M_x + \frac{2\kappa^2 n \psi^2}{\gamma} \right) \left(  \frac{2\kappa^2 }{\gamma} M_\Gamma + \kappa^2 M_P\right),\]
where $M_x := \left(\frac{M_\Gamma}{q_{\min}}\right) \frac{2\kappa^2 n\psi^2}{\gamma}$, $M_P := \frac{2n \kappa^2}{\gamma}M_\Gamma $, and $M_\Gamma := \max_{s} \norm{Q + (K_s^\stab)^\top R K_s^\stab } \leq q_{\max}+ r_{\max}\kappa^2$.
\end{lemma}
\begin{proof}
We prove this result by contradiction. We first establish some notation for the optimal dynamic policy. A classical fact is that the optimal dynamic policy for non-stationary LQR is also a linear state feedback policy, given via the following dynamic programming recursion:
\begin{align*}
    P_{T+1} &= 0, \\
    K_t &= -(R+B_t^\top P_{t+1} B_t) B_t^\top P_{t+1} A_t, \\
    P_t &= Q + K_t^\top R K_t + (A_t+B_tK_t)^\top P_{t+1} (A_t + B_t K_t), \\
    J_t &= \Tr(W\cdot P_{t+1}).
\end{align*}

Let $t$ be some time such that under the optimal dynamic policy, $\expct{\norm{x_{t}}^2}$ is larger than the bound in the Lemma statement. Define
\[ \tau = \max\{ s \leq t-1 : \expct{\norm{x_s}^2} \leq M_x \} \]
as the last time before $t$ when the expected squared norm of the state under the optimal policy is smaller than $M_x$. Similarly, define
\[ \tau' = \min\{ s \geq t : \norm{P_{s+1}} \leq M_P \} \]
as the first time including or after $t$ when the norm of $P_{s+1}$ is smaller than $M_P$. We will show that by deviating to a policy where $K'_s = K_s^\stab$ for $  s \in \{ \tau, \ldots, \tau' \}$ gives a policy with a smaller cost. Let $\{x'_s\}$ denote the state process for this new policy, $\{P'_s\}$ the relative value function matrices, and $J'_s := \Tr(W \cdot P'_s)$.

By the definition of the new policy, we must have $P'_s = P_s$ for $s \geq \tau'+1$. Recall the recursion for the relative value function for LQR:
\[ x_t^\top Q x_t + u_t^\top R u_t = x_t^\top P_t x_t + J_t - \expct{x_{t+1}^T P_{t+1} x_{t+1}}. \]
We will decompose the cost of the optimal policy into contributions due to the four intervals $\{1,\ldots, \tau-1\}$, $\{\tau, \ldots, t\}$, $\{t+1, \ldots, \tau'\}$ and $\{\tau'+1, \ldots, T\}$. Since both policies agree on the first interval, the total cost is the same, and hence we do not consider it henceforth. For the interval $\{ \tau, \ldots, t\}$ we lower bound the cost of the optimal policy as:
\begin{align}
    \expct{\sum_{s=\tau}^t x_s^\top Q x_s + u_s^\top R u_s} & \geq \sum_{s=\tau}^t q_{\min} \expct{\norm{x_s}^2}.
    \label{eqn:LB_term2}
\end{align}
For the interval $\{t+1 \ldots, \tau'\}$:
\begin{align*}
    \expct{\sum_{s=t+1}^{\tau'} x_s^\top Q x_s + u_s^\top R u_s} & = \expct{x_{t+1}^\top P_{t+1} x_{t+1}} + \sum_{s=t+1}^{\tau'}J_s - \expct{x_{\tau'+1}^\top P_{\tau'+1}x_{\tau'+1}} \\
    & \geq \sum_{s=t+1}^{\tau'+1} \Tr(W\cdot P_{s}) - \expct{x_{\tau'+1}^\top P_{\tau'+1}x_{\tau'+1}},
\end{align*}
where we have used $\expct{x_{t+1}^\top P_{t+1} x_{t+1}} \geq \expct{w_{t+1}^\top P_{t+1}w_{t+1}} = \Tr(W \cdot P_{t+1})$.
Finally, for the last interval, 
\begin{align*}
    \expct{\sum_{s=\tau'+1}^{T} x_s^\top Q x_s + u_s^\top R u_s} & = \expct{x_{\tau'+1}^\top P_{\tau'+1}x_{\tau'+1}} + \sum_{s=\tau'+1}^T J_s.
\end{align*}
It will be convenient to combine the lower bound for the interval $\{t+1,\ldots, T\}$ as:
 \begin{align}
 \nonumber
    \expct{\sum_{s=t+1}^{T} x_s^\top Q x_s + u_s^\top R u_s} & \geq \sum_{s=t+1}^{\tau'} \Tr(W\cdot P_{s}) + \sum_{s=\tau'}^T J_s \\
    & \geq \sum_{t=t+1}^{\tau'} \psi^2 \norm{P_s} + \sum_{s=\tau'}^T J_s. 
    \label{eqn:LB_term3}
\end{align}
We now proceed to upper bound the cost during these intervals for the modified policy $\{K'_s\}$. Denote $u'_s = K'_s x'_s$ as the control at time step $s$ under the new policy. We first summarize the results of Lemma~\ref{lem:norm_x_P_stab}, which bounds $\expct{\norm{x'_s}^2}$ and $\norm{P'_s}$ for $s \in \{ \tau+1, \ldots, \tau'\}$:
\begin{align*}
    \expct{\norm{x_s}^2} & \leq \kappa^2 \left( 1 - \frac{\gamma}{2}\right)^{2(s-\tau)} \expct{\norm{x_\tau}^2} + \frac{2\kappa^2 n \psi^2}{\gamma}, \\
    \norm{P'_s} & \leq \kappa^2 \left( 1 -\frac{\gamma}{2} \right)^{2(\tau'-s+1)} \norm{P_{\tau'+1}} + \frac{2\kappa^2}{\gamma} M_\Gamma.
\end{align*}
For the second interval, $\{\tau, \ldots, t\}$, we can upper bound the cost of the new policy by:
\begin{align}
\nonumber
    \expct{\sum_{s=\tau}^t {x'}_s^\top Q x'_s + {u'}_s^\top R u'_s} & = \expct{\sum_{s=\tau}^t {x'}_s^\top (Q + {K'}_s^\top R K'_s) x'_s } \\
    \nonumber
    & \leq M_\Gamma \sum_{s=\tau}^t \expct{\norm{x'_s}^2} \\
    & \leq M_\Gamma \left(  (t-\tau) \frac{2\kappa^2 n \psi^2}{\gamma} + \frac{2\kappa^2}{\gamma} \expct{\norm{x_\tau}^2}  \right),
        \label{eqn:UB_term2}
\end{align}
where we have used the bound on $\expct{\norm{x'_s}^2}$ from Lemma~\ref{lem:norm_x_P_stab}. 
Next, we upper bound the cost for interval $\{t+1,\ldots, T\}$:
\begin{align*}
    \expct{\sum_{s=t+1}^{\tau'} {x'}_s^\top Q x'_s + {u'}_s^\top R u'_s} & = \expct{{x'_{t+1}}^\top P'_{t+1} x'_{t+1}} + \sum_{s={t+1}}^T J'_s \\
    &= \expct{{x'_{t+1}}^\top P'_{t+1} x'_{t+1}}  + \sum_{s=t+2}^{\tau'} \Tr(W \cdot P'_s) +   \sum_{s={\tau'}}^T J'_s \\
    & \leq \expct{\norm{x'_{t+1}}^2} \norm{P'_{t+1}} + n \psi^2 \sum_{s={t+2}}^{\tau'} \norm{P'_s} + \sum_{s=\tau'}^T J_s.
\end{align*}
Using Lemma~\ref{lem:norm_x_P_stab} we can bound:
\[ \sum_{s={t+2}}^{\tau'} \norm{P'_s} \leq \sum_{s={t+2}}^{\tau'}\left(  \kappa^2 \left( 1 -\frac{\gamma}{2} \right)^{2(\tau'-s+1)} \norm{P_{\tau'+1}} + \frac{2\kappa^2}{\gamma} M_\Gamma \right) \leq  (\tau'-t-1) \frac{2\kappa^2}{\gamma}M_\Gamma + \frac{2\kappa^2}{\gamma} \norm{P_{\tau'+1}}. \]
Also, using Lemma~\ref{lem:norm_x_P_stab}:
\[ \expct{\norm{x'_{t+1}}^2} \norm{P'_{t+1}} \leq \left( \kappa^2 \expct{\norm{x_\tau}^2} + \frac{2\kappa^2 n \psi^2}{\gamma} \right) \left(  \kappa^2 \norm{P_{\tau'+1}}  + \frac{2\kappa^2}{\gamma} M_{\Gamma }\right).  \]
Putting together,
\begin{align}
\nonumber
    \expct{\sum_{s=t+1}^{\tau'} {x'}_s^\top Q x'_s + {u'}_s^\top R u'_s} & \leq \left( \kappa^2 \expct{\norm{x_\tau}^2} + \frac{2\kappa^2 n \psi^2}{\gamma} \right) \left(  \kappa^2 \norm{P_{\tau'+1}}  + \frac{2\kappa^2}{\gamma} M_{\Gamma }\right) \\
    & \qquad + n\psi^2 \left(  (\tau'-t) \frac{2\kappa^2}{\gamma}M_\Gamma + \frac{2\kappa^2}{\gamma} \norm{P_{\tau'+1}} \right) + \sum_{s=\tau'+1}^T J_s.
    \label{eqn:UB_term3}
    \end{align}
Combining the lower bounds on the optimal policy cost from \eqref{eqn:LB_term2}-\eqref{eqn:LB_term3} and upper bound on the cost of the modified policy from \eqref{eqn:UB_term2}-\eqref{eqn:UB_term3}:
    \begin{align*}
        & \expct{\sum_{s=\tau}^{T} {x}_s^\top Q x_s + {u}_s^\top R u_s} -         \expct{\sum_{s=\tau}^{T} {x'}_s^\top Q x'_s + {u'}_s^\top R u'_s} \\
        \geq & \   \sum_{s=\tau}^{t-1} \left(q_{\min}\expct{\norm{x_t}^2}  - M_\Gamma \frac{2\kappa^2 n \psi^2}{\gamma}  \right)   \\
        & \qquad + \psi^2 \sum_{s=t+1}^{\tau'} \left(  \norm{P_s} - \frac{2\kappa^2n}{\gamma}M_\Gamma \right) \\
        & \qquad + q_{\min} \expct{\norm{x_t}^2} - \frac{2\kappa^2}{\gamma} \norm{P_{\tau'+1}} - M_\Gamma \frac{2\kappa^2}{\gamma} \expct{\norm{x_{\tau}}^2}  \\
        & \qquad - \left( \kappa^2 \expct{\norm{x_\tau}^2} + \frac{2\kappa^2 n \psi^2}{\gamma} \right) \left(  \kappa^2 \norm{P_{\tau'+1}}  + \frac{2\kappa^2}{\gamma} M_{\Gamma }\right).
    \end{align*}
Recall that we choose $M_x := \left(\frac{M_\Gamma}{q_{\min}}\right) \frac{2\kappa^2 n \psi^2}{\gamma}$ as the threshold for $\expct{\norm{x_\tau}^2}$, and $M_P := \frac{2n \kappa^2}{\gamma}M_\Gamma $ as the threshold of $\norm{P_{\tau'+1}}$. This ensures the first two terms above are non-negative. Therefore, if 
\begin{align*} q_{\min}  \expct{\norm{x_t}^2} & \geq \frac{2\kappa^2}{\gamma} \norm{P_{\tau'+1}} + M_\Gamma \frac{2\kappa^2}{\gamma} \expct{\norm{x_{\tau}}^2}  +  \left( \kappa^2 \expct{\norm{x_\tau}^2} + \frac{2\kappa^2 n \psi^2}{\gamma} \right) \left(  \kappa^2 \norm{P_{\tau'+1}}  + \frac{2\kappa^2}{\gamma} M_{\Gamma }\right)  \\
& \geq \frac{2\kappa^2}{\gamma} M_P + M_\Gamma \frac{2\kappa^2}{\gamma} M_x  +  \left( \kappa^2 M_x + \frac{2\kappa^2 n \psi^2}{\gamma} \right) \left(  \kappa^2 M_P  + \frac{2\kappa^2}{\gamma} M_{\Gamma }\right),
\end{align*}
we get that the cost of the optimal policy is larger than the modified policy, a contradiction.
\end{proof}

\begin{lemma}\label{lem:norm_x_P_stab} For the alternate policy which chooses $K'_s = K^\stab_s$ for $s \in \{\tau, \ldots, t, \ldots, \tau'\}$, we have:
\begin{itemize}
    \item $\expct{\norm{x_s}^2} \leq \kappa^2 \left( 1 - \frac{\gamma}{2}\right)^{2(s-\tau)} \expct{\norm{x_\tau}^2} + \frac{2\kappa^2 n \psi^2}{\gamma}$,
    \item $\norm{P'_s}  \leq \kappa^2 \left( 1 -\frac{\gamma}{2} \right)^{2(\tau'-s+1)} \norm{P_{\tau'+1}} + \frac{2\kappa^2}{\gamma} M_\Gamma$,
\end{itemize}
where $M_\Gamma := \max_{s} \norm{Q + (K_{s}^\stab)^\top R K_s^\stab}$, and $\{x_s\}, \{P_s\}$ denote the optimal policy quantities.
\end{lemma}
\begin{proof}
Recall our notation $\Phi_t := A_t + B_t K_t^\stab$, $\Gamma_t = Q + (K_t^\stab)^\top R K_t^\stab$. 
Denote for $a \leq b$: 
\[ \Phi_{b:a} = \Phi_{b} \Phi_{b-1} \cdots \Phi_{a}.\]
We can write:
\begin{align*}
    x_s &= \Phi_{s-1} x_{s-1} + w_{s-1} \\
    &= \Phi_{s-1} \Phi_{s-2} x_{s-2} + \Phi_{s-1} w_{s-2} + w_{s-1} \\
    &=  \Phi_{s-1:\tau}  x_\tau + \sum_{\ell=\tau}^{s-1}  \Phi_{s-1:\ell+1}   w_{\ell},
\end{align*}
which gives:
\begin{align}
    \expct{\norm{x_s}^2} & \leq \norm{  \Phi_{s-1:\tau}}^2 \expct{\norm{x_\tau}^2} + \sum_{\ell=\tau}^{s-1} \norm{ \Phi_{s-1:\ell+1} }^2  \expct{\norm{w_{\ell}}^2} .
    \label{eqn:xnorm_bound}
\end{align}
By sequential strong stability:
\begin{align*}
    \norm{ \Phi_{s-1:\ell} } & = \norm{ H_{s-1}L_{s-1}H_{s-1}^{-1}H_{s-2}L_{s-2}H_{s-2}^{-1} \cdots H_{\ell}L_{\ell}H_{\ell}^-1 } \\
    & \leq \norm{ H_{s-1}} \norm{ L_{s-1}} \norm{ H_{s-1}^{-1}H_{s-2}} 
    \norm{L_{s-2}} \cdots \norm{ H_{\ell+1}^{-1} H_{\ell}} \norm{L_{\ell}} \norm{H_{\ell}^-1}\\
    & \leq \kappa (1-\gamma)^{s-\ell}(1+\gamma/2)^{s-\ell-1}, \\
    \intertext{which using $(1-\gamma)(1+\gamma/2)\leq (1-\gamma/2)$ gives}
    & \leq \kappa(1-\gamma/2)^{s-\ell}.
\end{align*}
Substituting the above in \eqref{eqn:xnorm_bound} and using $\expct{\norm{w_s}^2} = n \psi^2$:
\begin{align*}
    \expct{\norm{x_s}^2} & \leq \kappa^2 (1-\gamma/2)^{2(s-\tau)} \expct{\norm{x_\tau}^2} + n\psi^2 \kappa^2 \left( 1 + (1-\gamma/2)^2 + \cdots + (1-\gamma/2)^{2(s-\tau-1)}  \right)\\
    & \leq \kappa^2 (1-\gamma/2)^{2(s-\tau)} \expct{\norm{x_\tau}^2} + n\psi^2 \kappa^2 \frac{2}{\gamma}.
\end{align*}
This proves the first part. For the second part,
\begin{align*}
    P'_s & = Q + (K_{s}^\stab)^\top R K_{s}^\stab + \Phi_s^\top P'_{s+1} \Phi_s \\
    &= \Gamma_s + \Phi_s^\top P'_{s+1}\Phi_s \\
    &= \Gamma_s + \Phi_s^\top \Gamma_{s+1} \Phi_s + \Phi_s^\top \Phi_{s+1}^\top P'_{s+2} \Phi_{s+1} \Phi_{s} \\
    &=  \Phi_{\tau':s}^\top P_{\tau'+1}  \Phi_{\tau':s}  + \sum_{\ell=s}^{\tau'}  \Phi_{\ell-1:s}^\top \Gamma_\ell  \Phi_{\ell-1:s} .
\end{align*}
Therefore,
\begin{align*}
    \norm{P'_s} & \leq \norm{\prod_{m=s}^{\tau'} \Phi_{\tau':s} }^2  \norm{P_{\tau'+1}} + \sum_{\ell=s}^{\tau'} \norm{ \prod_{m=s}^{\ell-1} \Phi_{\ell-1:s} }^2 \norm{\Gamma_\ell} \\
    & \leq \kappa^2 (1-\gamma/2)^{2(\tau'+1-s)} \norm{P_{\tau'+1}} + M_\Gamma \kappa^2 \left( 1 + (1-\gamma/2)^2 + \cdots + (1-\gamma/2)^{2(\tau'-s)} \right) \\
    & \leq \kappa^2 (1-\gamma/2)^{2(\tau'+1-s)} \norm{P_{\tau'+1}} + M_\Gamma \frac{2\kappa^2}{\gamma}.
\end{align*}

\end{proof}

\section{Proof of Theorem~\ref{thm:dynalg_regret}}
\label{sec:proof_dynalg}
\subsection{Proofs for Section~\ref{sec:reg_decomp}}

\paragraph{Proof of Lemma~\ref{lem:stab_epoch}} 

Consider a stabilization epoch starting at time $\tau^\stab$ and ending at $\theta^\stab$. During the epoch, the dynamics of the state evolution is:
\[ x_{s} = \Phi_{s-1} x_{s-1} + w_{s-1} ,\]
where $\Phi_{s} := A_{s} + B_{s} K_{s}^\stab$. Denote for $a \leq b$: 
\[ \Phi_{b:a} = \Phi_{b} \Phi_{b-1} \cdots \Phi_{a}.\]
Then, 
\begin{align*}
\norm{x_s} & \leq \norm{\Phi_{s-1:\tau^\stab} } \norm{x_{\tau^\stab}} + \norm{\Phi_{s-1:\tau^\stab}} \norm{w_{\tau^\stab}} + \norm{\Phi_{s-1:\tau^\stab+1}} \norm{w_{\tau^\stab+1}} + \cdots + \norm{\Phi_{s-1}} \norm{w_{s-1}} \\
& \leq \kappa (1-\gamma/2)^{s-\tau^\stab} \norm{x_{\tau^\stab}} + \kappa (1-\gamma/2)^{s-\tau^\stab} \norm{w_{\tau^\stab}} + \kappa (1-\gamma/2)^{s-\tau^\stab-1} \norm{w_{\tau^\stab+1}} \\
& \qquad + \cdots + \kappa (1-\gamma/2) \norm{w_{s-1}} \\
& =: \kappa Y_{s-\tau^\stab},
\end{align*}
so that $\norm{x_{t+\tau^\stab}} \leq \kappa Y_t$ with $Y_0 = \norm{x_{\tau^\stab}}$, $W_t = \norm{w_{t+\tau^\stab}}$
and
\[ Y_{t+1} \leq (1-\gamma/2) Y_t + W_t .\]
Now applying Lemma~\ref{lem:hitting_time} with $\rho = \rho_0 =  (1-\gamma/2)$, and substituting $a = \frac{2\psi \sqrt{n}}{1-\rho_0}$, we get
\[ \rho_0 + (1-\rho_0) \frac{\expct{\norm{w_t}}}{2 \psi \sqrt{n}}  \leq \rho_0 + (1-\rho_0) \frac{\psi \sqrt{n}}{2 \psi \sqrt{n}} = \frac{1+\rho_0}{2}. \]
Therefore,
\[ \sum_{t = \tau^\stab}^{\theta^\stab} \norm{x_t}^2 \leq 2\kappa^2 \frac{\norm{x_{\tau^\stab}}^2}{1-\rho_0}. \]
Since 
\[ \expct{c_t \mid \cF_{t-1}, \cG_{t-1}} = x_t^\top Q x_t + [(A_t + B_t K_t)x_t]^\top R [(A_t + B_t K_t)x_t] + \sigma_t^2 \Tr(R) = {\cO}(\norm{x_t}^2 + 1),\]
the total cost during the stabilization epoch is bounded by $\cO(\norm{x_{\tau^\stab}}^2)$. We next show that this is $\cO(\ln T)$. Note that we have $\norm{x_{\tau^\stab-1}} \leq x_u = 2\kappa e^{C_{ss}} \left( \frac{\sqrt{8(n+d)}\beta}{\sqrt{1-\rho_0}} \sqrt{\log T} + \frac{(n+d)B}{1-\rho_0} \right)$. Therefore, 
\*
\expct{\norm{x_{\tau^\stab}}^2} &\leq \expct{ \sup_{\substack{{y_1, \ldots y_{T-1} \in \RR^n:} \\{\norm{y_t} \leq x_u \forall t \in [T-1]}}} \max_{t}  \norm{ (A_t + B_t K_t)y_t + B_t \sigma_t \eta_t + w_t  }^2 }.
\* 
Since $\max_{t} \max\{\norm{A_t}, \norm{B_t},\norm{K_t} \}$ are bounded, for some problem dependent constant $C_{13}$
\*
\expct{\norm{x_{\tau^\stab}}^2} &\leq C_{13}\left( \expct{ \max_{t} \norm{w_t}^2 + \max_{t} \norm{\eta_t}^2} + x_u \expct{ \max_t \norm{w_t} + \max_{t} \norm{\eta_t} } \right. \\
& \qquad \qquad \qquad  \qquad \qquad \left.+ \expct{\max_t \norm{w_t}} \cdot \expct{\max_t \norm{\eta_t}}   \right).
\* 
Using Lemma~\ref{lem:bound_max_chi_squared}, the expression above is $\cO(n + d + \ln T)$.

\paragraph{Proof of Lemma~\ref{lem:warmup}} 

Since $\cE_i$ is an exploration epoch, we have $\norm{x_{\tau_i}} \leq x_u$. During $\cB_{i,0}$ the dynamics are given by:
\[ x_{t+1} = \Phi_t x_t + \nu_0 \Xi_t  \eta_t + w_t, \]
where $\Phi_t := A_t + B_t K_t^\stab$ and $\Xi_t := B_t K^\stab_t$. Denote for $a\leq b$: 
\[ \Phi_{b:a}=\Phi_b \Phi_{b-1}\cdots \Phi_a.\]
By our assumption that $\eta_t$ and $w_t$ are a sequence of independent mean 0 Gaussian random vectors:
\begin{align*} \expct{\norm{x_{t}}^2 \mid \cF_{\tau_{i}-1},\cG_{\tau_{i}-1}} & = \norm{\Phi_{t:\tau_i} x_{\tau_i}}^2 + \sum_{s=\tau_i}^{t-1} \nu_0^2 \expct{\norm{ \Phi_{t:s+1} \Xi_s \eta_s}^2 \mid \cF_{\tau_i-1},\cG_{\tau_i-1}}  \\
& \qquad + \sum_{s=\tau_i}^{t-1} \expct{ \norm{\Phi_{t:s+1} w_s}^2 \mid \cF_{\tau_i-1},\cG_{\tau_i-1} } \\ 
& \leq  \kappa^2 \rho_0^{2(t-\tau_i)} \norm{x_{\tau_i}}^2 + \kappa^2 n \frac{\nu_0^2 C_6 + \psi^2 }{1-\rho_0^2},
\intertext{where $C_6 := \max_{t \in [T]} \norm{B_t K^\stab_t}^2$. This further gives the total during $\cB_{i,0}$} as
\expct{\sum_{t=\tau_i}^{\tau_i+ L-1}\norm{x_{t}}^2 \mid \cF_{\tau_i-1},\cG_{\tau_i-1}} & \leq  \kappa^2 \frac{x_{u}^2}{1-\rho_0^2} + L \kappa^2 n \cdot \frac{\nu_0^2 C_6 + \psi^2}{1-\rho_0^2} .
\end{align*}
 Finally, again using the fact that $\expct{c_t \mid \cF_{\tau_i-1}} = \tilde{\cO}( \expct{\norm{x_t}^2 \mid \cF_{\tau_i-1}}  + 1)$, and the definition of $L := \frac{16 (n+d)\log^3 T}{1-\rho_0}$, the bound in the lemma statement follows.

\paragraph{Proof of Lemma \ref{lem:expected_regret_goodint}}
We first prove a lemma that gives a regret decomposition for good intervals.

\begin{lemma}
\label{lem:regret_decomposition} For some epoch $\cE_i$, a block $\cB_{i,j}$ in epoch $\cE_i$, and a good interval $\cI^\good_{i,j,k} = [\tau, \theta]$ in block $\cB_{i,j}$, the following identity holds:
\begin{align*}
    \Reg^\pi(\cI^\good_{i,j,k}) & :=\sum_{t\in \cI^\good_{i,j,k}} x_t^\top Q x_t + u_t^\top R u_t - J^*_t \\ 
    &=  \sum_{t =\tau}^\theta J_t(K_t) - J^*_t  \\ 
\nonumber & \quad + x_\tau^\top P_\tau(K_\tau) x_\tau - x_{\theta+1}^\top P_\theta(K_\theta)x_{\theta+1} \\
\nonumber & \quad + \sum_{t=\tau}^{\theta-1} x_{t+1}^\top \left( P_{t+1}(K_{t+1}) - P_{t}(K_t) \right) x_{t+1} \\
\nonumber & \quad + \sum_{t=\tau}^\theta \left( x_{t+1}^\top P_t(K_t)  x_{t+1}  - \expct{x_{t+1}^\top P_t(K_t)  x_{t+1} \mid x_{t},\sigma_t} \right) \\
\label{eqn:regret_goodint} & \quad + \sum_{t=\tau}^\theta \sigma_t^2 \Tr\left( R +  B_t^\top P_t(K_t) B_t \right) .
\end{align*}
\end{lemma}
\begin{proof}
Note that for an interval lying within block $\cB_{i,j}$, the policy $K_t = K^*(\hat{\Theta}_{i,j-1})$ is fixed, however for generality, we use $K_t$.   
For dynamics given by $\Theta_t$, and control policy $u_t =  K_t x_t + \sigma_t \eta_t$ with $\eta_t \sim \cN(0,I_n)$, we have the following Bellman recursion:
\[ x_t^\top P_t(K_t) x_t = x_t^\top Q x_t + u_t^\top R u_t - J_t(K_t) - \sigma_t^2 \Tr\left(R + B_t^\top P_t(K_t) B_t \right) + \expct{ x_{t+1}^\top P_t(K_t)x_{t+1} \mid x_t, \sigma_t}.  \]
Rearranging terms, we get,
\begin{align*}
x_t^\top Q x_t &+ u_t^\top R u_t - J^*_t \\
&=  J_t(K_t) - J^*_t + x_t^\top P_t(K_t) x_t  - \expct{ x_{t+1}^\top P_t(K_t)x_{t+1} \mid x_t, \sigma_t} + \sigma_t^2 \Tr\left(R + B_t^\top P_t(K_t) B_t \right) \\
&= J_t(K_t) - J^*_t + \left( x_t^\top P_t(K_t) x_t - x_{t+1}^\top P_{t+1}(K_{t+1}) x_{t+1}\right) \\
& \qquad  + \left(x_{t+1}^\top P_{t+1}(K_{t+1})x_{t+1} - x_{t+1}^\top P_{t}(K_{t})x_{t+1} \right) \\
& \qquad + \left( x_{t+1}^\top P_{t}(K_{t})x_{t+1} - \expct{ x_{t+1}^\top P_t(K_t)x_{t+1} \mid x_t, \sigma_t}\right) + \sigma_t^2 \Tr\left(R + B_t^\top P_t(K_t) B_t \right)
\end{align*}
Summing the above for the entire interval gives the identity in the lemma.
\end{proof}

With Lemma~\ref{lem:regret_decomposition}, after taking expectation and using Lemma~\ref{lem:quadratic}, we prove Lemma~\ref{lem:expected_regret_goodint} below.

\begin{proof}
The second expression follows from the first by noting the definition of good intervals: for all $t \in \cI^\good_{i,j,k}$, $\norm{\hat{\Theta}_{i,j-1} - \Theta_t} \leq C_3$ and applying Lemma~\ref{lem:quadratic}.

To arrive at the first expression, we go through the expression in Lemma~\ref{lem:regret_decomposition} line-by-line. The expression in the second line is bounded by $\norm{x_\tau}^2 \norm{P_\tau(K_\tau)}$, which is $\tilde{\cO}\left( \frac{n+d+\log T}{1-\rho_0} \right)$. For the expression in the third line, noting that $K_t = K_{t+1}$, and that $\norm{P_t(K) - P_{t+1}(K)} \leq  C_{12} \norm{\Theta_t - \Theta_{t+1}}$ for a stabilizing controller $K$, and a constant $ C_{12}$, the sum is bounded by $ C_{12} \sum_{t=\tau}^\theta \norm{x_t}^2 \norm{\Theta_t - \Theta_{t+1}}$, which is $\tilde{\cO}\left( \frac{n+d+\log T}{1-\rho_0} \Delta_{i,j,k} \right)$. The expression in the fourth line is a mean 0 random variable and hence vanishes when we take the expectation. For the expression in the last line, note that in block $\cB_{i,j}$, for each $m=0,1,\ldots, j-1$ we start an exploration phase of scale $m$ (duration $L\cdot2^m$) with probability $\frac{1}{L\sqrt{2^j}\sqrt{2^m}}$ at each time $t$, and during an exploration of phase $m$, we choose $\sigma^2_t = \sqrt{\frac{C_0}{L2^m}}$. We will  upper bound $\expct{\sum_{t \in \cI^\good_{i,j,k}} \sigma_t^2}$ by allocating the entire exploration variance to the time $t$ at which an exploration phase begins. For a given time $t$, this gives the expected contribution due to scale $m$ as $\frac{1}{L\sqrt{2^j}\sqrt{2^m}} \times \sqrt{C_0 2^m/L} = \frac{C_0^{1/2}}{L^{3/2}\sqrt{2^j}}$. Summing over $m$ gives $\frac{jC_0^{1/2}}{L^{3/2}\sqrt{2^j}}$, and multiplying by $\size{\cI^\good_{i,j,k}}$ finally gives the expression in the Lemma.  
\end{proof}

\subsection{Proofs for Section~\ref{sec:bound_epochs}}

\paragraph{Proof of Lemma \ref{lemma:epoch_bound1}}

Consider the block $\cB_{i,j} = [s_{i,j},s_{i,j} + 2^jL -1 ]$ in $\cE_i$. We first show that no restart is triggered by \textsc{EndOfBlockTest}($i,j$). Let $t = \tau_i + 2^j L -1$, then $\Delta_{\cB_{i,j}}\leq \Delta_{[\tau_i,t]}\leq(t - \tau_i + 1 )^{-1/4}\leq |\cB_{i,j}|^{-1/4}$. Conditioning on  Event $\cE$, we have
\*
\norm{\Theta_{s_{i,j}} - \hat{\Theta}_{i,j}}_F\leq \bar{C}_\bias \Delta_{\cB_{i,j}}  + \bar{C}_\var |\cB_{i,j}|^{-1/4}.
\* 
Similarly, we also have 
\*
\norm{\Theta_{s_{i,j-1}} - \hat{\Theta}_{i,j-1}}_F\leq \bar{C}_\bias \Delta_{\cB_{i,j-1}}  + \bar{C}_\var |\cB_{i,j-1}|^{-1/4} 
\*
and
\*
\norm{\Theta_{s_{i,j-1}} - \hat{\Theta}_{[\tau_i, t]}}_F&\leq \Delta_{[\tau_i, t]}  + C_1|t - \tau_i +1|^{-1/4}, \\
\norm{\Theta_{s_{i,j}} - \hat{\Theta}_{[\tau_i, t]}}_F&\leq \Delta_{[\tau_i, t]}  + C_1|t - \tau_i +1 |^{-1/4} .
\*
Then 
\*
\norm{\hat{\Theta}_{i,j} - \hat{\Theta}_{i,j-1}}_F
&\leq 
\norm{\Theta_{s_{i,j}} - \hat{\Theta}_{i,j}}_F
 + \norm{\Theta_{s_{i,j}} - {\Theta}_{s_{i,j-1}}}_F
+\norm{\Theta_{s_{i,j-1}} - \hat{\Theta}_{i,j-1}}_F \\
&\leq
\bar{C}_\bias \Delta_{\cB_{i,j}}  + \bar{C}_\var |\cB_{i,j}|^{-1/4}
+ \Delta_{\tau_i,t}
+ \bar{C}_\bias \Delta_{\cB_{i,j-1}}  + \bar{C}_\var |\cB_{i,j-1}|^{-1/4} \\
& \leq (1+\bar{C}_\bias)  \Delta_{\tau_i,t} + 2 \bar{C}_\var |\cB_{i,j-1}|^{-1/4}  \\
& \leq (1 + \bar{C}_\bias + 2 \bar{C}_{\var}) |\cB_{i,j-1}|^{-1/4}. \* 
As a result, $\norm{\hat{\Theta}_{i,j} - \hat{\Theta}_{i,j-1}}_F^2\leq (1 + \bar{C}_\bias + 2 \bar{C}_{\var})^2|\cB_{i,j-1}|^{-1/2}$ and  $ \textsc{EndOfBlockTest}(i,j) = Pass$. Similarly, for any exploration interval $\cI = [s,e] \subset [\tau_i, t]$ with index $m\leq j-1$, note that $\Delta_{\cI }\leq \Delta_{[\tau_i,t]}\leq(t - \tau_i + 1 )^{-1/4}\leq |\cI|^{-1/4}$. Then 
\*
\norm{\hat{\Theta}_{i,j,(m,s)} - \hat{\Theta}_{i,j-1}}_F
&\leq \norm{\hat{\Theta}_{i,j,(m,s)} -{\Theta}_{s}}_F + \norm{\Theta_s - \Theta_{s_{i,j-1}}}_F + \norm{{\Theta}_{s_{i,j-1}} - \hat{\Theta}_{i,j-1}}_F\\
&\leq
\bar{C}_\bias \Delta_{\cI}  + \bar{C}_\var |\cI|^{-1/4}
+ \Delta_{\tau_i,t}
+ \bar{C}_\bias \Delta_{\cB_{i,j-1}}  + \bar{C}_\var |\cB_{i,j-1}|^{-1/4} \\
& \leq (1+\bar{C}_\bias)  \Delta_{\tau_i,t} + 2 \bar{C}_\var |\cI|^{-1/4}  \\
& \leq (1 + \bar{C}_\bias + 2 \bar{C}_{\var}) |\cI|^{-1/4}.
\*
Then $\norm{\hat{\Theta}_{i,j,(m,s)} - \hat{\Theta}_{i,j-1}}_F^2\leq (1 + \bar{C}_\bias + 2 \bar{C}_{\var})^2|\cI|^{-1/2}$ and $\textsc{EndOfExplorationTest}(i,j,m,s)= Pass$.

\paragraph{Proof of Corollary \ref{coro:epoch_bound1}}

By Lemma~\ref{lemma:epoch_bound1}, to end an epoch $\cE_i$ due to  detection of nonstationarity, we need $\Delta_{[\tau_i,t]}\geq \sqrt{\frac{C_0}{ |\cE_i|^{1/2}}}$. Then 
\*
\Delta_{T} \geq \sum_{i=1}^{E} \Delta_{[\tau_i,t]}   \geq  \sum_{i=1}^{E} \sqrt{\frac{C_0}{ |\cE_i|^{1/2}}}
\*
or
\[\sum_{i=1}^E |\cE_i|^{-\frac{1}{4}} \leq C_0^{-\frac{1}{2}}  \Delta_T . \]
By H\"{o}lder's inequality, 
\*
E & \leq \left(\sum_{i=1}^{E} |\cE_i|^{-\frac{1}{4}}\right)^{\frac{4}{5}} \left(\sum_{i=1}^{E} |\cE_i|\right)^{\frac{1}{5}} \\
& \leq \left( C_0^{-\frac{1}{2}}  \Delta_T \right)^{\frac{4}{5}} \left( T\right)^{\frac{1}{5}}  \\
&=   C_0^{-\frac{2}{5}}  \Delta_T^{\frac{4}{5}} T^{\frac{1}{5}}.
\*

\paragraph{Proof of Lemma \ref{lem:instab}}

By Assumption~\ref{assum:approx_stabilization}, a control based on an estimate $\hat{\Theta}_t$ of $\Theta_t$ such that $\norm{\Theta_t - \hat{\Theta}_t}_F \leq C_3/2$ guarantees that $K^*(\hat{\Theta}_t)$ is in fact $(\kappa,\gamma,\nu)$ sequentially strongly-stable for the epoch $\cE_i$ and parameters $\kappa,\gamma,\nu$ specified in Lemma~\ref{lem:nonstat_seq_stg_stab_bdd_norm}. We first show that under the assumption that $\{K_t\}$ is a $(\kappa, \gamma,\nu)$ sequentially strongly-stable sequence of controllers for the non-stationary dynamics in the interval $[s,e]$ ($1 \leq s \leq e \leq T$), then  Lemma~\ref{lem:hitting_upper} implies that with high probability $\max_{t \in [s,e]}\norm{x_t} \leq x_u$. 

The LQR dynamics are given by:
\[ x_{t+1} = (A_t + B_t K_t)x_t + \sigma_t B_t \eta_t + w_t. \]
Under the independence assumptions on $\{\eta_t\}, \{w_t\}$, we can use the analysis approach in Lemma~\ref{lem:stab_epoch} to show that $\norm{x_t} \leq \kappa e^{C_{ss} V_{[s:t-1]}} Y_t$ where $Y_t$ obeys $Y_{s-1} = \norm{x_{s-1}}$ and
\begin{align*} Y_{t+1} & \leq (1-\gamma) Y_t + \sum_{i=1}^d \beta_{i,t} |\hat{\eta}_{i,t}|  + \psi \sum_{j=1}^n |\hat{w}_{i,t}|.
\end{align*}
In the above, $\beta_{i,t}$ are the singular values of $B_t$, and $\hat{\eta}_{i,t}, \hat{w}_{i,t}$ are independent $\mathcal{N}(0,1)$ random variables. We have used the fact that $\sigma_t \leq 1$ for all $t$. Denoting:
\[ \beta = \max\left\{ \psi,  \max_{i,t} \beta_{i,t} \right\} \]
and applying Lemma~\ref{lem:hitting_upper},
\begin{align}
    \prob{\max_{t \in [s,e] } Y_t \geq Y_{s-1} + \left( \frac{\sqrt{8(n+d)}\beta}{\sqrt{1-\rho_0}} \sqrt{\log T} + \frac{(n+d)B}{1-\rho_0} \right)} & \leq \frac{|e-s+1|}{T^4}, \intertext{or,}
    \label{eqn:epoch_excursion}
    \prob{\max_{t \in [s,e] } \norm{x_t}e^{-C_{ss} V_{[s:e-1]}} /\kappa \geq \norm{x_{s-1}} + \left( \frac{\sqrt{8(n+d)}\beta}{\sqrt{1-\rho_0}} \sqrt{\log T} + \frac{(n+d)B}{1-\rho_0} \right)} & \leq \frac{|e-s+1|}{T^4}.
\end{align}
Note that we start an epoch with $\norm{x_{\tau_i}} \leq x_u$, and then use stabilizing controls for $L$ time steps. Using \eqref{lem:hitting_upper},
\begin{align}
\label{eqn:init_norm}
\prob{ \norm{x_{\tau_i+L}} \geq  \kappa^2 \left( \rho_0^{L} x_u + \left( \frac{\sqrt{6(n+d)}\beta}{\sqrt{1-\rho_0}} \sqrt{\log T} + \frac{(n+d)B}{1-\rho_0} \right) \right)}  \leq \frac{1}{T^3}. 
\end{align}
With our choice of $L$, $\rho_0^L x_u = o(1)$. 

Lemma~\ref{lemma:ols_esti_warmup} and Lemma~\ref{lemma:ols_esti_new} prove that under \textsc{Event 1} the OLS estimate $\hat{\Theta}_{i,j}$ based on block  $j \geq 0$ of epoch $\cI_i$ indeed satisfies $\norm{\Theta_t - \hat{\Theta}_t}_F^2 \leq C_3$. Thus it holds that the controllers $\{K_t\}$ are indeed $(\kappa, \gamma, \nu)$ sequentially strongly-stable for $t \in [\tau_i+L,  \theta_i]$. Combining \eqref{eqn:init_norm} and \eqref{eqn:epoch_excursion},  for epoch $\cE_i = [\tau_i,\theta_i]$ we get 
\begin{align}
        \prob{\max_{t \in [\tau_i+L,\theta_i] } \norm{x_t} \geq 2\kappa e^{C_{ss}} \left( \frac{\sqrt{8(n+d)}\beta}{\sqrt{1-\rho_0}} \sqrt{\log T} + \frac{(n+d)B}{1-\rho_0} \right)} & \leq \frac{2}{T^3}.
\end{align}
Therefore, with high probability, a restart of the epoch based on instability detection does not happen.

\subsection{Proofs for Section~\ref{sec:upper}}
\label{sec:proof_loss_function}

\begin{lemma}\label{lemma:phase_end}
Assume $\textsc{Event 1}$ holds. Let $\cI = [s,e]$ be an interval in $\cB_{i,j}$ satisfying $\Delta_\cI^2\leq\alpha_\cI = \frac{1}{\sqrt{|\cI|}}$ and $\varepsilon_\cI  = \norm{ \hat{\Theta}_{i,j-1} -  \Theta_{s} }_F^2  \geq C_5 \cdot \frac{1}{\sqrt{\cI}}$, where we define $C_5 \coloneqq (2 + 2\bar{C}_\bias + 3 \bar{C}_\var)^2$. Define $ \tilde{\varepsilon}_\cI \coloneqq \min\{ \varepsilon_\cI, C_3\}$. Then, there exists an index $m$, such that \emph{(1)} $C_5 \cdot \frac{1 }{\sqrt{2^{m+1} L}}\leq \tilde{\varepsilon}_\cI \leq C_5 \cdot  \frac{1}{\sqrt{2^{m} L}}$,  \emph{(2)} $2^m L \leq |\cI| $, and   \emph{(3)} if an exploration phase with index $m$ starts at some time $\tilde{s}$ within the interval $[s,e - 2^m L]$, then the algorithm starts a  new epoch at the end of the exploration phase.
\end{lemma}
\begin{proof}

By our assumption,  $\tilde{\varepsilon}_\cI  \leq \frac{C_5 }{\sqrt{L}}$. Note that $\cI\subset\cJ$, then $\tilde{\varepsilon}_\cI\geq  C_5 \cdot \frac{1}{\sqrt{\cI}}\geq  C_5 \cdot \frac{1}{\sqrt{2^j L}}$. Then there exist a index $m \in [j]$ satisfying  \emph{(1)}. \emph{(2)} is implied by $C_5 \cdot \frac{1 }{\sqrt{|\cI|}}\leq \tilde{\varepsilon}_\cI  \leq C_5 \cdot  \frac{1 }{\sqrt{2^{m} L}}$. 

To prove \emph{(3)}, let $\tilde{s}\in [s,e - 2^m L]$ be the starting time of $\cI$, condition on Event $\cE$ and note that $[\tilde{s},\tilde{s} + 2^mL ]\subset I $. We have 
\*
\norm{\hat{\Theta}_{i,j,(m,\tilde{s})} - \Theta_{\tilde{s}}}_F \leq \bar{C}_{\bias} \Delta_\cI +  \bar{C}_\var |\cI|^{-\frac{1}{2}}.
\*
By direct computation, 
\*
\norm{\hat{\Theta}_{i,j-1} - \hat{\Theta}_{i,j,(m,s)}}_F 
&\geq 
\norm{\hat{\Theta}_{i,j-1}- \Theta_{\tilde{s}}}_F  -  \norm{ \Theta_{\tilde{s}} - \Theta_s }_F -  \norm{ \Theta_s -\hat{\Theta}_{i,j,(m,\tilde{s})} }_F\\
&\geq  \sqrt{ C_5} \cdot {\cI}^{-\frac{1}{4}}- \Delta_\cI - \bar{C}_\bias \Delta_\cI -  \bar{C}_\var |\cI|^{-\frac{1}{2}}\\
&\geq  \sqrt{ C_5} \cdot {\cI}^{-\frac{1}{4}}  - (1+\bar{C}_{\bias}){\cI}^{-\frac{1}{4}} - C_{\var} |\cI|^{-\frac{1}{2}}\\
&\geq( \sqrt{ C_5} - 1 - \bar{C}_\bias - \bar{C}_\var) {\cI}^{-\frac{1}{4}} .
\*
Hence 
\*
\norm{\hat{\Theta}_{i,j-1} - \hat{\Theta}_{i,j,(m,\tilde{s})}}_F^2 \geq ( \sqrt{ C_5} - 1 - \bar{C}_\bias - \bar{C}_\var)^2 {\cI}^{-\frac{1}{2}}\geq (1+\bar{C}_\bias + 2 \bar{C}_\var)^2 {\cI}^{-\frac{1}{2}}
\*
and 
$\textsc{EndOfExplorationTest}(i,j,m,s) = \textit{Fail}$.
\end{proof}

\paragraph{Proof of Lemma~\ref{lem:interval_loss}}
Starting with the definition in \eqref{eqn:defn_L},
\*
\cL(I) &:= \sum_{t \in \cI} \min\left\{ C_4 \norm{\hat{\Theta}_{i,j-1} - \Theta_t}_F^2 , C_3 \right\} \\
& \leq  C_4 \sum_{t \in \cI} \norm{\Theta_t - \hat{\Theta}_{i,j-1}}_F^2\\
& \leq  2 C_4 |\cI| \norm{ \hat{\Theta}_{i,j-1} - \Theta_s }_F^2 
+  2 C_4 \sum_{t \in \cI} \norm{\Theta_t - \Theta_s}_F^2 \\
& \leq 2C_4|\cI|\left( \left( \alpha_{\cI} + \varepsilon_{\cI} \mathbbm{1}\{ \varepsilon_{\cI} \geq \alpha_{\cI}\}\right) +  \Delta^2_{\cI} \right) .
\*

\paragraph{Proof of Lemma~\ref{lemma:partition}}

We create the partition using Algorithm \ref{alg:create_partition}, where we  check the truncating condition \emph{current interval ends and a new interval is created at time $t\in\cJ$ whenever} $ \Delta_{\left[s_{k}, t\right]} \leq \sqrt{\frac{\log|J|}{(t-s_{k})^{1/2}+1}} \text { and } \Delta_{\left[s_{k}, t+1\right]}>\sqrt{\frac{\log|J|}{(t-s_{k})^{1/2}+2}}$ at each time $t\in \cJ$. 
\begin{algorithm}[th!]
 \caption{\textsc{Creating Partition}} 
 \label{alg:create_partition}
 \SetKw{KwInit}{Initialize:}
 \SetKw{KwInput}{Input:}
\SetAlgoLined
\KwInput{an block $\cJ = [s,e]$.}\;
\KwInit{ Set $k = 1$; $s_1 = s$; $t = s$.}\;
 \While{$t \leq e$}{
  \If{$\Delta_{\left[s_{k}, t\right]} \leq \sqrt{\frac{\log|J|}{(t-s_{k})^{1/2}+1}} \text { and } \Delta_{\left[s_{k}, t+1\right]}>\sqrt{\frac{\log|J|}{(t-s_{k})^{1/2}+2}}$}
  {Let $e_k \leftarrow t$; $\cI_k \leftarrow [s_k, e_k]$; $k \leftarrow k+1$.  } 
  $t  \leftarrow t+1$
  }
 \If{$s_k\leq e$}{
 $e_k \leftarrow e$; $\cI_k \leftarrow [s_k, e_k]$. 
 }
\end{algorithm}

To calculate an upper bound for the number of intervals $\Gamma$, consider the inequality
\*
\Delta_{[s, e]} \geq \Delta_{\left[s_{1}, e_{1}+1\right]}+\Delta_{\left[s_{2}, e_{2}+1\right]}+\ldots+\Delta_{\left[s_{\Gamma-1}, e_{\Gamma-1}+1\right]} \geq \sum_{k=1}^{\Gamma-1} \sqrt{\frac{\log|J|}{(e_{k}-s_{k})^{1/2}+2}}=\sum_{k=1}^{\Gamma-1} \sqrt{\frac{\log|J|}{\mathcal{I}_{k}^{1/2}+1}}.
\*
On the other hand, by Holder's inequality, 
\*
\left(\sum_{k=1}^{\Gamma-1} \sqrt{\frac{\log|J|}{\left|\mathcal{I}_{k}\right|^{1/2}+1}}\right)^{\frac{2}{3}}\left(\sum_{k=1}^{\Gamma-1}\left(\left|\mathcal{I}_{k}\right|^{1/2}+1\right)\right)^{\frac{1}{3}} \geq(\Gamma-1)\left(\log|J|\right)^{\frac{1}{3}}.
\*
Combining the two inequalities, we have
\*
\Gamma-1 \leq\left(\log|J|\right)^{-\frac{1}{3}}\left(\sum_{k=1}^{\Gamma-1}\left(\left|\mathcal{I}_{k}\right|^{1/2}+1\right)\right)^{\frac{1}{3}} \Delta_{[s, e]}^{\frac{2}{3}} \leq \mathcal{O}\left(\left(\log|J|\right)^{-\frac{2}{5}}|\mathcal{J}|^{\frac{1}{5}} \Delta_{[s, e]}^{\frac{4}{5}}+1\right).
\*
To prove the upper bound using $S_{\mathcal{J}}$,  recall the condition $\Delta_{\left[s_{k}, t+1\right]}>\sqrt{\frac{\log|J|}{(t-s_{k})^{1/2}+2}}$. Each distribution switch only creates one interval, then  we have $\Gamma -1 \leq S_\cJ - 1 $.

\paragraph{Proof of Lemma~\ref{lemma:block_upperbound}}
In Section~\ref{sec:upper}, we sketched the proof for an upper bound for regret for block $J$, where we only considered the first $\Gamma-1$ complete intervals. Here we show the omitted details in the proof. Following the technique in \cite{chen2019new}, we define $\cJ^\prime\coloneqq [\tau_i, \tau_i+2^j L]$ to be the block that differs from $\cJ$ only in that $\cJ$ is assumed not triggering the restart. Note that following the same partitioning procedure, we have $ \cJ^\prime = \mathcal{I}_{1}^\prime \cup \mathcal{I}_{2}^\prime \cup \cdots \cup \mathcal{I}_{\Gamma^\prime}^\prime$. We can check that $\Gamma\leq\Gamma^\prime$, $\mathcal{I}_{k}^\prime =  \mathcal{I}_{k}$ for $k = 1,2,\dots, \Gamma-1$. Moreover, let $\cI_\Gamma = [s_{\Gamma},e_{\Gamma}]$ and $\cI^\prime_{\Gamma} = [s^\prime_{\Gamma},e^\prime_{\Gamma}]$. We have $s_{\Gamma} = s^\prime_{\Gamma}$ and $e_{\Gamma}\leq e^\prime_{\Gamma}$.

By direct computation, we bound the first term in \eqref{eq:block_upperbound} by
\*
\sum_{k=1}^{\Gamma}|{\cI_k}|\alpha_{{\cI_k}}  &= \sum_{k=1}^{\Gamma-1}|{\cI_k^\prime}|\alpha_{{\cI_k^\prime}} + |{\cI_\Gamma}|\alpha_{{\cI_\Gamma}}\\
&\leq \sum_{k=1}^{\Gamma-1}|{\cI_k^\prime}|\alpha_{{\cI_k^\prime}} + |{\cI_\Gamma^\prime}|\alpha_{{\cI_\Gamma^\prime}}\\
&\leq \sum_{k=1}^{\Gamma} \sqrt{\left|\mathcal{I}_{k}^\prime\right|} \log \left|\mathcal{I}_{k}^\prime\right|\\
&\leq \sqrt{\Gamma \sum_{k=1}^{\Gamma}|\cI_k^\prime|}\\
&\leq \cO\left( \min\left\{|\mathcal{J}|^{\frac{3}{5}} \Delta_\cJ^{\frac{2}{5}}, \sqrt{S\cJ} \right\}\right),
\*
where the last inequality comes from applying Cauchy-Schwarz inequality and plugging in
\[\Gamma=\mathcal{O}\left(\min \left\{S_{J},\left(\log|J|\right)^{-\frac{2}{5}} \Delta_{\mathcal{J}}^{\frac{4}{5}}|\mathcal{J}|^{\frac{1}{5}}+1\right\}\right).\]

In the following, we upper bound the second term in \eqref{eq:block_upperbound}. Note that $\cI_\Gamma \subset \cI_\Gamma^\prime$,  we only need to bound $\sum_{k=1}^{\Gamma} |{\cI_k^\prime}|\varepsilon_{\cI_k^\prime} \mathbbm{1}\{\varepsilon_{\cI_k^\prime} \geq \alpha_{\cI_k^\prime} \}$.
We follow \cite{chen2019new} and prove the following adapted lemma.
\begin{lemma}
With probability at least $1-\delta$, it holds that
\*
\sum_{k=1}^{\Gamma} |{\cI_k^\prime}|\varepsilon_{\cI_k^\prime} \mathbbm{1}\{\varepsilon_{\cI_k^\prime} \geq \alpha_{\cI_k^\prime} \}\leq \cO\left( \min\left\{|\mathcal{J}|^{\frac{3}{5}} \Delta_\cJ^{\frac{2}{5}}, \sqrt{S\cJ} \right\}\right).
\*
\end{lemma}

\begin{proof}
Define $\cM  = \{k\in [\Gamma]| \varepsilon_{\cI_k^\prime} \geq \alpha_{\cI_k^\prime} \}$.
Let $m_k$ be the index defined in Lemma~\ref{lemma:phase_end}. By definition,
\*
\sum_{k=1}^{\Gamma} |{\cI_k^\prime}|\varepsilon_{\cI_k^\prime} \mathbbm{1}\{\varepsilon_{\cI_k^\prime} \geq \alpha_{\cI_k^\prime} \} & = \sum_{k\in\cM} |{\cI_k^\prime}|\varepsilon_{\cI_k^\prime} \\
&\leq \sum_{k\in\cM}( |{\cI_k^\prime}|-2^{m_k} L)\varepsilon_{\cI_k^\prime}+\sum_{k\in\cM} 2^{m_k} L \times\varepsilon_{\cI_k^\prime} .
\*
Following a similar derivation as in \citet[Lemma 26]{chen2019new}, the first term is bounded by $\cO\left(\sqrt{|\cJ|   }\log T\right)$ with probability at least $1-\delta$. Specifically,
\*
 \sum_{k\in\cM}( |{\cI_k^\prime}|-2^{m_k} L)\varepsilon_{\cI_k^\prime} &=  \sum_{k\in\cM} \sum_{t\in [s_k+2^{m_k} L ,e_k]}\varepsilon_{\cI_k^\prime}\\
 &\leq \sum_{k\in\cM} \sum_{t\in [s_k+2^{m_k} L ,e_k]} C_{5} \cdot \frac{1}{\sqrt{2^{m_k} L}}\\
 &=  \sum_{k\in\cM} \sum_{t\in [s_k^\prime+2^{m_k} L ,e_k^\prime]}C_{5} \cdot \frac{1}{\sqrt{2^{m_k} L}}\mathbbm{1}\{t\leq e_\Gamma\}\\
 &= \varphi(e_\Gamma),
\*
where we define $\varphi(\tau) =  \sum_{k\in\cM} \sum_{t\in [s_k^\prime+2^{m_k} L ,e_k^\prime]}C_{5} \cdot \frac{1}{\sqrt{2^{m_k} L}}\mathbbm{1}\{t\leq \tau\}$. 
By definition, we have $\prob{\phi(e_\Gamma) > \phi(\tau)}\leq \prob{ e_\Gamma> \tau}$. By Lemma~\ref{lemma:phase_end}, if the algorithm has not been restarted till time $\tau$, for all $k$ such that $e_k^\prime \leq \tau$, the algorithm must have missed all opportunities to start an exploration phase with index $m_k$. And for the $k$ with $\tau\in\left[s_{k}^{\prime}, e_{k}^{\prime}\right]$  the algorithm must have missed all opportunities to start a exploration phase with index in $\left[s_{k}^{\prime}, \tau-2^{m_{k}} L\right]$. Define $p_m = \frac{1}{L} 2^{-j / 2} 2^{-m / 2}$. Hence, we have
\*
 \prob{e_\Gamma>\tau} &\leq \prod_{k \in \mathcal{M}} \prod_{t \in\left[s_{k}^{\prime}, e_{k}^{\prime}-2^{m_{k}} L\right]}\left(1-p_{m_{k}} \mathbf{1}\left\{t \leq \tau-2^{m_{k}} L\right\}\right)\\
 &\leq \prod_{k \in \mathcal{M}} \prod_{t \in\left[s_{k}^{\prime}+2^{m_{k}} L, e_{k}^{\prime}\right]}\left(1-q_{m_{k}} \mathbf{1}\{t \leq \tau\}\right)\\
 &\leq \prod_{k \in \mathcal{M}} \prod_{t \in\left[s_{k}^{\prime}+2^{m_{k}} L, e_{k}^{\prime}\right]} \exp \left(-q_{m_{k}} \mathbf{1}\{t \leq \tau\}\right)\\
 &\leq  \prod_{k \in \mathcal{M}} \prod_{t \in\left[s_{k}^{\prime}+2^{m_{k}} L, e_{k}^{\prime}\right]}\left(1-q_{m_{k}} \mathbf{1}\{t \leq \tau\}\right)\\
 &\leq  \exp \left(-   \sum_{k \in \mathcal{M}} \sum_{t \in\left[s_{k}^{\prime}+2^{m_{k}} L, e_{k}^{\prime}\right]} q_{m_{k}} \mathbf{1}\{t \leq \tau\}\right)\\
 &= \exp \left(-   \frac{\varphi(\tau)}{C_5\sqrt{2^j L}} \mathbf{1}\{t \leq \tau\}\right)\\
  &=  \exp \left(-   \frac{\varphi(\tau)}{C_5\sqrt{|\cJ|}} \mathbf{1}\{t \leq \tau\}\right).
\*
Define $z = \left(\frac{1}{\sqrt{L|\cJ|}}+\log(1/\delta)\right)C_5\sqrt{|\cJ|}$ and pick $\tau$  such that $\varphi(\tau) \leq z  \leq \varphi(\tau+1) $. If no such $z$ exists, then $\prob{\phi\left(e_{\Gamma}\right)>\phi(\tau)} = 0$. Then we have $\varphi(\tau) >  \varphi(\tau+1)  - \frac{C_5}{\sqrt{L}}> z - \frac{C_5}{\sqrt{L}}$ and 
\*
\prob{\phi\left(e_{\Gamma}\right)>z} \leq \prob{\phi\left(e_{\Gamma}\right)>\phi(\tau)}\leq \exp\left( - \frac{z}{C_5\sqrt{|\cJ|}} + \frac{C_5}{C_5\sqrt{|\cJ|}\sqrt{L}}\right)  = \delta.
\*
Hence, $\phi\left(e_{\Gamma}\right)\leq \left(\frac{1}{\sqrt{L|\cJ|}}+\log(1/\delta)\right)C_5\sqrt{|\cJ|}$ with probability at least $1-\delta$.

The second term is bounded as 
\*
\sum_{k\in\cM} 2^{m_k} L \times \varepsilon_{\cI_k^\prime}&\leq  \sum_{k\in\cM} 2^{m_k} L \times C_{5}  \frac{1}{\sqrt{2^{m_k} L}}  \\
&= \sum_{k\in\cM} C_5 \sqrt{|\cI_k| }\\
&\leq C_5  \sqrt{\Gamma\sum_{k\in\cM}|\cI_k| } .
\*
Plugging in $\Gamma=\mathcal{O}\left(\min \left\{S_{J},\left(\log|J|\right)^{-\frac{2}{5}} \Delta_{\mathcal{J}}^{\frac{4}{5}}|\mathcal{J}|^{\frac{1}{5}}+1\right\}\right)$ concludes the proof.
\end{proof}

\clearpage
\section{Proof of Theorem~\ref{thm:lowerbound_window_based}}
\label{sec:proof_lower_bound}

Our goal is to prove that for the randomized instance described in Section~\ref{sec:lower}, algorithm RestartLQR with the optimally tuned window size $W$ and exploration noise $\sigma$ yields regret $\Omega(V_T^{1/3} T^{2/3})$.

We first begin by noting that using the sequence of controllers $K_t = K^*(\Theta_t)$ incurs a total cost of at most $\sum_{t \in [T]} J^*_t + \mathcal{O}(S)$, where $S$ denotes the number of switches in the hypothesis. This is because for an interval $\{\tau_1, \tau_1+1,\ldots, \tau_2 \}$ where the dynamics remain fixed at $\theta$ with optimal parameters $p^*,k^*,J^*$, we have
\[ \sum_{t = \tau_1}^{\tau_2} x_t^2 + u_t^2 = \sum_{t = \tau_1}^{\tau_2} x_t^2 + (k^* x_t)^2  = \sum_{t=\tau_1}^{\tau_2} J^* + p^* x_{\tau_1}^2 - p^*\expct{x^2_{\tau_2 + 1}} \leq  \sum_{t=\tau_1}^{\tau_2} J^* + p^* x_{\tau_1}^2.\]
Furthermore, since the optimal controllers yield $|a+b_t k_t^*| = \left| a \frac{1-b_t^2 p_t^*}{1+b_t^2 p_t^*} \right| \leq a = \frac{1}{\sqrt{5}}$, $\expct{x_t}$ is bounded for all $t \in [T]$. 

We next show that the loss for the optimally tuned RestartLQR algorithm is at least $\sum_{t} J^*_t + \Omega\left(V_T^{1/3} T^{2/3}\right)$. We will use the following lemma from \cite{cassel2020logarithmic}.

\begin{lemma}[Lemma~14 in \cite{cassel2020logarithmic}] Let $I = \{\tau_1, \ldots, \tau_2\}$ be an interval with dynamics $a=1/\sqrt{5}$, $b_t = b$ with $|b| \leq 0.05$, $\expct{w_t^2}=\psi^2$, and optimal policy parameters $k^*, J^*$. Then for an arbitrary admissible control policy $\{u_t\}$,
\begin{align}
\label{eqn:regret_ineq1}
      \expct{ \sum_{t\in I }^{} x_t^2 + u_t^2} - |I| J^*  & \geq 0.99 \expct{ \sum_{t \in I } (u_t - k^* x_t)^2} - 4 \psi^2,
\end{align}
as well as:
\begin{align}
\label{eqn:regret_ineq2}
      \expct{ \sum_{t \in I } x_t^2 + u_t^2} - |I| J^*  & \geq \frac{1}{3} \expct{ \sum_{t \in I } u_t ^2} - \frac{5}{6}  \psi^2 (k^*)^2 |I|.
\end{align}
\end{lemma}

We begin by defining the random variables that specify the instance. Let $\{\mu_t\}$ ($t=1,2,\ldots, T$) be the sequence specifying the \textit{magnitude} of changes in $b_t$, defined so that $\mu_1 = \epsilon$, and $\mu_2, \ldots, \mu_T$ are i.i.d.~random variables with the following distribution:
\begin{align*}
    \mu_t &= \begin{cases}
    0.05 & \mbox{with probability } \frac{V_T}{2T}, \\
    \epsilon & \mbox{with probability } \left( \frac{V_T}{T}\right)^{5/6}, \\
    0 & \mbox{otherwise},
    \end{cases}
    \intertext{where}
\epsilon &= 0.05 \cdot \left(V_T/T \right)^{1/6}. 
\end{align*}
Let $\{\chi_t\}$ be the sequence specifying the \textit{sign} of changes in $b_t$, defined so that $\chi_1 = 1$ and $\chi_t$ for $t \geq 2$ are i.i.d.~Rademacher random variables (i.e., $\pm 1$ with equal probability). 
Given the above, the sequence $b_t$ is defined as 
\[ b_t = b_{t-1} \cdot \ind_{\mu_t = 0} + \mu_t \cdot \chi_t. \]
Let 
\[ \mathcal{H}_t = \{ w_s, \eta_s, \sigma_s, \mu_s, \chi_s \}_{s=1}^{t} \]
denote the history of the dynamics and instances until time $t$.
Recall that the RestartLQR($W$) family of algorithms partition the horizon into contiguous non-overlapping windows of size $W$. We will use $I_i = [W \cdot (i-1) + 1 , \ldots, W \cdot i ]$ to denote the $i$-th window. The control for $t \in I_i$ is chosen as $\hat{k}_{(i)} + \sigma_t \eta_t$ where $\eta_t $ are i.i.d.~$\mathcal{N}(0,1)$ and $\sigma_t$ is an arbitrary adapted sequence of exploration energy injected by the algorithm. With some abuse of notation, we will use $k^*(b)$ to denote the optimal linear feedback controller as a function of $b$ (with $a=1/\sqrt{5}$ and $w_t \sim \mathcal{N}(0,\psi^2)$ implicit), and note that $k^*(b) = - k^*(-b)$.

We will partition our windows into three sets:
\begin{enumerate}
    \item $\mathcal{I}_1$: windows $i$ which have at least one $\mu_t = 0.05$ for $t \in I_i$; let $\tau_1(i) \in I_i$  be the first time such that $\mu_{\tau_1(i)} \neq 0$,
    \item $\mathcal{I}_\epsilon$: \textit{pairs} of contiguous windows $(i,i+1)$ with $\mu_t=0$ for all $t \in I_i \cup I_{i+1}$, and $|b_t|=\epsilon$,
    \item $\mathcal{I}_2$: the remaining windows.
\end{enumerate}
Note that this partition is not unique. In particular, there could be many ways to pair up contiguous windows with small $b_t$ and no change of dynamics to create the second set. We pick any such maximal partition.

We can use \eqref{eqn:regret_ineq1} and \eqref{eqn:regret_ineq2} to express the total cost of the algorithm as:
\begin{align}
\nonumber
& \expct{\sum_{t \in [T]} x_t^2 + u_t^2 } - \sum_{t \in [T]} J_t^* \\
& \qquad \qquad \geq \   0.99 \expct{\sum_{t \in T} (u_t - k^*_t x_t)^2} - 4 \psi^2 S \nonumber\\
\nonumber
& \qquad \qquad \geq \sum_{i \in \mathcal{I}_1} 0.99 \expct{ \sum_{t = \tau_1(i)}^{i\cdot W}  (u_t - k^*_t x_t)^2} \\
& \qquad \qquad \quad + \sum_{(i,i+1) \in \mathcal{I}_\epsilon} \left( \frac{1}{3} \expct{ \sum_{t \in I_i } u_t ^2} - \frac{5}{6}  \psi^2 (k^*(\epsilon))^2 W + 0.99 \expct{\sum_{t \in I_{i+1}} (u_t - k^*_t x_t)^2}  \right) - \psi^2 S.
    \label{eqn:lb_inter_1}
\end{align}
Begin by considering the event $\mathcal{E}_{i,1} := \{i \in I_1\}$. Conditioning on this event, $\tau_1(i)$ is uniformly distributed in $I_i$. Furthermore, the sign of $b_{\tau_1(i)}$ is $\pm 1$ with equal probability. We thus bound the contribution to regret due to windows in $\cI_1$ as:
\begin{align}
    \nonumber
    &\expct{ \left. \sum_{t = \tau_1(i)}^{i\cdot W}  (u_t - k^*_t x_t)^2 \ \right| \  \mathcal{E}_{i,1}, \cH_t} \\
    \nonumber
    &=     \expct{ \left. \sum_{t = \tau_1(i)}^{i\cdot W}  ( \sigma_t \eta_t + (\hat{k}_{(i)} - k^*_t) x_t)^2 \ \right| \  \mathcal{E}_{i,1}, \cH_t} \\
    \nonumber
    & \geq  \expct{ \left. \sum_{t = \tau_1(i)}^{i\cdot W}  (\hat{k}_{(i)} - k^*_t)^2 x_t^2 \ \right| \  \mathcal{E}_{i,1}, \cH_t} \\
    \nonumber
    &  \geq  \psi^2 \expct{ \left. \sum_{t = \tau_1(i)}^{i\cdot W}  (\hat{k}_{(i)} - k^*_t)^2  \ \right| \  \mathcal{E}_{i,1}, \cH_t} \\
    \nonumber
    &  \geq  \psi^2 \expct{ \left. \sum_{t = (i-1)W + 1}^{i\cdot W}  \ind_{\{ t \geq \tau_1(i), |b_t|=1 \}}(\hat{k}_{(i)} - k^*_t)^2  \ \right| \  \mathcal{E}_{i,1}, \cH_t} \\
    \nonumber
    &  \geq  \psi^2 \expct{| t: t \geq \tau_1(i) , |b_t|=1 |}
    \expct{ \left. (\hat{k}_{(i)} - k^*(b))^2  \ \right| \  \mathcal{E}_{i,1}, \cH_t},\\
    \intertext{where $b$ denotes a random variable that is $\pm 0.05$ with equal probability; }
    \nonumber
    & \hspace{-0.2in} = \psi^2 \cdot \expct{| t: t \geq \tau_1(i) , |b_t|=0.05 |}  \cdot \left( \frac{1}{2}(\hat{k}_{(i)} - k^*(0.05))^2 +  \frac{1}{2}(\hat{k}_{(i)} + k^*(0.05))^2  \right) \\
    \nonumber
    & \geq  \psi^2 \cdot \expct{| t: t \geq \tau_1(i) , |b_t|=0.05 |}  \cdot k^*(0.05)^2 \\
    \nonumber
    & \geq \psi^2 \cdot \expct{| t: t \geq \tau_1(i) , |b_t|=0.05 |} \cdot \frac{1}{4000}.
\end{align}
Therefore, 
\begin{align*}
    & \sum_{i \in \cI_1} \expct{ \left. \sum_{t = \tau_1(i)}^{i\cdot W}  (u_t - k^*_t x_t)^2 \right| \mathcal{E}_{i,1}, \cH_t} \ =\  \sum_{i } \expct{ \left. \sum_{t = \tau_1(i)}^{i\cdot W}  (u_t - k^*_t x_t)^2 \right| \mathcal{E}_{i,1}, \cH_t}   \expct{\ind_{\cE_{i,1}}} \\
    & \qquad \qquad  \geq \sum_i \psi^2 \cdot \expct{| t: t \geq \tau_1(i) , |b_t|=0.05 |} \cdot \frac{1}{4000} \expct{\cE_{i,1}} \\
    & \qquad \qquad  = \frac{\psi^2}{4000}\sum_i \sum_{t \in I_i} \prob{\mu_t = 0.05} \expct{\min\{ i\cdot W - t + 1, \mbox{Geom}(V_T/T + (V_T/4T)^{5/6}) \}},
\end{align*}
where $\mbox{Geom}(p)$ denotes a Geometric random variable with success probability $p$. For any non-negative integer-valued random variable $X$ with median $\hat{X}$ and non-negative integer $a$, we have the identity,
\begin{align*}
    \expct{\min\{X,a\}} = \sum_{x=1}^a \prob{X \geq x}  \geq \sum_{x=1}^{\min\{a,\hat{X}\}} \prob{X \geq x}  \geq \frac{\min\{\hat{X},a\}}{2}.
\end{align*}
For $X \sim \mbox{Geom}(p)$, we have $\hat{X} \geq \frac{1}{5p}$, which finally gives,
\begin{align}
    \nonumber
    \sum_{i \in \cI_1} \expct{ \left. \sum_{t = \tau_1(i)}^{i\cdot W}  (u_t - k^*_t x_t)^2 \right| \mathcal{E}_{i,1}, \cH_t} 
    & \geq  \frac{\psi^2}{4000}\sum_i \sum_{t \in I_i}\frac{V_T}{2T} \frac{\min\{ W, 0.1(4T/V_T)^{5/6} \}}{2} 
    \\
    \label{eqn:lb_inter_2}
    & = \frac{\psi^2}{8000} V_T \min\left\{ W, 0.1(4T/V_T)^{5/6} \right\}.
\end{align}
Note that if $W = \Omega\left( (T/V_T)^{2/3} \right)$, then \eqref{eqn:lb_inter_2} already gives the regret lower bound of the Theorem. Therefore, henceforth we will assume $W = \cO\left( (T/V_T)^{2/3} \right)$.

Next, we turn to windows in $\cI_\epsilon$. Specifically, pick a pair $(i,i+1)$, and our goal is to bound
\[ \frac{1}{3} \expct{ \sum_{t \in I_i } u_t ^2} - \frac{5}{6}  \psi^2 (k^*(\epsilon))^2 W + 0.99 \expct{\sum_{t \in I_{i+1}} (u_t - k^*_t x_t)^2}. \]
We next invoke yet another useful lemma from \cite{cassel2020logarithmic}. 

\begin{lemma}[Lemma~15 in \cite{cassel2020logarithmic}] Let $\mathbb{P}_+$ and $\mathbb{P}_-$ denote the probability laws of $\{x_{t}\}_{t \in I_i}$ under $b_t = +\epsilon$ and $b_t = -\epsilon$ ($\forall t \in I_i$), respectively. Then, the total variation distance between these is upper bounded as 
\begin{align*}
    TV(\mathbb{P}_+, \mathbb{P}_-) \leq \frac{\epsilon}{\psi} \sqrt{ \expct{\sum_{t \in I_i} u_t^2} }.
\end{align*}
\end{lemma}
We will use the notation of the above lemma for the rest of the proof to bound the regret due to windows $(i,i+1)$. As before, we bound the regret in the window $I_{i+1}$ by:
\begin{align*}
     0.99 \expct{\sum_{t \in I_{i+1}} (u_t - k^*_t x_t)^2} & \geq 0.99 \psi^2 W \expct{ \left( \hat{k}_{(i+1)} - k^*_t \right)^2 }  \\
    & = 0.99 \psi^2 W \left( \frac{1}{2} \expctsub{+}{ \left( \hat{k}_{(i+1)} - k^*(\epsilon) \right)^2 } + \frac{1}{2} \expctsub{-}{ \left( \hat{k}_{(i+1)} + k^*(\epsilon) \right)^2 } \right) .
\end{align*}
Let $F_+,F_-$ denote the distribution of $\hat{k}_{(i+1)}$ under $\mathbb{P}_+,\mathbb{P}_-$, respectively, and let $g_+(k) := \left( \hat{k}_{(i+1)} - k^*(\epsilon) \right)^2 $ and $g_-(k) := \left( \hat{k}_{(i+1)} + k^*(\epsilon) \right)^2$. Note that $g_+, g_-$ are non-negative and 
\[ \frac{1}{2}\left( g_+(k) + g_-(k) \right) \geq k^*(\epsilon)^2. \]
Therefore, 
\begin{align*}
    \frac{1}{2} \expctsub{+}{ \left( \hat{k}_{(i+1)} - k^*(\epsilon) \right)^2 } &+ \frac{1}{2} \expctsub{-}{ \left( \hat{k}_{(i+1)} + k^*(\epsilon) \right)^2 } \\
    &= \frac{1}{2}\int_{\Re} g_+(k) dF_+(k) + \frac{1}{2} \int_{\Re} g_-(k) dF_-(k)  \\
    &= \sup_{F \in \Gamma(F_+,F_+) } \int_{\Re \times \Re} \frac{1}{2}\left( g_+(k_1) + g_-(k_2) \right) dF(k_1,k_2) 
    \intertext{where $\Gamma(F_+,F_-)$ denotes the set of stochastic couplings of measures $F_+,F_-$,}
    & \geq  \sup_{F \in \Gamma(F_+,F_+) } \int_{\Re \times \Re} \frac{1}{2}\left( g_+(k_1) + g_-(k_2) \right) \ind_{\{k_1=k_2\}} dF(k_1,k_2) \\
    & \geq  \sup_{F \in \Gamma(F_+,F_+) } \int_{\Re \times \Re} k^*(\epsilon)^2 dF(k_1,k_2) \\
    & \geq k^*(\epsilon)^2 (1 - TV(\mathbb{P}_+, \mathbb{P}_-)).
\end{align*}
We therefore have,
\begin{align}
    \nonumber
    & \frac{1}{3} \expct{ \sum_{t \in I_i } u_t ^2} - \frac{5}{6}  \psi^2 (k^*(\epsilon))^2 W + 0.99 \expct{\sum_{t \in I_{i+1}} (u_t - k^*_t x_t)^2} \\
    \nonumber
    & \geq \frac{\psi^2 TV(\mathbb{P}_+,\mathbb{P}_-)^2}{3\epsilon^2} + \psi^2 k^*(\epsilon)^2 W \left( 0.99(1 - TV(\mathbb{P}_+,\mathbb{P}_-)) - \frac{5}{6}  \right) \\
    \nonumber
    & \geq \min\left\{  \frac{\psi^2 }{300 \epsilon^2} , \frac{\psi^2 k^*(\epsilon)^2W }{20} \right\} \\
    & \geq \min\left\{  \frac{\psi^2 }{300 \epsilon^2} , \frac{\psi^2 \epsilon^2 W }{200} \right\} .
    \label{eqn:lb_inter_3}
\end{align} 
Since we are assuming $W = \cO\left( (T/V_T)^{2/3} \right) = o\left( (T/V_T)^{5/6} \right) $ (the mean duration between switches in $b_t$), the expected number of pairs $(i,i+1)$ in any maximal choice of $\cI_\epsilon$ is $\Omega\left( T/W \right)$, which gives the total regret contribution due to intervals in $\cI_\epsilon$ of at least 
 \[ T \cdot \min\left\{  \frac{\psi^2  }{300 W \epsilon^2} , \frac{\psi^2 \epsilon^2 }{200} \right\} .\]
The expression above is decreasing in $W$, and for $W = \cO\left( (T/V_T)^{2/3} \right)$ is $\Omega\left(V_T^{1/3}T^{2/3}\right)$. ~ \hfill$\Box$
\end{appendix}



%
%

\end{document}